\pdfoutput=1

\documentclass[11pt]{article}

\usepackage{acl}

\usepackage{times}
\usepackage{latexsym}

\usepackage{graphicx}
\usepackage{todonotes}
\usepackage{hyperref}
\usepackage{caption}
\usepackage{multirow}
\usepackage{subcaption}
\usepackage{enumitem,kantlipsum}
\usepackage{arydshln}
\usepackage{sidecap}
\usepackage{balance}
\usepackage{float}
\usepackage[T1]{fontenc}

\usepackage[utf8]{inputenc}

\usepackage{microtype}

\usepackage{inconsolata}

\usepackage{lipsum}

\newcommand\blfootnote[1]{%
  \begingroup
  \renewcommand\thefootnote{}\footnote{#1}%
  \addtocounter{footnote}{-1}%
  \endgroup
}

\newcommand{\Di}{\texttt{Dynabench}}
\newcommand{\Dii}{\texttt{Waseem}}
\newcommand{\Diii}{\texttt{Hatexplain}}
\newcommand{\Div}{\texttt{Founta}}
\newcommand{\Dv}{\texttt{Toxigen}}
\newcommand{\Dvi}{\texttt{OLID}}
\newcommand{\Dvii}{\texttt{Davidson}}

\title{Probing Critical Learning Dynamics of PLMs \\for Hate Speech Detection}

\author{Sarah Masud$^{*1}$, 
  Mohammad Aflah Khan$^{*1}$, \\
  \bf Vikram Goyal$^1$, Md Shad Akhtar$^1$,
  Tanmoy Chakraborty$^2$\\
  $^1$IIIT Delhi, $^2$IIT Delhi\\
\texttt{\{sarahm,aflah20082,vikram,shad.akhtar\}@iiitd.ac.in,  tanchak@iitd.ac.in}}

\begin{document}
\maketitle
\begin{abstract}
Despite the widespread adoption, there is a lack of research into how various critical aspects of pretrained language models (PLMs) affect their performance in hate speech detection. Through five research questions, our findings and recommendations lay the groundwork for empirically investigating different aspects of PLMs' use in hate speech detection. We deep dive into comparing different pretrained models, evaluating their seed robustness, finetuning settings, and the impact of pretraining data collection time. Our analysis reveals early peaks for downstream tasks during pretraining, the limited benefit of employing a more recent pretraining corpus, and the significance of specific layers during finetuning. We further call into question the use of domain-specific models and highlight the need for dynamic datasets for benchmarking hate speech detection\blfootnote{* Equal Contribution}.
\end{abstract}

\section{Introduction}
\begin{figure}[!t]
\includegraphics[width=\columnwidth]{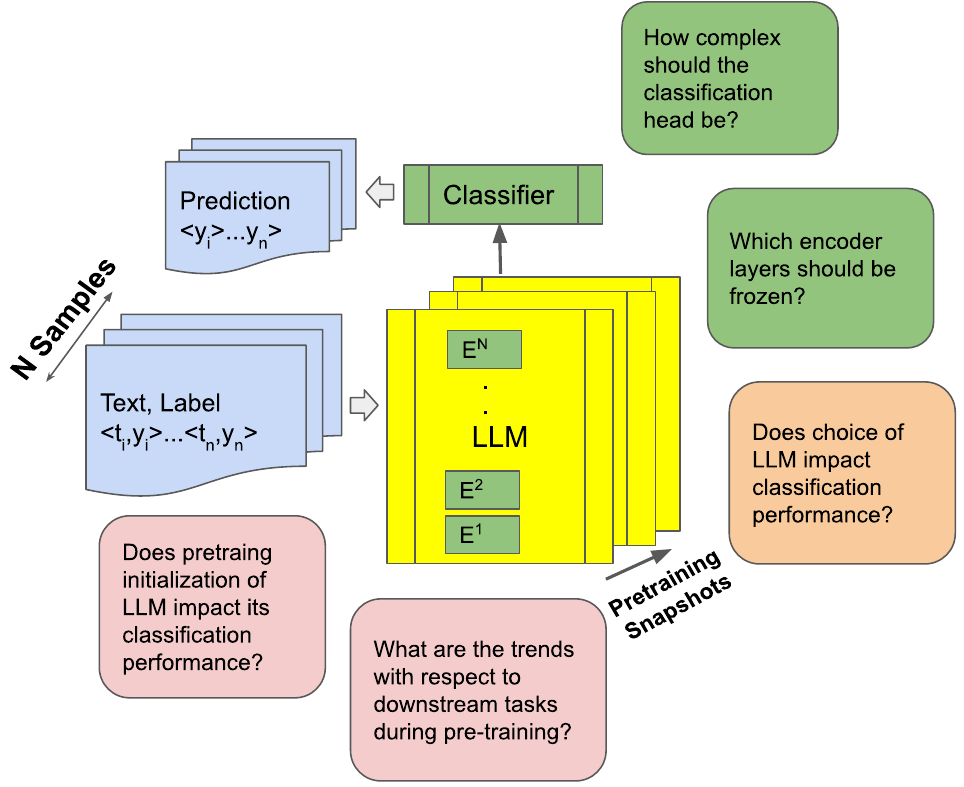}
  \caption{Research Overview: The study comprises five research questions (RQs) to empirically analyze the pretraining and finetuning strategies for PLM variants employed for hate detection. A typical PLM-inspired pipeline involves working with one or more checkpoints, i.e., PLM model weights obtained after pretraining. The checkpoint is then finetuned for downstream tasks by keeping one or more layers of PLM trainable along with a trainable classification head (CH). Finally, the PLM + CH generates predictions on incoming test samples.}
  \label{fig:overview_qa}
\vspace{-3mm}
\end{figure}
The transformer-based language models (LMs) \cite{NIPS2017_3f5ee243,devlin-etal-2019-bert,DBLP:journals/corr/abs-1907-11692} have been a game-changer in NLP. Consequently, researchers have adopted pretrained language models (PLMs) to detect hate speech. However, the choice of the PLM employed for hate detection is often arbitrary and relies on default hyperparameters \cite{ccl}. Despite PLMs being prone to variability in performance \cite{sellam2022multiberts}, there is limited research comparing training settings for subjective tasks like hate speech detection. Note, this study follows the definition of hate speech provided by \citet{waseem-hovy-2016-hateful} -- ``a language targeted at a group or individual intended to derogatory, humiliate, or insult.''  

\begin{table*}[!t]
\scriptsize
\centering
\begin{tabular}{l|l|c|c|c|ccc}
\hline
\multirow{2}{*}{Dataset} &
  \multirow{2}{*}{Source} &
  \multirow{2}{*}{Labels} &
  \multirow{2}{*}{Platform of origin} &
  \multirow{2}{*}{Time of collection} &
  \multicolumn{3}{c}{Dataset size} \\ \cline{6-8} 
   &               &            &                             &                       & \multicolumn{1}{c|}{Train}  & \multicolumn{1}{c|}{Dev}   & Test  \\ \hline
\Dii\ & \citet{waseem-hovy-2016-hateful}        & H, NH      & Twitter                     & Prior to  Jun '16     & \multicolumn{1}{c|}{6077}   & \multicolumn{1}{c|}{2026}  & 2701  \\ 
\Dvii\ & \citet{hateoffensive}      & H, NH      & Twitter                     & Prior to Mar '17      & \multicolumn{1}{c|}{13940}  & \multicolumn{1}{c|}{4647}  & 6196  \\ 
\Div\ & \citet{Founta_Djouvas_Chatzakou_Leontiadis_Blackburn_Stringhini_Vakali_Sirivianos_Kourtellis_2018}      & H, NH      & Twitter                     & March '17 - April '17 & \multicolumn{1}{c|}{33293}  & \multicolumn{1}{c|}{11098} & 14798 \\
\Dvi* & \citet{zampieri-etal-2019-semeval}   & OFF, NOT   & Twitter                     & Prior to Jun '19      & \multicolumn{1}{c|}{9930}   & \multicolumn{1}{c|}{3310}  & 860   \\ 
\Diii\ &  \citet{mathew2021hatexplain}    & H, NH      & Twitter \& Gab              & Jan '19 - June '20    & \multicolumn{1}{c|}{11303}  & \multicolumn{1}{c|}{3768}  & 5024  \\
\Di\ & \citet{vidgen-etal-2021-learning}     & H, NH      & Synthetic (human-generated) & Sept '20 - Jan '21    & \multicolumn{1}{c|}{23143}  & \multicolumn{1}{c|}{7715}  & 10286 \\
\Dv\ & \citet{hartvigsen-etal-2022-toxigen}     & H, NH      & Synthetic (LLM generated)   & Prior to Jul '22      & \multicolumn{1}{c|}{141159} & \multicolumn{1}{c|}{47054} & 62738 \\\hline
\end{tabular}
\caption{Datasets employed in this study. Abbreviation: H: Hate, NH: Not Hate, OFF: Offensive, NOT: Not Offensive. Datasets with * have a predefined train-dev-test split. For others, we take a 75-25\% split for train-test sets, with another 25\% of the train reserved as a development set.}
\label{tab:dataset_info}
\vspace{-5mm}
\end{table*}

\textbf{Research questions.} Figure \ref{fig:overview_qa} provides an overview of our research questions (RQ). We broadly study two critical elements of PLMs by analyzing (i) the impact of different pretraining strategies and (ii) the impact of different finetuning strategies. Section \ref{sec:pretraining} primarily focuses on whether there is a significant performance difference in downstream hate speech detection w.r.t variability in pretraining seeding (RQ1), checkpoints (RQ2), and training corpus (RQ3). Meanwhile, Section \ref{sec:finetuning} deals with layer-level training and its impact on hate speech detection (RQ4). We further examine these setups across five different BERT-based PLMs (RQ5) widely employed for hate detection. \textcolor{black}{While these RQs have been studied in some other aspects of NLP \cite{sellam2022multiberts, 10.1145/3357384.3358028}, their employment for hate speech detection is a unique perspective given the subjective nature of the task.} Each selected question targets a fundamental yet taken-for-granted aspect of PLM through the lens of hate speech detection. We hope this study helps researchers make informed choices, from selecting the underlying PLMs, trainable layers, and classification heads.  

\textbf{Contributions.} \textcolor{black}{While previous studies on hate speech modeling perform hyperparameter tuning, they examine either a single architecture \cite{10.1145/3292522.3326028}, a single PLM \cite{vidgen-etal-2021-learning}, or a single dataset \cite{mathew2021hatexplain}. One of our work's core contributions is to examine different PLMs, seeds, and datasets under one study.} Consequently, we observe that the dynamics of PLMs for hate detection differ significantly from the other use cases \cite{sellam2022multiberts,durrani-etal-2022-transformation}. There are interesting trends in pretraining learning dynamics, with peaks at early checkpoints. We find pretraining over newer data unhelpful. Consequently, on the pretraining end, we observe that general-purpose PLMs with a complex classification head can be as efficient as domain-specific PLMs \cite{caselli-etal-2021-hatebert}. Unlike BERT \cite{ccl}, for mBERT finetuning, the last layer is not the most effective for hate detection. To the best of our knowledge, we are the first to evaluate PLMs' learning dynamics for hate speech detection\footnote{Source Code of our work is available at \url{https://github.com/LCS2-IIITD/HateFinetune}}. \textcolor{black}{Overall, the study examines seven datasets under diverse settings. The aim is not to derive a consistent pattern but rather to examine whether any pattern exists among the datasets w.r.t. different settings discussed in the RQs.}

\section{Related Work}
Early attempts at hate speech detection employed linguistic features \cite{waseem-hovy-2016-hateful} and recurrent architectures \cite{10.1145/3292522.3326028, Badjatiya_2017}. However, with the arrival of the transformer architecture \cite{NIPS2017_3f5ee243}, hate speech tasks also gained a significant boost \cite{mathew2021hatexplain, caselli-etal-2021-hatebert,10.1145/3534678.3539161}. However, most studies adopted the default setting to finetune PLMs.

Meanwhile, deep learning models are criticized to be black boxes. While heuristics such as LIME \cite{10.1145/2939672.2939778} and SHAP \cite{10.5555/3295222.3295230}, among others, attempt to make these models interpretable, they are limited to perturbations in the input space rather than the latent space. More recently, work on mechanistic interpretability \cite{elhage2021mathematical} attempts to understand how transformers build their predictions across layers. Control over high-level properties of the generated text, such as toxicity, can be obtained by tweaking and promoting certain concepts in the vocabulary space \cite{geva-etal-2022-lm}. Interpretability \cite{vijayaraghavan2021interpretable}, finding best practices \cite{DBLP:conf/iclr/KhanYJG23} and sufficiency \cite{balkir-etal-2022-necessity} in hate speech have always been open research areas. 
While toxicity and biases encoded by pretrained PLMs \cite{ousidhoum-etal-2021-probing} is an essential area of research, our work focuses on the downstream finetuning of PLMs for hate detection.

\section{Experimental Setup}
\label{sec:exp}
\textbf{Dataset.} As this research focuses on classifying hateful text, we utilize seven publicly available hate detection datasets in English (Table \ref{tab:dataset_info}). \Dii, \Div, \Dvii\, \& \Dvi\ are chosen based on their prominence in literature. \Dvi\ is obtained from a shared task, and we employ task A of \Dvi. More recently curated datasets, such as \Diii\, as well as synthetically generated ones (either by humans, like \Di\, or by LLMs, like \Dv), are also picked.

\textcolor{black}{\textbf{Note on Dataset Characteristics.}
During our preliminary analysis, we performed data drift experiments to see how distinguishable the HS datasets are from each other \cite{10.1145/3580305.3599896}. From Table \ref{tab:dataset_char}, we observe that, on average, the datasets are differentiable on the latent space with a macro F1 of 60-80\%. \Dv\ was more distinguishable than the rest, with a macro F1 of 85-90\%, yet it does not show major deviations in patterns for the RQs. As \Diii\ provides multiple annotator responses for each sample, we consider those samples as hateful, where a majority of annotators labeled them as either hateful or offensive, and the rest are considered non-hateful.}

\begin{table}[!t]
\centering
\resizebox{\columnwidth}{!}{
\begin{tabular}{l|l|l|l|l|l|l|l}
\hline
\textbf{Dataset}           & Davidson & Dynabench & Founta & Hateexplain & OLID  & Toxigen & Waseem \\ \hline
Davidson & 0.00     &           &        &             &       &         &        \\
Dynabench         & 62.60    & 0.00      &        &             &       &         &        \\
Founta            & 70.26    & 59.47     & 0.00   &             &       &         &        \\
Hateexplain       & 66.23    & 64.12     & 71.91  & 0.00        &       &         &        \\
OLID              & 63.66    & 74.21     & 80.82  & 80.82       & 0.00  &         &        \\
Toxigen           & 91.09    & 85.88     & 80.86  & 91.70       & 94.76 & 0.00    &        \\
Waseem            & 69.47    & 79.06     & 84.59  & 67.70       & 57.20 & 96.00   & 0.00 \\ \hline 
\end{tabular}}
\caption{\textcolor{black}{Data drift experiment measuring the lexical difference between the dataset corpora in macro F1 \%.}}
\label{tab:dataset_char}
\vspace{-3mm}
\end{table}

\textbf{Backbone PLMs} We provide an overview of the various PLMs ({\em aka} backbone models) employed in this study in Table \ref{tab:LLM_details}\footnote{For some models, the release date is not publicly available and is taken to be the publication date of its research.}. As the work focuses on finetuning the most commonly employed LMs for hate speech detection, we focused on the BERT and RoBERTa family of models (PLMS), the same as previous studies on hate speech \cite{antypas-camacho-collados-2023-robust}. Trends common across these models are likely relevant to a broader set of PLMs employed for hate detection. \textcolor{black}{Further note that for RQ1, 2, and 3, only English variants of the PLM are available, necessitating the study to focus on English datasets for uniform comparison.}

\textbf{Classification Head.} We use three seeds hereby referred to as the \emph{MLP seeds} ($ms=\{12,127,451\}$) to initialize the classification head (CH) of varying complexity:
\begin{enumerate}[noitemsep,nolistsep,topsep=0pt,leftmargin=1em]
    \item \textit{Simple CH}: A linear layer followed by Softmax.
    \item \textit{Medium CH}: Two linear layers with intermediate $dim=128$ and intermediate activation function as ReLU followed by a Softmax.
    \item \textit{Complex CH}: Two linear layers with an intermediate $dim=512$, ReLU activation, and an intermediate dropout layer with a dropout probability of $0.1$, followed by a softmax layer. We borrow this setup from \citet{ilan-vilenchik-2022-harald}.
\end{enumerate}

\begin{table}[!t]
\resizebox{\columnwidth}{!}{
\begin{tabular}{p{8em}|l|p{14em}|p{10em}}
\hline
Model    & YoR & Dataset used & Training strategy \\ \hline
BERT \cite{devlin-etal-2019-bert}     & 2018            & Book Corpus \& English Wikipedia & MLM + NSP         \\
mBERT \cite{devlin-etal-2019-bert}     & 2018 & BERT Pretrained on all Wikipedia data for 104 languages with the most representation in Wikipedia & MLM + NSP                         \\ 
HateBERT \cite{caselli-etal-2021-hatebert} & 2020 & RAL-E (Reddit Comments) - 1.5M Comments                                                  & Retrained BERT with MLM Objective \\ 
BERTweet \cite{nguyen-etal-2020-bertweet} & 2020            & 850M Tweets                      & Only MLM          \\ 
RoBERTa \cite{DBLP:journals/corr/abs-1907-11692} & 2019 & Book Corpus, Common Crawler, WebText  \& Stories & Dynamic MLM + NSP \\\hline
\end{tabular}}
\caption{Overview of PLMs employed in this study. YoR is the year of release (either the public model or the source research paper). We also enlist the data source employed for training. The systems use masked language modeling (MLM) and next-sentence prediction (NSP) as pretraining strategies.}
\label{tab:LLM_details}
\vspace{-3mm}
\end{table}

\textbf{Hyperparameter}
All experiments are run with NVIDIA RTX A6000 (48GB), RTX A5000 (25GB) \& Tesla V100 (32GB) GPUs. Significance tests are run with a random seed value of $150$. \textcolor{black}{We employ the two-sided t-test and Cohen-d for measuring the effect size.} We remove emojis, punctuations, and extra whitespaces to preprocess the textual content. URLs and usernames (beginning with '@') are also replaced with <URL> and <USER>, respectively. We train the classifiers for two epochs for all our experiments. The setups employ PLMs that are publicly available on HuggingFace \cite{wolf-etal-2020-transformers}. The classifiers use AdamW optimizer \cite{DBLP:conf/iclr/LoshchilovH19} with a batch size of $16$ and sentences padded to a max length of the respective PLM. \textcolor{black}{We keep the learning rate (LR) at 0.001 (for all RQs) to be in line with the default Adam-W optimizer setting in Huggingface's implementation.} We also use a linear scheduler for the optimizer with a warmup.

\section{Analysis of the Pretrained Backbones}
\label{sec:pretraining}
Variability in pretraining strategies should lead to variability in the performance of downstream tasks. To explore this for hate speech detection, we start with analyzing pretraining weight initialization on the final checkpoint and then move to investigate intermediate checkpoints and pretraining corpus.

\subsection*{\underline{RQ1:} How do variations in pretraining weight initialization of PLMs impact hate detection?}
\label{sec:RQ1}

\textbf{Hypothesis.}
With no guarantee of attaining global minima via gradient descent, some seed initialization of weights during pretraining could lead to better performance downstream. On the one hand, in a study over multiple seeded BERT \cite{sellam2022multiberts}, it was observed that the GLUE benchmark \cite{wang-etal-2018-glue} is susceptible to randomness in finetuning and especially pretraining seed strategy. Meanwhile, for auto-regressive models, it has been observed that the order of training samples during pretraining has a very low correlation with what the final model memorizes \cite{biderman2023pythia}. We hypothesize that hate detection should follow the former patterns.

\begin{table}[]
\centering
\resizebox{\columnwidth}{!}{
\begin{tabular}{l|l|l|l}
\hline
Dataset & Min F1 & Max F1 & ES \\ \hline
\Dii & $S_{451,0}$: 0.675 & $S_{12,10}$: 0.731 & 0.446* \\
\Dvii & $S_{451,0}$: 0.745 & $S_{12,15}$: 0.792 & 0.582** \\
\Div & $S_{12,5}$: 0.872 & $S_{127,20}$: 0.888 & 0.473** \\
\Dvi & $S_{451,0}$: 0.647 & $S_{451,10}$: 0.731 & 0.287* \\
\Diii & $S_{127,5}$: 0.630 & $S_{451,10}$: 0.680 & 0.676** \\
\Di & $S_{451,15}$: 0.625 & $S_{12,20}$: 0.660 & 0.724** \\ 
\Dv & $S_{451,5}$: 0.767 & $S_{127,10}$: 0.771 & 0.226 \\ \hline
\end{tabular}}
\caption{{\bf RQ1:} Comparison of minimum and maximum macro F1 obtained under varying seed combinations by each dataset. $S_{ms, ps}$ represents the combination of MLP seed ($ms$) and pretraining seed ($ps$). ES stands for effect size. ** and * indicate whether the difference in minimum and maximum macro F1 is significant by $\le0.05$ and $\le0.001$ p-value, respectively.}
\label{tab:multibert}
\vspace{-5mm}
\end{table}

\textbf{Setup.}
We utilize the publicly available $25$ different final checkpoints of BERT \cite{sellam2022multiberts}, each trained under the same architecture and hyperparameters but with different random weight (random seed) initializations and shuffling of the training corpus. We randomly picked five pretrained checkpoints for our analysis. The seeds employed for selecting the five checkpoints will be referred to as the {\em pretraining seed set} ($ps=\{0,5,10,15,20\}$). To better capture the impact of pretraining weight randomization, the PLM is frozen, and only the classification head is trained. Further, to control for the randomness in the MLP layer, we use the MLP seeds ($ms$) and run differently-seeded $(ms, ps)$ combination.

\textbf{Findings.}
At the macro level, as outlined in Table \ref{tab:multibert}, the performance appears to be significantly impacted by different seed ($ms, ps$) combinations. We perform a $p$-test on each dataset's overall minimum and maximum macro F1 seed pairs to establish the same. The difference in performance is significant for $5$ out of $7$ datasets with medium to high effect sizes. \textcolor{black}{Similar to prior work \cite{sellam2022multiberts}, we look at the variability in performance when considering one set of seeds to be fixed. Keeping $ms$ constant at the micro-level produces more variability than $ps$ (Appendix \ref{app:rq1}). It follows from the fact that in finetuning settings, the MLP layer initialized with $ms$ is trainable, while the pretrained model initialized with $ps$ may be fully or partially set to non-trainable (fully in our case).} In this investigation, the machine-generated dataset (\Dv) is the only one immune to variation in seeding. \emph{However, due to randomness in weight initialization, the PLMs encode subjectivity across different datasets for hate detection.}

\begin{table}[]
\resizebox{\columnwidth}{!}{
\begin{tabular}{l|lll|lll}
\hline
\multirow{2}{*}{Dataset} & \multicolumn{3}{c|}{Simple} & \multicolumn{3}{c}{Complex} \\ \cline{2-7}
                         & $S_{12}$    & $S_{127}$    & $S_{451}$    & $S_{12}$     & $S_{127}$    & $S_{451}$    \\ \hline
                        \Dii & $C_{3}$: 0.660       & $C_{3}$: 0.668        & $C_{2}$: 0.691         & $C_{2}$: 0.734         & $C_{2}$: 0.738         & $C_{2}$: 0.756         \\
                        \Dvii & $C_{2}$: 0.739        & $C_{2}$: 0.740        & $C_{2}$: 0.775         & $C_{2}$: 0.824         & $C_{3}$: 0.810         & $C_{2}$: 0.764         \\
                        \Div & $C_{3}$: 0.870        & $C_{2}$: 0.861         & $C_{3}$: 0.869         & $C_{2}$: 0.879         & $C_{2}$: 0.880        & $C_{2}$: 0.878        \\
                        \Dvi & $C_{2}$: 0.660        & $C_{2}$: 0.649         & $C_{2}$: 0.654         & $C_{2}$: 0.667         &$C_{2}$: 0.693         & $C_{2}$: 0.672         \\ 
                        \Diii & $C_{2}$: 0.646         & $C_{2}$: 0.666        & $C_{4}$: 0.647 &        $C_{2}$: 0.694
         & $C_{2}$:  0.672         & $C_{2}$: 0.700         \\
                        \Di & $C_{2}$:  0.626       & $C_{2}$: 0.629        &  $C_{2}$: 0.625       & $C_{2}$: 0.627         & $C_{2}$: 0.623         & $C_{2}$: 0.631        \\
                        \Dv & $C_{2}$: 0.733        & $C_{2}$: 0.732         & $C_{2}$: 0.733        & $C_{2}$: 0.764         & $C_{2}$: 0.763        & $C_{2}$: 0.764 \\
                        
                          \hline     
\end{tabular}}
\caption{{\bf RQ2:} We report the $n^{th}$ checkpoint ($C_n$) which leads to maximum macro F1 obtained for simple and complex classification head respectively. For each head, we analyze MLP seeds ($S_i \in ms$).}
\label{tab:simple_complex_max}
\vspace{-5mm}
\end{table}

\subsection*{\underline{RQ2:} How do variations in saved checkpoint impact hate detection?}
\label{sec:RQ2}

\textbf{Hypothesis.}
In RQ1, we examine the variability only at the last checkpoint. Meanwhile, in RQ2, we analyze the trends these models may follow for hate detection over intermediate checkpoints. To study the impact of intermediate checkpoints on downstream tasks, \citet{elazar2023measuring} released $84$ intermediate pretrained checkpoints, one for each training epoch of the RoBERTa. This question is necessary as we hypothesize the model's performance will grow during the early checkpoints and then saturate. It should allow one to find a sweet spot to pretrain task-specific PLMs for a shorter duration, saving compute resources.
 
\textbf{Setup.}
Provided by \citet{elazar2023measuring}, we employ the $84$ RoBERTa pretraining checkpoints ($C_n \in C_1, C_2,\ldots, C_{84}$).  In our analysis, each pretrained checkpoint PLM is frozen, and simple and complex classification heads are trained. We train a classification head for each pretrained checkpoint separately for all MLP seeds ($ms$).

\textbf{Findings.}
Contrary to our hypothesis, we observe the performance peaks early (mostly around checkpoint 2) and then rapidly falls. This trend is consistent across different datasets, seeds, and CH complexity as captured by the highest macro F1 reported in Table \ref{tab:simple_complex_max} and Appendix \ref{app:rq2}. The trends in performance indicate that each checkpoint possesses hate detection capacity to varying degrees. We extend our analysis of the superiority of early checkpoints, especially checkpoint \#2 over \#3, with varying learning rates (LR), -- 0.001 (default), 0.01, and 0.1. Averaged across the three MLP seeds, we observe that for a given quadruple <dataset, learning rate, checkpoint, classifier complexity> triplet, checkpoint \#2 is consistently at par with checkpoint \#3, as highlighted by the difference (diff) row in Table \ref{tab:lr_analysis}.
The analysis suggests that a fully pretrained model may not be necessary for hate-related tasks. We concur this may be due to a mismatch between the model's training on well-written datasets such as Wikipedia and Book Corpus and the noisy nature of hate speech. \emph{When the model has not yet fully learned the English language syntax, it could be better suited to capture the noisy information in the hate speech text.}

\begin{table*}[!h]
\resizebox{\textwidth}{!}{
\begin{tabular}
{l|l|l|l|l|l|l|l|l|l}
\hline
CH      & Checkpoints  & LR    & Davidson & Dynabench & Founta & Hateexplain & OLID  & Toxigen & Waseem \\ \hline
Simple  & C2           & 0.001 & 0.75     & 0.63      & 0.867  & 0.657       & 0.653 & 0.73    & 0.637  \\
        & C3           & 0.001 & 0.547    & 0.553     & 0.86   & 0.62        & 0.517 & 0.72    & 0.653  \\ 
        & Diff (C2-C3) &       & 0.203    & 0.077     & 0.007  & 0.037       & 0.136 & 0.01    & -0.016 \\ \hline
Complex & C2           & 0.001 & 0.78     & 0.627     & 0.88   & 0.687       & 0.677 & 0.76    & 0.743  \\ 
        & C3           & 0.001 & 0.763    & 0.577     & 0.857  & 0.613       & 0.55  & 0.74    & 0.69   \\ 
        & Diff (C2-C3) &       & 0.017    & 0.05      & 0.023  & 0.074       & 0.127 & 0.02    & 0.053  \\ \hline
Simple  & C2           & 0.01  & 0.813    & 0.493     & 0.827  & 0.683       & 0.657 & 0.73    & 0.743  \\
        & C3           & 0.01  & 0.76     & 0.52      & 0.843  & 0.543       & 0.623 & 0.72    & 0.72   \\ 
        & Diff (C2-C3) &       & 0.053    & -0.027    & -0.016 & 0.14        & 0.034 & 0.01    & 0.023  \\ \hline
Complex & C2           & 0.01  & 0.837    & 0.593     & 0.863  & 0.623       & 0.617 & 0.73    & 0.753  \\
        & C3           & 0.01  & 0.643    & 0.517     & 0.867  & 0.617       & 0.597 & 0.72    & 0.723  \\ 
        & Diff (C2-C3) &       & 0.194    & 0.076     & -0.004 & 0.006       & 0.02  & 0.01    & 0.03   \\ \hline
Simple  & C2           & 0.1   & 0.75     & 0.52      & 0.777  & 0.62        & 0.577 & 0.72    & 0.75   \\ 
        & C3           & 0.1   & 0.76     & 0.543     & 0.823  & 0.517       & 0.567 & 0.717   & 0.68   \\ 
        & Diff (C2-C3) &       & -0.01    & -0.023    & -0.046 & 0.103       & 0.01  & 0.003   & 0.07   \\ \hline
Complex & C2           & 0.1   & 0.76     & 0.35      & 0.487  & 0.543       & 0.527 & 0.447   & 0.677  \\ 
        & C3           & 0.1   & 0.45     & 0.35      & 0.57   & 0.467       & 0.42  & 0.433   & 0.71   \\ 
        & Diff (C2-C3) &       & 0.31     & 0         & -0.083 & 0.076       & 0.107 & 0.014   & -0.033 \\ \hline
\end{tabular}}
\caption{\textbf{RQ2: } Macro F1 for checkpoints 2 and 3 with varying LR (0.001,0.01,0.1) and classification head (CH) as simple and complex. Diff (C2-C3) depicts the difference in performance of two checkpoints.}
\label{tab:lr_analysis}
\end{table*}

\subsection*{\underline{RQ3:} Does newer pretraining data impact downstream hate speech detection?}
\label{sec:RQ3}

\begin{figure*}[!t]
  \centering
\includegraphics[width=0.85\textwidth]{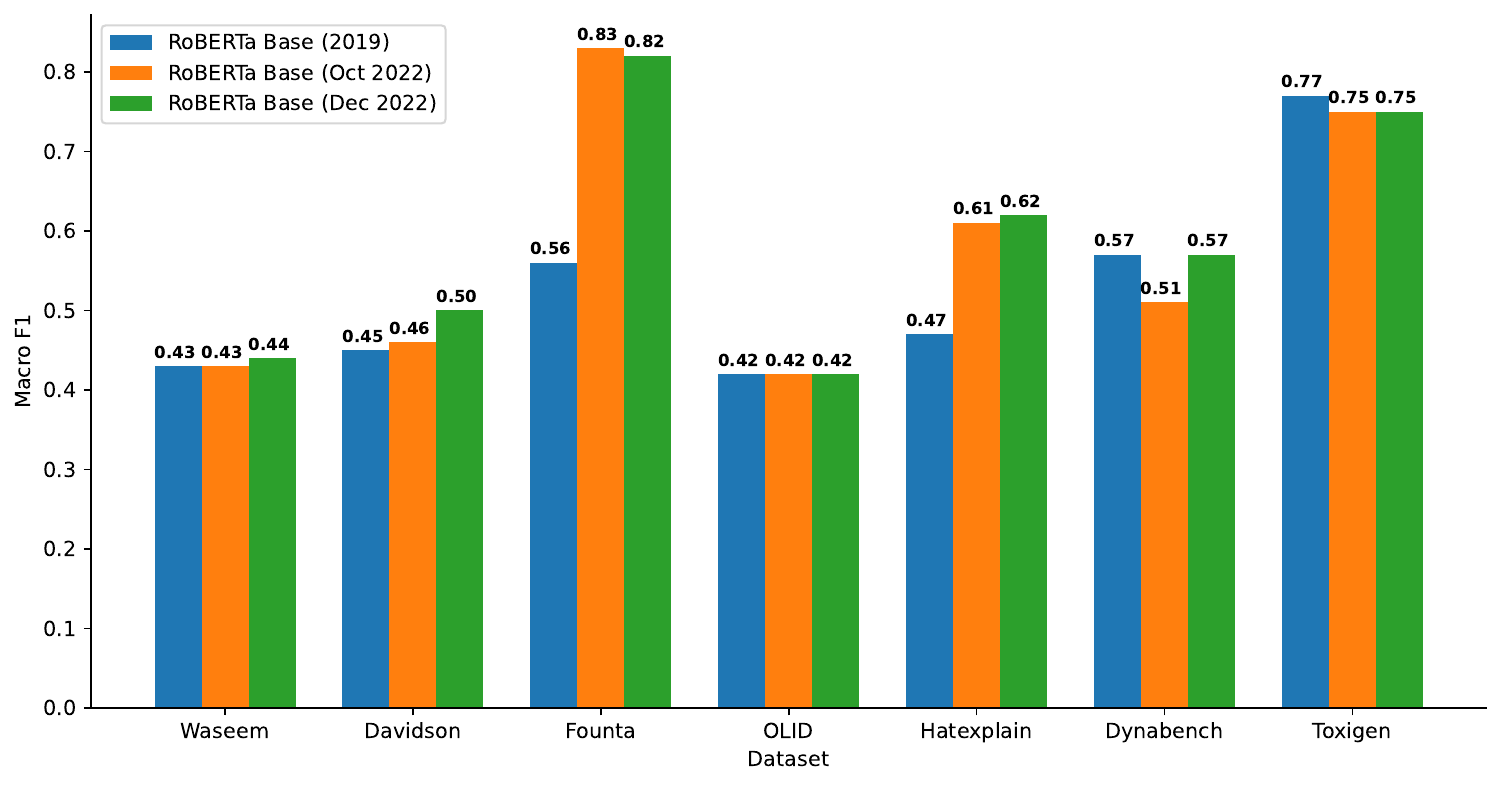}
  \caption{{\bf RQ3:} Macro F1 on different datasets finetuned with an MLP classifier on RoBERTa variants. The variants employed are from June 2019 ($R_{J19}$), October 2022 ($R_{O22}$), and December 2022 ($R_{D22}$). Each variant is trained on a training corpus from Wikipedia, and Common-Crawl is curated and updated before the date associated with the model. $R_{J19}$ is the original RoBERTa model and $R_{O22}$ and $R_{D22}$ are its more recent variants.}
  \label{fig:hf_monthly}
\end{figure*}

\textbf{Hypothesis.}
Hate speech is evolving and often collected from the web in a static/one-time fashion. Pretraining/continued training PLMs on more recent data should capture the emerging hateful world knowledge and enhance the detection of hate.

\textbf{Setup.}
We use checkpoints released by the Online Language Modeling Community\footnote{\url{https://huggingface.co/olm}} (details on OLM provided in Appendix \ref{app:rq3}) for RoBERTa variants trained on more recent data from October ($R_{O22}$) and December 2022 ($R_{D22}$) respectively. We compare these variants against RoBERTa initially released in June 2019 ($R_{J19}$). 

\begin{figure*}[!t]
    \centering
    \begin{subfigure}[b]{0.49\columnwidth}
        \includegraphics[width=\linewidth]{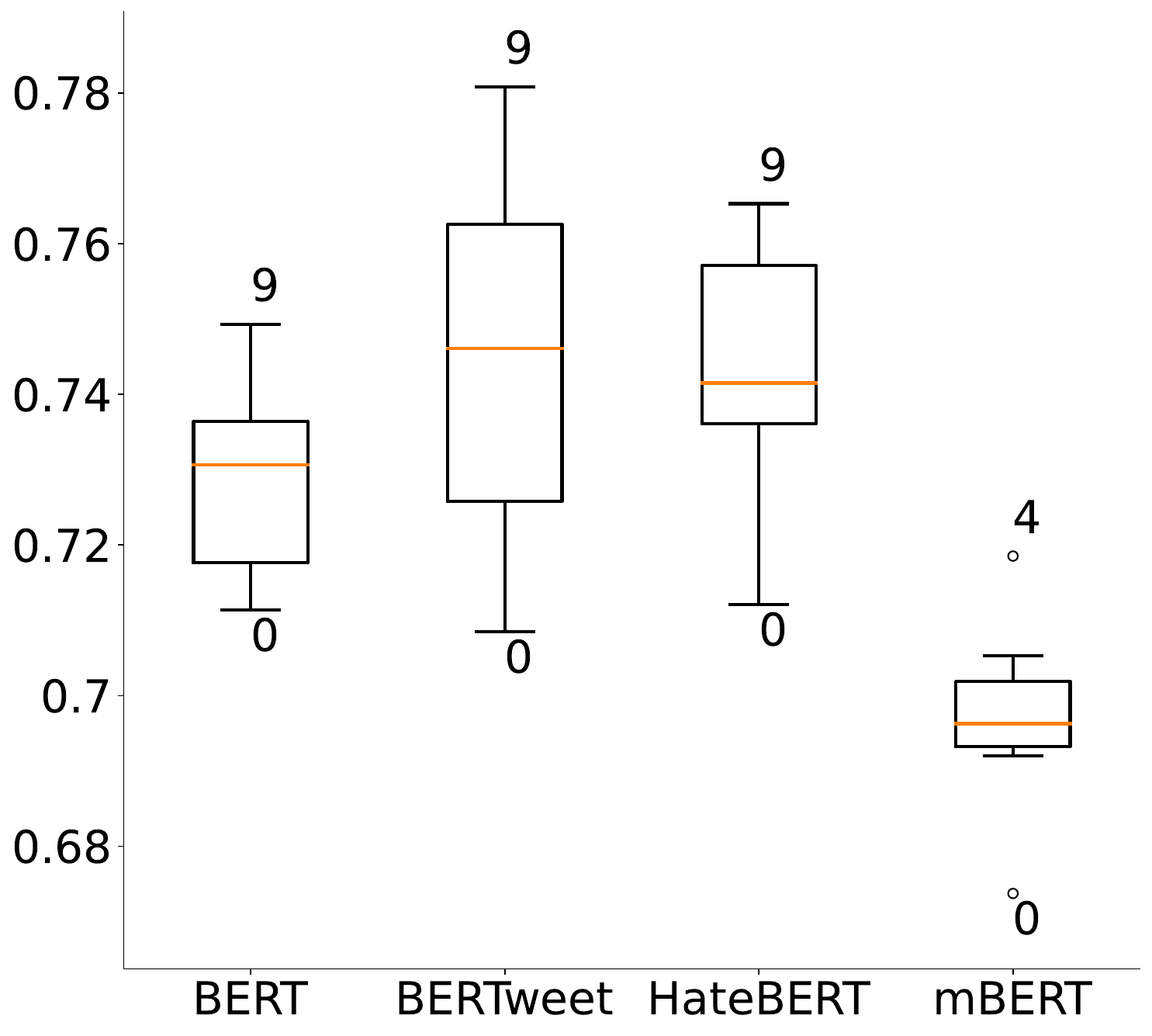}
        \caption{Dynabench}
    \label{fig:sub_dynabench_label_boxplot_one_layer_at_a_time}
    \end{subfigure}
    \begin{subfigure}[b]{0.49\columnwidth}
        \includegraphics[width=\linewidth]{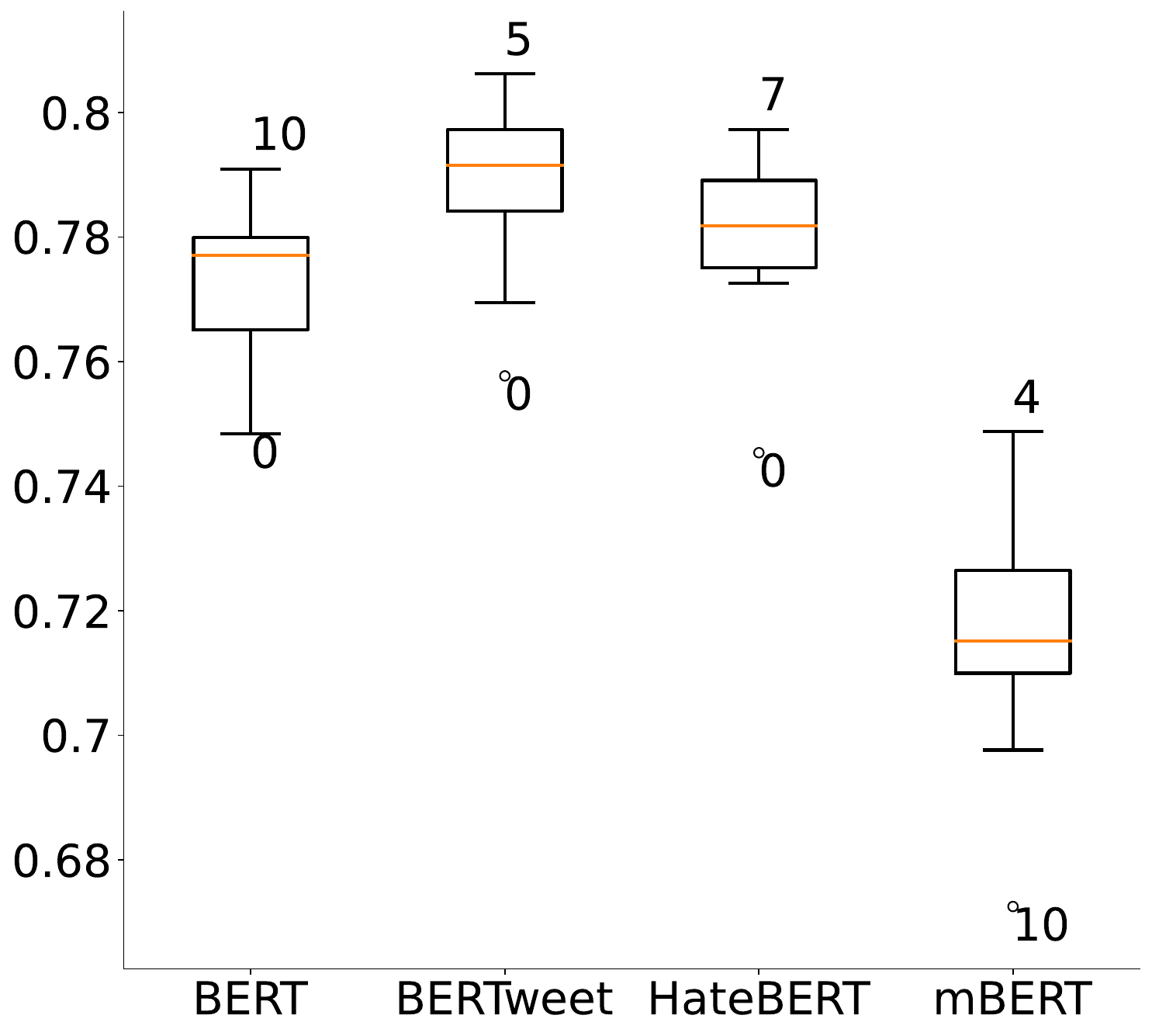}
        \caption{OLID}
        \label{fig:sub_olid_taska_boxplot_one_layer_at_a_time}
    \end{subfigure}
    \begin{subfigure}[b]{0.49\columnwidth}
       \includegraphics[width=\linewidth]{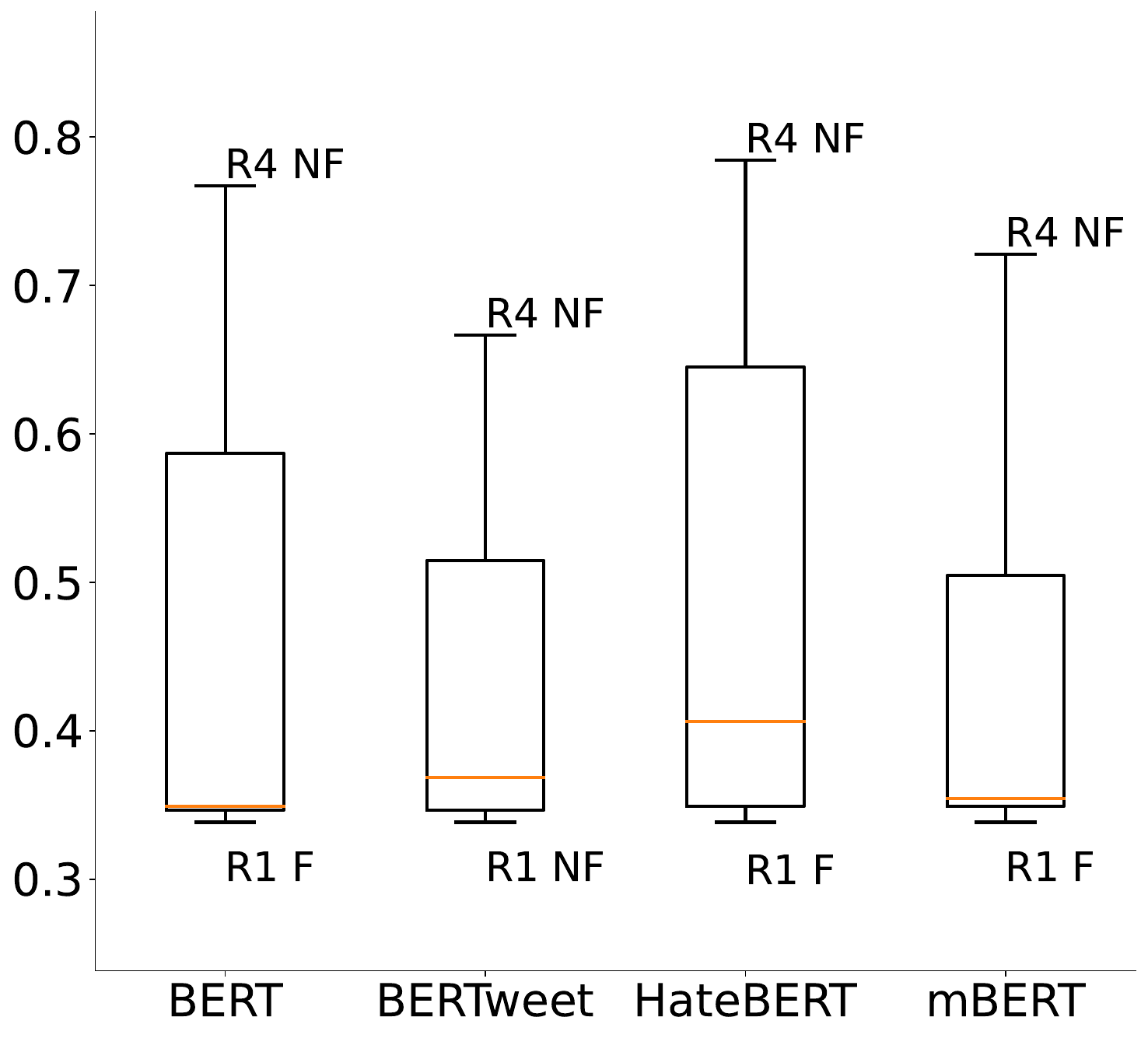}
        \caption{Dynabench}
        \label{fig:dynabench_label_boxplot_grouped}
    \end{subfigure}
    \begin{subfigure}[b]{0.49\columnwidth}
        \includegraphics[width=\linewidth]{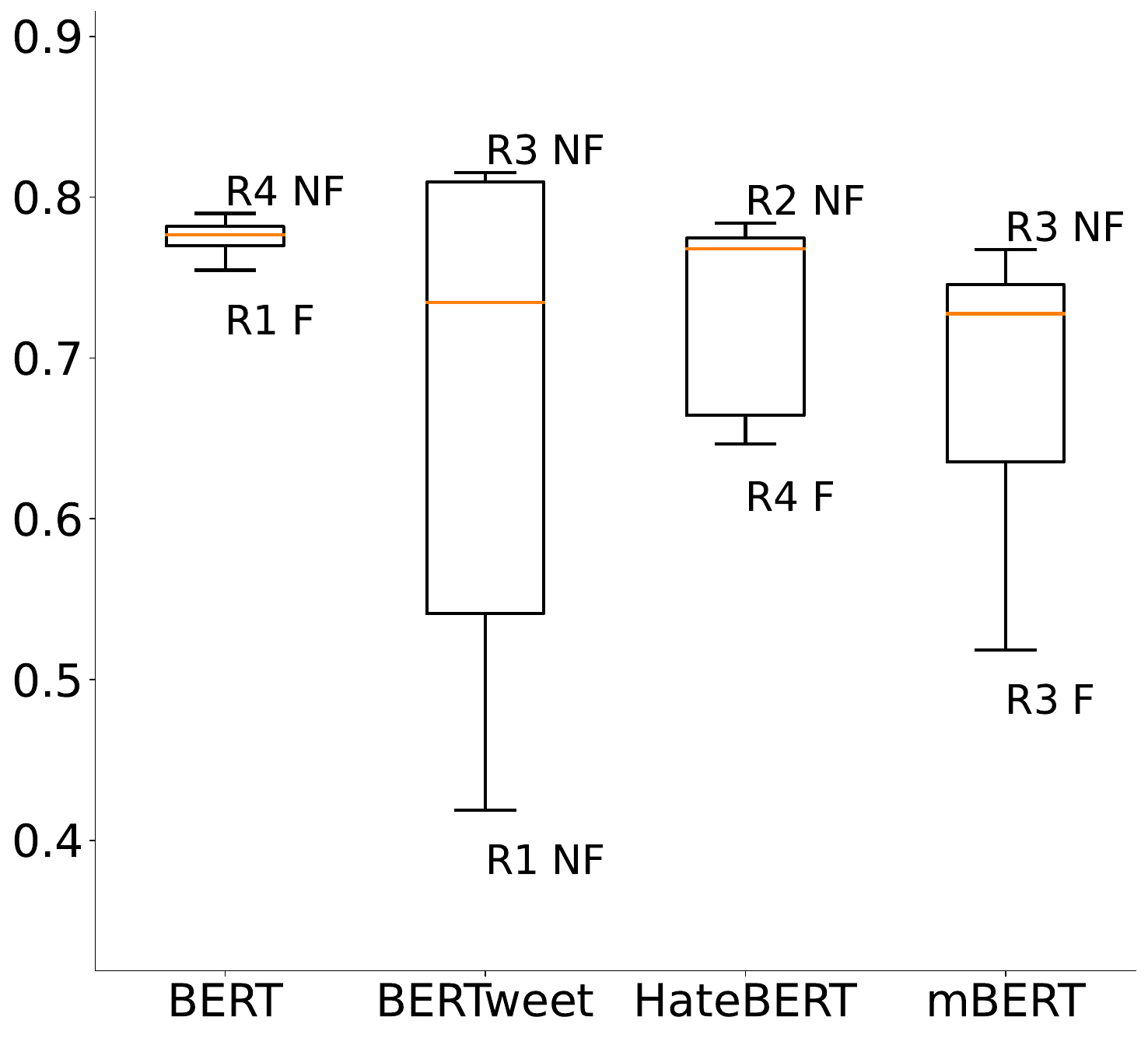}
        \caption{OLID}
        \label{fig:olid_taska_boxplot_grouped}
    \end{subfigure}
    \caption{{\bf RQ4:} (a) \Di\ and (b) \Dvi\ -- Descriptive statistics of macro F1 when finetuning on top of individual layers of the BERT-variant highlighting the layer ($L_i$) that on average over MLP seeds ($ms$) leads to minimum and maximum macro F1. Here, the $L_i$ is trainable while other layers are frozen. (c) \Di\ and (d) \Dvi\ -- Descriptive statistics of macro F1 when finetuning while constraining a region of layers to be frozen (Suffix F) or non-frozen while all others are frozen (Suffix NF) for different BERT-variant highlighting the region ($R_i$) that on average over MLP seeds ($ms$) leads to minimum and maximum macro F1.
    }
\label{fig:one_layer_at_a_time}
\end{figure*}

\textbf{Findings.}
\textcolor{black}{To assess the impact of differently updated PLMs on downstream hate detection, the performance should be interpreted at the individual dataset level and not across datasets.} Figure \ref{fig:hf_monthly} reveals that only three datasets register a sharp jump in performance. 
We attribute this to the fact that most of the datasets employed in this study were collected years ago (Table \ref{tab:dataset_info}). Consequently, events present in these datasets were already sufficiently represented in the original model ($R_{J19}$). \textcolor{black}{Interestingly, the $25$ macro F1 jump for \Div\ may indicate that the models may have seen the data before. Previous literature hypothesized the same when they observed a substantial improvement in NLP performance \cite{zhu2023multilingual}.} \emph{The findings in RQ3 shed light on the problem of stale hate speech datasets and highlight the need to address the dynamic nature of hate speech.} 

\section{Analysis of the Finetuning Schemes}
\label{sec:finetuning}
During finetuning, the PLM layers closer to the classification head capture the maximum task-specific information \cite{durrani-etal-2022-transformation}. Hence, setting the lower layers parameters untrainable is a standard finetuning practice. While layer-wise analyses have been explored in various NLP tasks \cite{de-vries-etal-2020-whats, 10.1145/3357384.3358028}, a comprehensive examination across models, datasets and finetuning scenarios has been notably absent in the hate speech domain. Experiments in this section are run on four BERT variants -- BERT \cite{devlin-etal-2019-bert}, BERTweet \cite{nguyen-etal-2020-bertweet}, HateBERT \cite{caselli-etal-2021-hatebert}, and Multilingual-BERT (mBERT) \cite{devlin-etal-2019-bert}.

\begin{table*}[!h]
\resizebox{\textwidth}{!}{
\begin{tabular}{l|lll|lll|lll|lll}
\hline
\multirow{2}{*}{Dataset} & \multicolumn{3}{l}{BERT} & \multicolumn{3}{l}{BERTweet} & \multicolumn{3}{l}{HateBERT} & \multicolumn{3}{l}{mBERT} \\ \cline{2-13}
       & Min F1 & Max F1 & ES & Min F1 & Max F1 & ES & Min F1 & Max F1 & ES & Min F1 & Max F1 & ES \\ \hline

waseem & $S_{12},L_6$: 0.758 & $S_{12},L_{11}$: 0.806 & 0.484** & $S_{127},L_6$: 0.758 & $S_{127},L_{11}$: 0.810 & 0.944** & $S_{451},L_1$: 0.752 & $S_{127},L_{10}$: 0.813 & 0.619** & $S_{451},L_9$: 0.732 & $S_{127},L_5$: 0.793 & 0.617**\\

davidson & $S_{12},L_{11}$: 0.887        & $S_{451},L_4$: 0.931         & 0.854**    & $S_{12},L_6$: 0.899         &  $S_{12},L_5$: 0.935      & 1.824** & $S_{12},L_{10}$: 0.904**     & $S_{127},L_5$: 0.932        & 0.561**        & $S_{12},L_{10}$: 0.852     & $S_{451},L_4$: 0.922        & 1.367**    \\
 
founta & $S_{12},L_7$: 0.916 & $S_{127},L_5$: 0.929 & 0.485** & $S_{127},L_0$: 0.918 & $S_{451},L_3$: 0.930 & 0.486** & $S_{12},L_2$: 0.915 & $S_{12},L_9$: 0.928 & 0.484** & $S_{12},L_{11}$: 0.890 & $S_{12},L_4$: 0.924 & 1.120** \\

olid & $S_{127},L_0$: 0.732 & $S_{451},L_{11}$: 0.802 & 0.420* & $S_{12},L_0$: 0.747 & $S_{127},L_9$: 0.817 & 0.438* & $S_{451},L_0$: 0.738 & $S_{127},L_8$: 0.806 & 0.383* & $S_{127},L_{10}$: 0.624 & $S_{451},L_4$: 0.764 & 0.595** \\

hatexplain & $S_{451},L_{11}$: 0.639 & $S_{12},L_10$: 0.766 & 1.807** & $S_{12},L_6$: 0.586 & $S_{12},L_9$: 0.770 & 2.616** & $S_{12},L_7$: 0.638 & $S_{12},L_4$: 0.766 & 1.671** & $S_{451}L_9$: 0.615 & $S_{12},L_7$: 0.739 & 1.796** \\

dynabench & $S_{127},L_6$: 0.665        & $S_{451},L_9$: 0.756         & 2.082**    & $S_{12},L_0$: 0.705         &  $S_{127},L_{11}$: 0.783       & 1.824** & $S_{127},L_0$: 0.706     & $S_{451},L_{11}$: 0.770        & 1.564**        & $S_{12},L_0$: 0.635     & $S_{451},L_4$: 0.720        & 1.737**    \\

toxigen & $S_{12},L_0$: 0.767 & $S_{12},L_{11}$: 0.806 & 2.126** & $S_{12},L_1$: 0.0.786 & $S_{12},L_{11}$: 0.827 & 2.621** & $S_{127},L_0$: 0.775 & $S_{127},L_{11}$: 0.816 & 2.386** & $S_{451},L_0$: 0.746 & $S_{12},L_4$: 0.777 & 1.821** \\ \hline

\end{tabular}
}
\caption{\textbf{RQ4:} Comparison of $L_i^{th}$ layer which leads to minimum and maximum macro F1. Note the layers for the BERT-variant may come from different MLP seed values ($S_{ms}$). ES stands for effect size. ** and * indicate whether the difference in minimum and maximum macro F1 is significant by $\le0.05$ and $\le0.001$ $p$-value, respectively.}
\label{tab:layer_wise}
\vspace{-3mm}
\end{table*}

\subsection*{\underline{RQ4:} What impact do individual/grouped layers have on hate detection?}
\label{sec:RQ4}
Different layers or groups of layers in the PLM will be of varying importance for hate detection. Borrowing from the popular finetuning settings \cite{ccl}, one expects training the last few (higher) layers to yield better than training earlier (lower) layers. Further, the setting where more layers are trainable is likely better, giving the model more ability to learn the latent space.

\textbf{Setup.}
We freeze (set to non-trainable) all parameters except the probed layer and the classification head initialized with MLP seeds ($ms$). We probe the impact of layers beginning with the analysis of setting (un)trainable individual layers $L_1, L_2,\ldots, L_{12}$ and then setting (un)trainable groups of layers, {\em aka} region. A $12$ layer PLM comprises $4$ regions ($R_1, R_2, R_3, R_4$) of $3$ consecutive layers  with $R_1=\{L_1, L_2, L_3\}$ and so on. For the layer-wise case the classification head is placed on top of the trainable layer.

\textbf{Findings.}
Table \ref{tab:layer_wise} shows that trainable higher layers (closer to the classification head) lead to higher macro-F1 for most BERT-variants. However, no single layer emerges as a clear winner across all datasets and models, as illustrated in Figure \ref{fig:one_layer_at_a_time}(a,b). When examining specific datasets, such as \Di\ in Figure \ref{fig:sub_dynabench_label_boxplot_one_layer_at_a_time}, it appears that layer \#$9$ is quite dominant, while layer \#$0$ consistently performs poorly across all models. On the other hand, in the case of \Dvi\ (Figure \ref{fig:sub_olid_taska_boxplot_one_layer_at_a_time}), no such trend is observed. The variation in macro F1 when keeping the same MLP seed ($ms$) across BERT-variants is enlisted in Appendix \ref{app:rq4}. Here, we observe that, on average, \Dvii\ and \Div\ seem to be favoring the lower layer for max F1; however, looking at Table 11, we again see that across seeds, \Dvii\ is the only dataset that significantly reaches Max F1 via lower layers. However, overall, the trend for higher layers leading to substantially better performance holds significantly for 5 out of 7 datasets and partially for \Div.

Interestingly, we also observe that layer-wise trends for generating maximum macro F1 are more similar for BERT and BERTweet than BERT-HateBERT or BERTweet-HateBERT comparisons (Table \ref{tab:layer_wise}). Further, the notion of higher layers being important applies to BERT, HateBERT, and BERTweet; the results do not hold for mBERT. As we observe from Table \ref{tab:layer_wise} for mBERT, layer \#$4$ seems to dominate across datasets. \textcolor{black}{While obtaining the best performance from the middle layers of PLMs is counterintuitive in a general setup, similar behavior regarding mBERT has been reported earlier \cite{de-vries-etal-2020-whats}. We hypothesize that this behavior stems from mBERT's need to be simultaneously equally generalized vs. informative for all languages. Thus, the higher dependence on mBERT's lower layers may stem from training on a corpus of multiple languages.}

\begin{figure*}[!t]
    \centering
    \begin{subfigure}[b]{0.49\textwidth}
        \includegraphics[width=\linewidth]{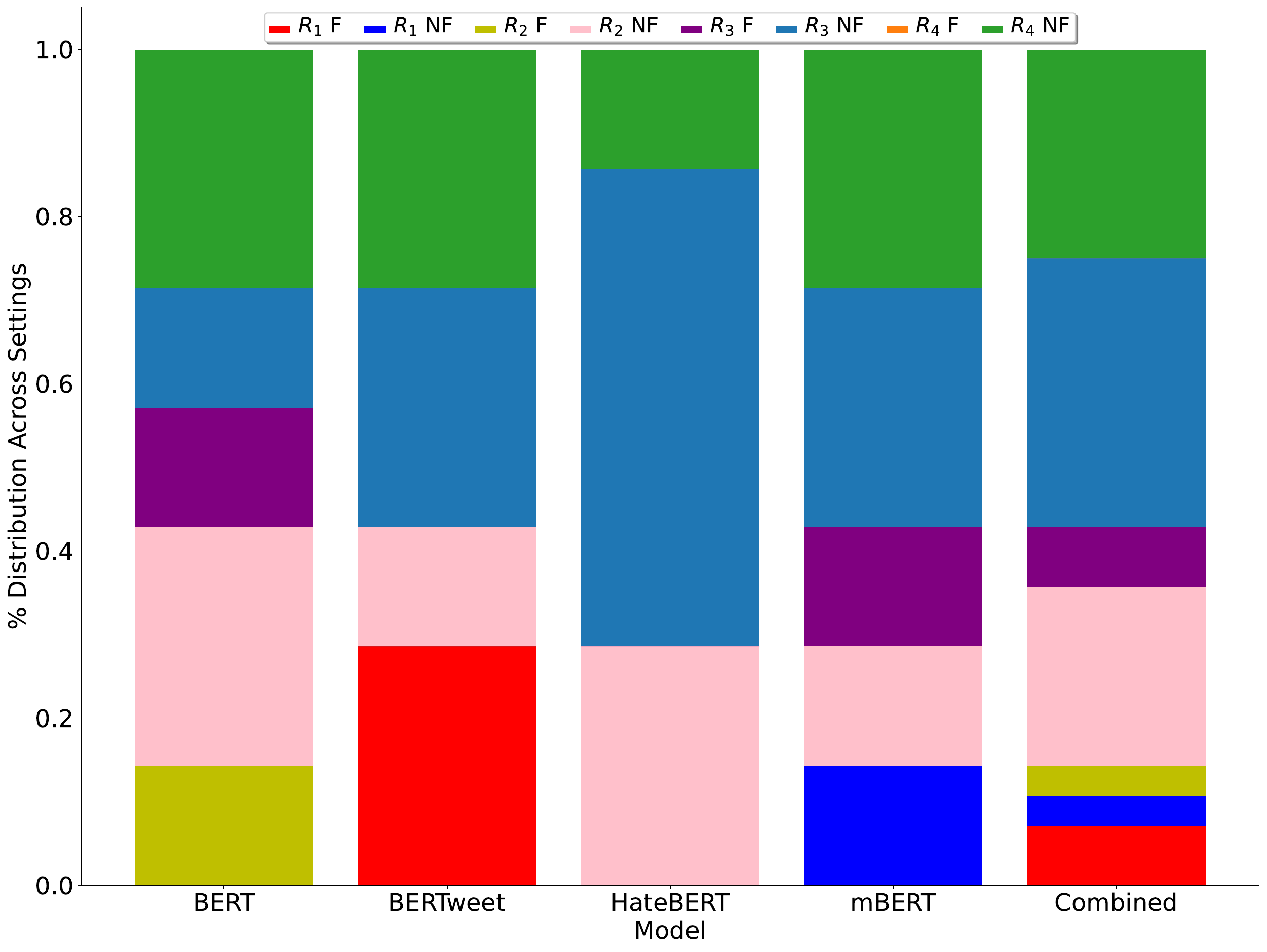}
        \caption{Best region distribution.}
        \label{fig:best_regions}
    \end{subfigure}
    \hfill
    \begin{subfigure}[b]{0.49\textwidth}
        \includegraphics[width=\linewidth]{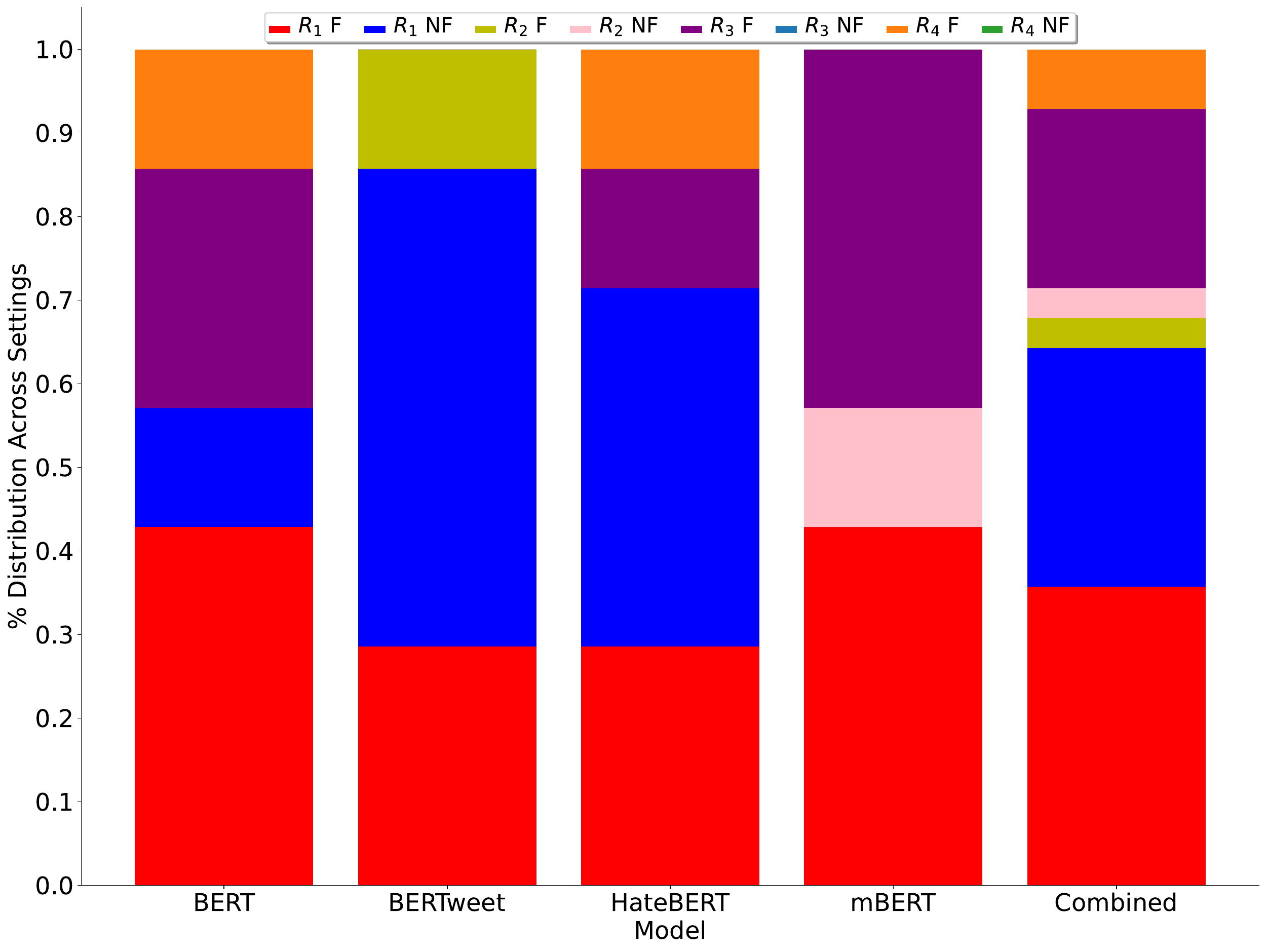}
        \caption{Worst region distribution.}
        \label{fig:worst_regions}
    \end{subfigure}
    \caption{{\bf RQ4:} Percentage distribution of best and worst performing regions across datasets. The divisions on each bar enlist the \% of datasets where the given configuration performs best (a) or worst (b) for a BERT-variant. Combined captures the overall trend across all BERT-variants and datasets. Region $R_1$ includes layers $L_1$ to $L_3$, $R_2$ from $L_4$ to $L_6$, $R_3$ from $L_7$ to $L_{9}$ and $R_4$ from $L_{10}$ to $L_{12}$. Suffix $F$ implies that the region was frozen while other regions were trainable, and the $NF$ suffix implies all other regions were frozen while only that region was trainable.}
    \label{fig:grouped_dist}
\vspace{-2mm}
\end{figure*}

Our findings on region-wise analysis indicate that training the last region performs better than the other settings where only other regions is trained (as shown in Figure \ref{fig:grouped_dist}), i.e., the latter regions are more likely to be better than earlier regions (Figure \ref{fig:best_regions}). Also, when the last region is frozen, it is never the best combination for any dataset or model (Figure \ref{fig:worst_regions}), further validating the status quo. However, no clear region dominates significantly across all datasets (Appendix \ref{app:rq4}). In the case of \Di\ (Figure \ref{fig:dynabench_label_boxplot_grouped}), when $R_4$ is not frozen, it performs the best consistently, while $R_1$ being frozen performs the worst consistently. This is not so black and white for all datasets, as seen in the case of \Dvi\ (Figure \ref{fig:olid_taska_boxplot_grouped}), where there is no one best scheme across models. \emph{In general, layers closer to the classification head appear more critical for hate detection, except in the case of mBERT.}

\subsection*{\underline{RQ5:} Does the complexity of the classifier head impact hate speech detection?}
\label{sec:RQ5}

\textbf{Hypothesis.}
There is an increasing trend in obtaining domain-specific PLMs that are continuously pretrained on domain corpus. Meanwhile, when finetuning, most downstream tasks employ a simple classification head to retain maximum latent information from the pretrained PLMs. In reproducing the work by \cite{ilan-vilenchik-2022-harald}, we observed their use of a complex classification head for HateBERT outperformed a simple one. It prompts the study of the relationship between PLMs and CHs. We hypothesize that employing a relatively complex classification head should perform better than its simpler counterpart.   

\textbf{Setup.}
We run our experiments on three classification heads (CH) of three complexity levels -- simple, medium, and complex (described in Section \ref{sec:exp}). The pretrained model is frozen for this set of experiments to capture the variability introduced by the trainable CH's complexity.

\begin{figure*}[!t]
    \centering
    \begin{subfigure}[b]{0.49\textwidth}
        \includegraphics[width=\linewidth]{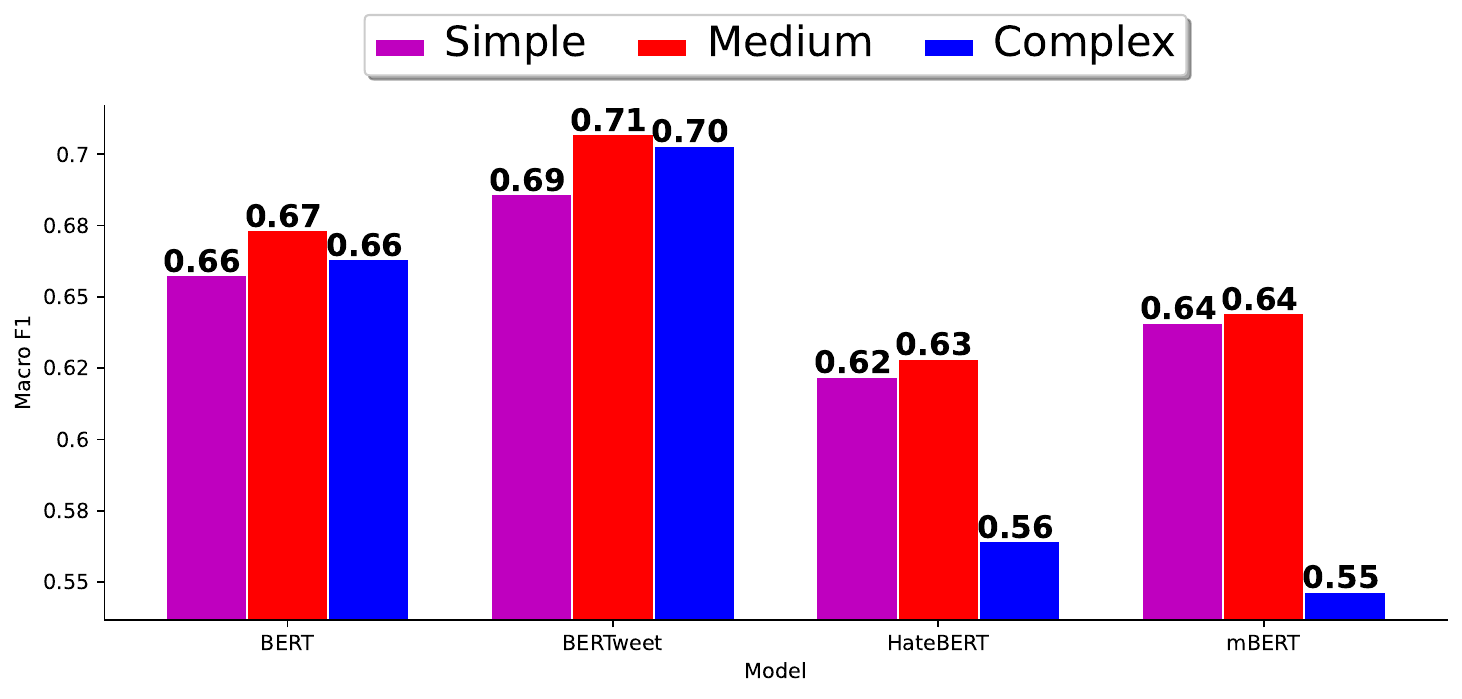}
        \caption{Dynabench}
        \label{fig:cc_Dynabench}
    \end{subfigure}
    \hfill
    \begin{subfigure}[b]{0.49\textwidth}
        \includegraphics[width=\linewidth]{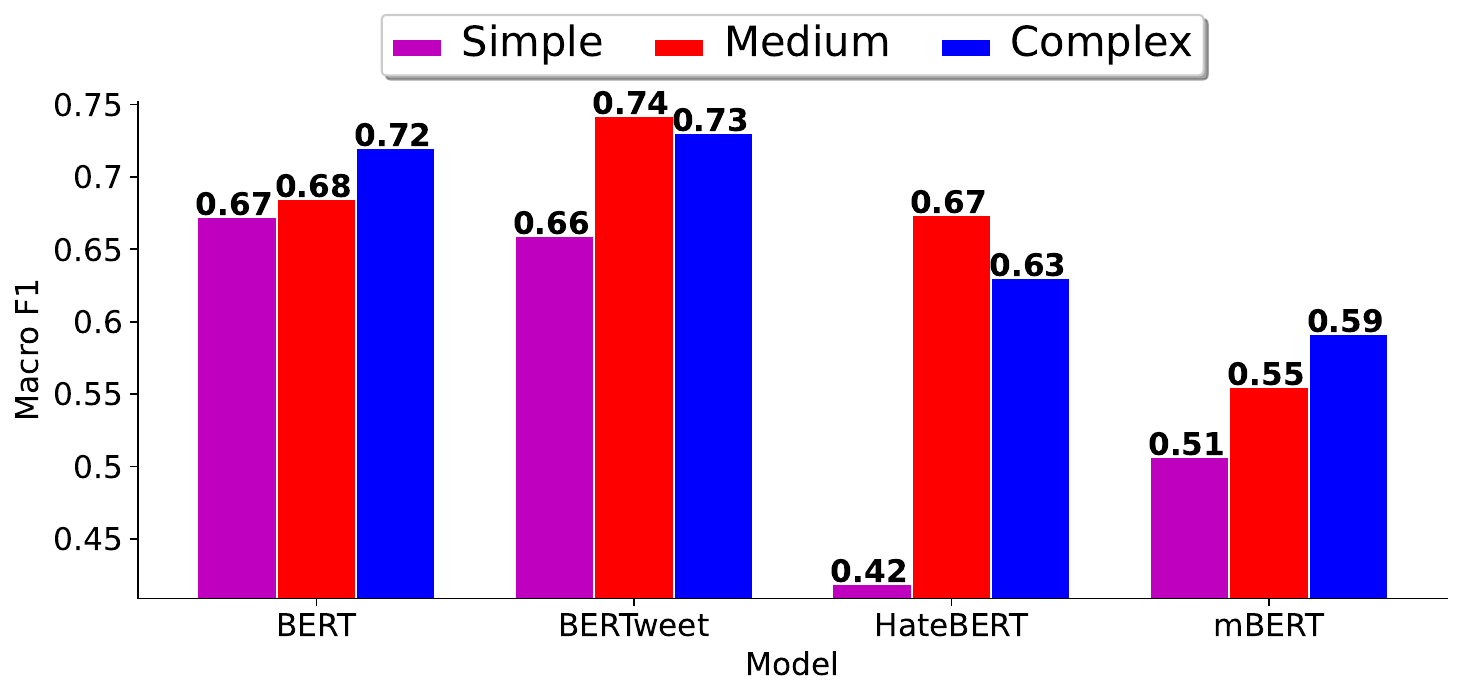}
        \caption{OLID}
        \label{fig:cc_olid_taska}
    \end{subfigure}
    \caption{{\bf RQ5:} Macro F1 scores (averaged over MLP seeds $ms$) for (a) \Di\ and (b) \Dvi\ datasets employing BERT-variants (BERT, BERTweet, HateBERT, and mBERT). Classification heads of varying complexity (simple, medium, and complex) are utilized to capture their effect on BERT-variants employed for hate detection.}
    \label{fig:cc_analysis}
\vspace{-3mm}
\end{figure*}

\textbf{Findings.}
We observe from Figure \ref{fig:cc_analysis} that \textcolor{black}{compared to a simple classification head (CH), a more sophisticated one (either medium or complex) is better.} Full dataset results and analysis are enlisted in Appendix \ref{app:rq5} and reflect similar patterns. Surprisingly, BERTweet, a relatively lesser-used PLM for hate speech detection, outperforms its supposedly superior domain-specific counterpart, HateBERT. Additionally, BERT with a complex classification head demonstrates comparable performance to domain-specific PLMs and even outperforms them in several cases. We also note that mBERT's performance is lost on English-specific datasets. It would be interesting to see how this compares to non-English hate speech datasets that employ mBERT. We further note that HateBERT's performance is highly dependent on the classification head used, with a more complex one often needed to enhance its performance to bring it to part with its coevals. \emph{Interestingly, we observe that a general-purpose pretrained model with a complex classification head may mimic the results of a domain-specific pretrained model. If true for other tasks, it questions the resource allocation for curating domain-specific PLMs.}

\section{Takeaways and Recommendations}
This section summarises the major takeaways that would allow practitioners to make effective choices when modeling PLMs for hate speech detection.
\begin{enumerate}[noitemsep,nolistsep,topsep=0pt,leftmargin=1em]
\item In RQ1, we established that different seed initializations of the classification head and the underlying pretrained model (during its training) could significantly affect PLMs' performance on hate speech detection. However, finding the best-suited hyperparameters is sub-optimal and resource-intensive. \emph{Therefore, we recommend reporting results averaged over more than one seed for the hate detection tasks.}

\item In RQ2, while analyzing the training dynamics of PLMs concerning downstream tasks, we observed early peaks w.r.t hate speech detection. We hypothesize that different NLP tasks may display different peak patterns. \emph{Our first recommendation is to make intermediate checkpoints available if pretraining is involved.} An open research direction is the \emph{intermediate-evaluation test cases to record the PLM's finetuning performance and early stopping if desired thresholds are obtained.} For instance, if we assume the same training setup as used by \citet{elazar2023measuring} and if the training was stopped just after $8$-$10$ epochs noticing the performance drop on the downstream task, $8$-$10\times$ compute, could have been saved. Though their use case differed, this can hold for training models for tasks such as sentiment analysis.

\item In RQ3, we found that pretraining of PLMs on newer data does not help hate speech detection. This is counter-intuitive as one would expect newer data to enhance a model's world knowledge. However, most datasets employed in this study are older than the models being released. Further, the datasets are on the side of explicit hate, and any hateful event regarding them should already be captured in the world knowledge gained by the PLM via the training corpus. Throughout examination in this work, the two synthetically generated datasets, \Di\ and \Dv, do not record any significant deviation from overall trends, even though \Di\ is human-generated while \Dv\ is machine-generated. The only notable difference is that \Di\ is less prone to the complexity of classification heads, as we observe in both RQ2 and RQ5. Whether it is a function of its synthetic nature or large test size is not apparent. \emph{We recommend that benchmark datasets must be regularly updated for subjective tasks like hate speech detection.}  

\textcolor{black}{As the use of generative AI tools for crowdsourcing is on the rise \cite{doi:10.1073/pnas.2305016120, liu2023geval}, it is imperative to equip hate speech researchers to deal with a broader AI-assisted system than just finetuning PLMs. \emph{Moreover, using computational methods at every step of the hate detection pipeline should always be human-aided.}}
 
\item In RQ4, we reinstated the status quo of finetuning the last few layers to obtain the best performance to largely hold for hate detection. Yet, in the case of mBERT, we observed that the middle and lower layers are much more critical. \emph{We recommend that tasks employing multilingual or non-English hate speech detection using mBERT should start with keeping the middle layers unfrozen for finetuning.} By comparing four BERT variants on seven datasets and three seeds, it appears that the region-wise performance of PLMs is a characteristic of the underlying PLM and the task domain at hand and is less impacted by variation in datasets. Such intuitions can help narrow the experiments one has to run to obtain better classification configurations.

\begin{table}[!t]
\resizebox{\columnwidth}{!}{
\begin{tabular}{l|llll}
\hline
\multirow{2}{*}{Test} & \multicolumn{4}{c}{Train}                                                                   \\ \cline{2-5} 
 & \multicolumn{1}{l|}{OLID Min} & \multicolumn{1}{l|}{OLID Max} & \multicolumn{1}{l|}{Dynabench Min} & Dynabench Max \\ \hline
OLID                  & \multicolumn{1}{l|}{0.747} & \multicolumn{1}{l|}{0.817} & \multicolumn{1}{l|}{0.435} & 0.520 \\ \hline
Dynabench             & \multicolumn{1}{l|}{0.435} & \multicolumn{1}{l|}{0.491} & \multicolumn{1}{l|}{0.705} & 0.783 \\ \hline
\end{tabular}}
\caption{\textcolor{black}{\textbf{RQ4:} Macro F1 based on BERTweet cross-dataset generalization. The min and max define the seed+layer combination that led to min and max macro F1 in the in-domain experiments, as reported in Table \ref{tab:layer_wise}. In each row, two columns with the same dataset name as the one in the row correspond to in-domain evaluation, the others correspond to out-of-domain evaluation.}}
\label{tab:cross_dataset}
\vspace{-4mm}
\end{table}

\textcolor{black}{Further, based on the best seed, layer, and PLM combinations obtained in RQ4 (Table \ref{tab:layer_wise}), we randomly picked \Di\ and \Dvi\ to perform a cross-dataset generalization experiment and examine the impact of hyperparameters associated with minimum and maximum in-domain PLM (BERTweet in this case) on cross-domain testing. From Table \ref{tab:cross_dataset}, in line with previous studies \cite{10.1016/j.ipm.2021.102524} on cross-dataset generalization, we observe a poor performance on out-of-domain testing. Our results do hint that the best finetuning setting may also correspond to the best out-of-domain generalization. \emph{Such settings can be useful to narrow down the hyperparameter search in balancing in-domain vs. out-of-domain performance gains.}}

\item In RQ5, we uncovered that finetuning a general-purpose model, like BERT, with a more complex classification head can mimic the performance of a domain-specific pretrained model, like HateBERT. Our analysis also brought out the superiority of BERTweet over HateBERT. While HateBERT is continued-pretrained on offense subreddits, BERTweet is continued-pretrained on Tweets. Given that most datasets are either directly drawn from Twitter or synthesized in a short-text fashion, BERTweet could be indirectly capturing both short-text syntax and offense from the Tweet corpus. \emph{Hence, we recommend practitioners employing HateBERT to report their findings on BERTweet as well.} Further, we observe a slight decrease in performance across datasets comparing mBERT and BERT for English datasets. Given that mBERT has more parameters than BERT (178M vs. 110M in base version), \emph{we suggest not using mBERT unless the hate speech is itself multilingual.} 

When even a random set of test samples can help steal model weights \cite{49091} in NLP tasks, it points to limited domain-specific learning in light of the adversary. Thus, more experiments are needed to establish their superiority over general-purpose models.
\end{enumerate}

\section{Conclusion}
Due to the subjective nature of hate speech, no standard benchmarking exists. We take this opportunity to explore the patterns in finetuning PLMs for hate detection through a series of experiments over five research questions. We hope each experiment in this study lays the ground for future work to improve our understanding of how PLMs model hatefulness and their deployment to detect hate. In the future, we would like to extend our analysis against adversarial settings, bias mitigation, a broader language set, and auto-regressive LLMs.

\newpage
\section{Acknowledgements}
Sarah Masud would like to acknowledge the support of the Prime Minister Doctoral Fellowship and Google PhD Fellowship. The authors also acknowledge the support of our research partner Wipro AI.

\section{Limitations}
Despite examining multiple pretraining and finetuning settings in this study, there are certain limitations that we would like to highlight. First and foremost, the parameters evaluated in this study regarding PLMs, random seeds, and classification heads are not exhaustive due to constraints on computing resources. Secondly, due to BERT and ROBERTA checkpoint variants \cite{sellam2022multiberts,elazar2023measuring} employed in RQ1-RQ3 being available only in English, we were constrained to pick hate speech datasets only in English. While non-English datasets can be utilized to some extent in RQ4 and RQ5, there are again constraints of BERTweet and HateBERT variants being available in those languages. However, results should hold on to other hate speech datasets curated from Twitter. Lastly, we acknowledge that hate speech datasets \cite{madukwe-etal-2020-data} and automatic hate speech detection \cite{schmidt-wiegand-2017-survey}, especially those derived from PLMs, are not without flaws. Blind-sided usage of PLM in hate speech detection can further the stereotypes already present in PLMs \cite{ousidhoum-etal-2021-probing}. 

\section{Ethical Considerations}
Hate speech is a severe issue plaguing society and needs efforts beyond computational methods from different factions of researchers and practitioners. Our aim with this study is not to spread harmful content, nor do we support the hateful content analyzed in this study. In this regard, we hope our experiments help build better and more robust hate speech systems. Further, note that we do not create any new dataset or model in this study and instead employ existing publicly available open-sourced datasets and HuggingFace PLMs in agreement with their data-sharing licenses. The datasets and models are duly cited. Further, given the computationally expensive nature of probing and the carbon footprint incurred, we hope our experiments help narrow the parameter search for future research. During our experimentation, care was taken to inoculate the code against memory leakage, and early stopping, where applicable, was invoked.    

\bibliography{eacl}

\begin{thebibliography}{43}
\expandafter\ifx\csname natexlab\endcsname\relax\def\natexlab#1{#1}\fi

\bibitem[{Antypas and Camacho-Collados(2023)}]{antypas-camacho-collados-2023-robust}
Dimosthenis Antypas and Jose Camacho-Collados. 2023.
\newblock \href {https://doi.org/10.18653/v1/2023.woah-1.25} {Robust hate speech detection in social media: A cross-dataset empirical evaluation}.
\newblock In \emph{The 7th Workshop on Online Abuse and Harms (WOAH)}, pages 231--242, Toronto, Canada. Association for Computational Linguistics.

\bibitem[{Badjatiya et~al.(2017)Badjatiya, Gupta, Gupta, and Varma}]{Badjatiya_2017}
Pinkesh Badjatiya, Shashank Gupta, Manish Gupta, and Vasudeva Varma. 2017.
\newblock \href {https://doi.org/10.1145/3041021.3054223} {Deep learning for hate speech detection in tweets}.
\newblock In \emph{Proceedings of the 26th International Conference on World Wide Web Companion - {WWW} {\textquotesingle}17 Companion}. {ACM} Press.

\bibitem[{Balkir et~al.(2022)Balkir, Nejadgholi, Fraser, and Kiritchenko}]{balkir-etal-2022-necessity}
Esma Balkir, Isar Nejadgholi, Kathleen Fraser, and Svetlana Kiritchenko. 2022.
\newblock \href {https://doi.org/10.18653/v1/2022.naacl-main.192} {Necessity and sufficiency for explaining text classifiers: A case study in hate speech detection}.
\newblock In \emph{Proceedings of the 2022 Conference of the North American Chapter of the Association for Computational Linguistics: Human Language Technologies}, pages 2672--2686, Seattle, United States. Association for Computational Linguistics.

\bibitem[{Biderman et~al.(2023)Biderman, Schoelkopf, Anthony, Bradley, O'Brien, Hallahan, Khan, Purohit, Prashanth, Raff, Skowron, Sutawika, and van~der Wal}]{biderman2023pythia}
Stella Biderman, Hailey Schoelkopf, Quentin Anthony, Herbie Bradley, Kyle O'Brien, Eric Hallahan, Mohammad~Aflah Khan, Shivanshu Purohit, USVSN~Sai Prashanth, Edward Raff, Aviya Skowron, Lintang Sutawika, and Oskar van~der Wal. 2023.
\newblock \href {http://arxiv.org/abs/2304.01373} {Pythia: A suite for analyzing large language models across training and scaling}.

\bibitem[{Caselli et~al.(2021)Caselli, Basile, Mitrovi{\'c}, and Granitzer}]{caselli-etal-2021-hatebert}
Tommaso Caselli, Valerio Basile, Jelena Mitrovi{\'c}, and Michael Granitzer. 2021.
\newblock \href {https://doi.org/10.18653/v1/2021.woah-1.3} {{H}ate{BERT}: Retraining {BERT} for abusive language detection in {E}nglish}.
\newblock In \emph{Proceedings of the 5th Workshop on Online Abuse and Harms (WOAH 2021)}, pages 17--25, Online. Association for Computational Linguistics.

\bibitem[{Davidson et~al.(2017)Davidson, Warmsley, Macy, and Weber}]{hateoffensive}
Thomas Davidson, Dana Warmsley, Michael Macy, and Ingmar Weber. 2017.
\newblock Automated hate speech detection and the problem of offensive language.
\newblock In \emph{Proceedings of the 11th International AAAI Conference on Web and Social Media}, ICWSM '17, pages 512--515.

\bibitem[{de~Vries et~al.(2020)de~Vries, van Cranenburgh, and Nissim}]{de-vries-etal-2020-whats}
Wietse de~Vries, Andreas van Cranenburgh, and Malvina Nissim. 2020.
\newblock \href {https://doi.org/10.18653/v1/2020.findings-emnlp.389} {What{'}s so special about {BERT}{'}s layers? a closer look at the {NLP} pipeline in monolingual and multilingual models}.
\newblock In \emph{Findings of the Association for Computational Linguistics: EMNLP 2020}, pages 4339--4350, Online. Association for Computational Linguistics.

\bibitem[{Devlin et~al.(2019)Devlin, Chang, Lee, and Toutanova}]{devlin-etal-2019-bert}
Jacob Devlin, Ming-Wei Chang, Kenton Lee, and Kristina Toutanova. 2019.
\newblock \href {https://doi.org/10.18653/v1/N19-1423} {{BERT}: Pre-training of deep bidirectional transformers for language understanding}.
\newblock In \emph{Proceedings of the 2019 Conference of the North {A}merican Chapter of the Association for Computational Linguistics: Human Language Technologies, Volume 1 (Long and Short Papers)}, pages 4171--4186, Minneapolis, Minnesota. Association for Computational Linguistics.

\bibitem[{Durrani et~al.(2022)Durrani, Sajjad, Dalvi, and Alam}]{durrani-etal-2022-transformation}
Nadir Durrani, Hassan Sajjad, Fahim Dalvi, and Firoj Alam. 2022.
\newblock \href {https://aclanthology.org/2022.emnlp-main.97} {On the transformation of latent space in fine-tuned {NLP} models}.
\newblock In \emph{Proceedings of the 2022 Conference on Empirical Methods in Natural Language Processing}, pages 1495--1516, Abu Dhabi, United Arab Emirates. Association for Computational Linguistics.

\bibitem[{Elazar et~al.(2023)Elazar, Kassner, Ravfogel, Feder, Ravichander, Mosbach, Belinkov, Schütze, and Goldberg}]{elazar2023measuring}
Yanai Elazar, Nora Kassner, Shauli Ravfogel, Amir Feder, Abhilasha Ravichander, Marius Mosbach, Yonatan Belinkov, Hinrich Schütze, and Yoav Goldberg. 2023.
\newblock \href {http://arxiv.org/abs/2207.14251} {Measuring causal effects of data statistics on language model's `factual' predictions}.

\bibitem[{Elhage et~al.(2021)Elhage, Nanda, Olsson, Henighan, Joseph, Mann, Askell, Bai, Chen, Conerly, DasSarma, Drain, Ganguli, Hatfield-Dodds, Hernandez, Jones, Kernion, Lovitt, Ndousse, Amodei, Brown, Clark, Kaplan, McCandlish, and Olah}]{elhage2021mathematical}
Nelson Elhage, Neel Nanda, Catherine Olsson, Tom Henighan, Nicholas Joseph, Ben Mann, Amanda Askell, Yuntao Bai, Anna Chen, Tom Conerly, Nova DasSarma, Dawn Drain, Deep Ganguli, Zac Hatfield-Dodds, Danny Hernandez, Andy Jones, Jackson Kernion, Liane Lovitt, Kamal Ndousse, Dario Amodei, Tom Brown, Jack Clark, Jared Kaplan, Sam McCandlish, and Chris Olah. 2021.
\newblock A mathematical framework for transformer circuits.
\newblock \emph{Transformer Circuits Thread}.
\newblock Https://transformer-circuits.pub/2021/framework/index.html.

\bibitem[{Fortuna et~al.(2021)Fortuna, Soler-Company, and Wanner}]{10.1016/j.ipm.2021.102524}
Paula Fortuna, Juan Soler-Company, and Leo Wanner. 2021.
\newblock \href {https://doi.org/10.1016/j.ipm.2021.102524} {How well do hate speech, toxicity, abusive and offensive language classification models generalize across datasets?}
\newblock \emph{Inf. Process. Manage.}, 58(3).

\bibitem[{Founta et~al.(2018)Founta, Djouvas, Chatzakou, Leontiadis, Blackburn, Stringhini, Vakali, Sirivianos, and Kourtellis}]{Founta_Djouvas_Chatzakou_Leontiadis_Blackburn_Stringhini_Vakali_Sirivianos_Kourtellis_2018}
Antigoni Founta, Constantinos Djouvas, Despoina Chatzakou, Ilias Leontiadis, Jeremy Blackburn, Gianluca Stringhini, Athena Vakali, Michael Sirivianos, and Nicolas Kourtellis. 2018.
\newblock \href {https://doi.org/10.1609/icwsm.v12i1.14991} {Large scale crowdsourcing and characterization of twitter abusive behavior}.
\newblock \emph{Proceedings of the International AAAI Conference on Web and Social Media}, 12(1).

\bibitem[{Founta et~al.(2019)Founta, Chatzakou, Kourtellis, Blackburn, Vakali, and Leontiadis}]{10.1145/3292522.3326028}
Antigoni~Maria Founta, Despoina Chatzakou, Nicolas Kourtellis, Jeremy Blackburn, Athena Vakali, and Ilias Leontiadis. 2019.
\newblock \href {https://doi.org/10.1145/3292522.3326028} {A unified deep learning architecture for abuse detection}.
\newblock In \emph{Proceedings of the 10th ACM Conference on Web Science}, WebSci '19, page 105–114, New York, NY, USA. Association for Computing Machinery.

\bibitem[{Geva et~al.(2022)Geva, Caciularu, Dar, Roit, Sadde, Shlain, Tamir, and Goldberg}]{geva-etal-2022-lm}
Mor Geva, Avi Caciularu, Guy Dar, Paul Roit, Shoval Sadde, Micah Shlain, Bar Tamir, and Yoav Goldberg. 2022.
\newblock \href {https://aclanthology.org/2022.emnlp-demos.2} {{LM}-debugger: An interactive tool for inspection and intervention in transformer-based language models}.
\newblock In \emph{Proceedings of the 2022 Conference on Empirical Methods in Natural Language Processing: System Demonstrations}, pages 12--21, Abu Dhabi, UAE. Association for Computational Linguistics.

\bibitem[{Gilardi et~al.(2023)Gilardi, Alizadeh, and Kubli}]{doi:10.1073/pnas.2305016120}
Fabrizio Gilardi, Meysam Alizadeh, and Maël Kubli. 2023.
\newblock \href {https://doi.org/10.1073/pnas.2305016120} {Chatgpt outperforms crowd workers for text-annotation tasks}.
\newblock \emph{Proceedings of the National Academy of Sciences}, 120(30):e2305016120.

\bibitem[{Hartvigsen et~al.(2022)Hartvigsen, Gabriel, Palangi, Sap, Ray, and Kamar}]{hartvigsen-etal-2022-toxigen}
Thomas Hartvigsen, Saadia Gabriel, Hamid Palangi, Maarten Sap, Dipankar Ray, and Ece Kamar. 2022.
\newblock \href {https://doi.org/10.18653/v1/2022.acl-long.234} {{T}oxi{G}en: A large-scale machine-generated dataset for adversarial and implicit hate speech detection}.
\newblock In \emph{Proceedings of the 60th Annual Meeting of the Association for Computational Linguistics (Volume 1: Long Papers)}, pages 3309--3326, Dublin, Ireland. Association for Computational Linguistics.

\bibitem[{Ilan and Vilenchik(2022)}]{ilan-vilenchik-2022-harald}
Tal Ilan and Dan Vilenchik. 2022.
\newblock \href {https://aclanthology.org/2022.findings-emnlp.165} {{HARALD}: Augmenting hate speech data sets with real data}.
\newblock In \emph{Findings of the Association for Computational Linguistics: EMNLP 2022}, pages 2241--2248, Abu Dhabi, United Arab Emirates. Association for Computational Linguistics.

\bibitem[{Khan et~al.(2023)Khan, Yadav, Jain, and Goyal}]{DBLP:conf/iclr/KhanYJG23}
Mohammad~Aflah Khan, Neemesh Yadav, Mohit Jain, and Sanyam Goyal. 2023.
\newblock \href {https://openreview.net/pdf?id=1yXbt6\_o6av} {The art of embedding fusion: Optimizing hate speech detection}.
\newblock In \emph{The First Tiny Papers Track at {ICLR} 2023, Tiny Papers @ {ICLR} 2023, Kigali, Rwanda, May 5, 2023}. OpenReview.net.

\bibitem[{Krishna et~al.(2020)Krishna, Tomar, Parikh, Papernot, and Iyyer}]{49091}
Kalpesh Krishna, Gaurav~Singh Tomar, Ankur Parikh, Nicolas Papernot, and Mohit Iyyer. 2020.
\newblock Thieves of sesame street: Model extraction on bert-based apis.

\bibitem[{Kulkarni et~al.(2023)Kulkarni, Masud, Goyal, and Chakraborty}]{10.1145/3580305.3599896}
Atharva Kulkarni, Sarah Masud, Vikram Goyal, and Tanmoy Chakraborty. 2023.
\newblock \href {https://doi.org/10.1145/3580305.3599896} {Revisiting hate speech benchmarks: From data curation to system deployment}.
\newblock In \emph{Proceedings of the 29th ACM SIGKDD Conference on Knowledge Discovery and Data Mining}, KDD '23, page 4333–4345, New York, NY, USA. Association for Computing Machinery.

\bibitem[{Liu et~al.(2023)Liu, Iter, Xu, Wang, Xu, and Zhu}]{liu2023geval}
Yang Liu, Dan Iter, Yichong Xu, Shuohang Wang, Ruochen Xu, and Chenguang Zhu. 2023.
\newblock \href {http://arxiv.org/abs/2303.16634} {G-eval: Nlg evaluation using gpt-4 with better human alignment}.

\bibitem[{Liu et~al.(2019)Liu, Ott, Goyal, Du, Joshi, Chen, Levy, Lewis, Zettlemoyer, and Stoyanov}]{DBLP:journals/corr/abs-1907-11692}
Yinhan Liu, Myle Ott, Naman Goyal, Jingfei Du, Mandar Joshi, Danqi Chen, Omer Levy, Mike Lewis, Luke Zettlemoyer, and Veselin Stoyanov. 2019.
\newblock \href {http://arxiv.org/abs/1907.11692} {Roberta: {A} robustly optimized {BERT} pretraining approach}.
\newblock \emph{CoRR}, abs/1907.11692.

\bibitem[{Loshchilov and Hutter(2019)}]{DBLP:conf/iclr/LoshchilovH19}
Ilya Loshchilov and Frank Hutter. 2019.
\newblock \href {https://openreview.net/forum?id=Bkg6RiCqY7} {Decoupled weight decay regularization}.
\newblock In \emph{7th International Conference on Learning Representations, {ICLR} 2019, New Orleans, LA, USA, May 6-9, 2019}. OpenReview.net.

\bibitem[{Lundberg and Lee(2017)}]{10.5555/3295222.3295230}
Scott~M. Lundberg and Su-In Lee. 2017.
\newblock A unified approach to interpreting model predictions.
\newblock In \emph{Proceedings of the 31st International Conference on Neural Information Processing Systems}, NIPS'17, page 4768–4777, Red Hook, NY, USA. Curran Associates Inc.

\bibitem[{Madukwe et~al.(2020)Madukwe, Gao, and Xue}]{madukwe-etal-2020-data}
Kosisochukwu Madukwe, Xiaoying Gao, and Bing Xue. 2020.
\newblock \href {https://doi.org/10.18653/v1/2020.alw-1.18} {In data we trust: A critical analysis of hate speech detection datasets}.
\newblock In \emph{Proceedings of the Fourth Workshop on Online Abuse and Harms}, pages 150--161, Online. Association for Computational Linguistics.

\bibitem[{Masud et~al.(2022)Masud, Bedi, Khan, Akhtar, and Chakraborty}]{10.1145/3534678.3539161}
Sarah Masud, Manjot Bedi, Mohammad~Aflah Khan, Md~Shad Akhtar, and Tanmoy Chakraborty. 2022.
\newblock \href {https://doi.org/10.1145/3534678.3539161} {Proactively reducing the hate intensity of online posts via hate speech normalization}.
\newblock In \emph{Proceedings of the 28th ACM SIGKDD Conference on Knowledge Discovery and Data Mining}, KDD '22, page 3524–3534, New York, NY, USA. Association for Computing Machinery.

\bibitem[{Mathew et~al.(2021)Mathew, Saha, Yimam, Biemann, Goyal, and Mukherjee}]{mathew2021hatexplain}
Binny Mathew, Punyajoy Saha, Seid~Muhie Yimam, Chris Biemann, Pawan Goyal, and Animesh Mukherjee. 2021.
\newblock Hatexplain: A benchmark dataset for explainable hate speech detection.
\newblock In \emph{Proceedings of the AAAI Conference on Artificial Intelligence}, volume~35, pages 14867--14875.

\bibitem[{Nguyen et~al.(2020)Nguyen, Vu, and Tuan~Nguyen}]{nguyen-etal-2020-bertweet}
Dat~Quoc Nguyen, Thanh Vu, and Anh Tuan~Nguyen. 2020.
\newblock \href {https://doi.org/10.18653/v1/2020.emnlp-demos.2} {{BERT}weet: A pre-trained language model for {E}nglish tweets}.
\newblock In \emph{Proceedings of the 2020 Conference on Empirical Methods in Natural Language Processing: System Demonstrations}, pages 9--14, Online. Association for Computational Linguistics.

\bibitem[{Ousidhoum et~al.(2021)Ousidhoum, Zhao, Fang, Song, and Yeung}]{ousidhoum-etal-2021-probing}
Nedjma Ousidhoum, Xinran Zhao, Tianqing Fang, Yangqiu Song, and Dit-Yan Yeung. 2021.
\newblock \href {https://doi.org/10.18653/v1/2021.acl-long.329} {Probing toxic content in large pre-trained language models}.
\newblock In \emph{Proceedings of the 59th Annual Meeting of the Association for Computational Linguistics and the 11th International Joint Conference on Natural Language Processing (Volume 1: Long Papers)}, pages 4262--4274, Online. Association for Computational Linguistics.

\bibitem[{Ribeiro et~al.(2016)Ribeiro, Singh, and Guestrin}]{10.1145/2939672.2939778}
Marco~Tulio Ribeiro, Sameer Singh, and Carlos Guestrin. 2016.
\newblock \href {https://doi.org/10.1145/2939672.2939778} {{"Why Should I Trust You?": Explaining the Predictions of Any Classifier}}.
\newblock In \emph{Proceedings of the 22nd ACM SIGKDD International Conference on Knowledge Discovery and Data Mining}, KDD '16, page 1135–1144, New York, NY, USA. Association for Computing Machinery.

\bibitem[{Schmidt and Wiegand(2017)}]{schmidt-wiegand-2017-survey}
Anna Schmidt and Michael Wiegand. 2017.
\newblock \href {https://doi.org/10.18653/v1/W17-1101} {A survey on hate speech detection using natural language processing}.
\newblock In \emph{Proceedings of the Fifth International Workshop on Natural Language Processing for Social Media}, pages 1--10, Valencia, Spain. Association for Computational Linguistics.

\bibitem[{Sellam et~al.(2022)Sellam, Yadlowsky, Tenney, Wei, Saphra, D'Amour, Linzen, Bastings, Turc, Eisenstein, Das, and Pavlick}]{sellam2022multiberts}
Thibault Sellam, Steve Yadlowsky, Ian Tenney, Jason Wei, Naomi Saphra, Alexander D'Amour, Tal Linzen, Jasmijn Bastings, Iulia~Raluca Turc, Jacob Eisenstein, Dipanjan Das, and Ellie Pavlick. 2022.
\newblock \href {https://openreview.net/forum?id=K0E_F0gFDgA} {The multi{BERT}s: {BERT} reproductions for robustness analysis}.
\newblock In \emph{International Conference on Learning Representations}.

\bibitem[{Sun et~al.(2019)Sun, Qiu, Xu, and Huang}]{ccl}
Chi Sun, Xipeng Qiu, Yige Xu, and Xuanjing Huang. 2019.
\newblock How to fine-tune bert for text classification?
\newblock In \emph{Chinese Computational Linguistics}, pages 194--206, Cham. Springer International Publishing.

\bibitem[{van Aken et~al.(2019)van Aken, Winter, L\"{o}ser, and Gers}]{10.1145/3357384.3358028}
Betty van Aken, Benjamin Winter, Alexander L\"{o}ser, and Felix~A. Gers. 2019.
\newblock \href {https://doi.org/10.1145/3357384.3358028} {How does bert answer questions? a layer-wise analysis of transformer representations}.
\newblock In \emph{Proceedings of the 28th ACM International Conference on Information and Knowledge Management}, CIKM '19, page 1823–1832, New York, NY, USA. Association for Computing Machinery.

\bibitem[{Vaswani et~al.(2017)Vaswani, Shazeer, Parmar, Uszkoreit, Jones, Gomez, Kaiser, and Polosukhin}]{NIPS2017_3f5ee243}
Ashish Vaswani, Noam Shazeer, Niki Parmar, Jakob Uszkoreit, Llion Jones, Aidan~N Gomez, \L~ukasz Kaiser, and Illia Polosukhin. 2017.
\newblock \href {https://proceedings.neurips.cc/paper_files/paper/2017/file/3f5ee243547dee91fbd053c1c4a845aa-Paper.pdf} {Attention is all you need}.
\newblock In \emph{Advances in Neural Information Processing Systems}, volume~30. Curran Associates, Inc.

\bibitem[{Vidgen et~al.(2021)Vidgen, Thrush, Waseem, and Kiela}]{vidgen-etal-2021-learning}
Bertie Vidgen, Tristan Thrush, Zeerak Waseem, and Douwe Kiela. 2021.
\newblock \href {https://doi.org/10.18653/v1/2021.acl-long.132} {Learning from the worst: Dynamically generated datasets to improve online hate detection}.
\newblock In \emph{Proceedings of the 59th Annual Meeting of the Association for Computational Linguistics and the 11th International Joint Conference on Natural Language Processing (Volume 1: Long Papers)}, pages 1667--1682, Online. Association for Computational Linguistics.

\bibitem[{Vijayaraghavan et~al.(2021)Vijayaraghavan, Larochelle, and Roy}]{vijayaraghavan2021interpretable}
Prashanth Vijayaraghavan, Hugo Larochelle, and Deb Roy. 2021.
\newblock \href {http://arxiv.org/abs/2103.01616} {Interpretable multi-modal hate speech detection}.

\bibitem[{Wang et~al.(2018)Wang, Singh, Michael, Hill, Levy, and Bowman}]{wang-etal-2018-glue}
Alex Wang, Amanpreet Singh, Julian Michael, Felix Hill, Omer Levy, and Samuel Bowman. 2018.
\newblock \href {https://doi.org/10.18653/v1/W18-5446} {{GLUE}: A multi-task benchmark and analysis platform for natural language understanding}.
\newblock In \emph{Proceedings of the 2018 {EMNLP} Workshop {B}lackbox{NLP}: Analyzing and Interpreting Neural Networks for {NLP}}, pages 353--355, Brussels, Belgium. Association for Computational Linguistics.

\bibitem[{Waseem and Hovy(2016)}]{waseem-hovy-2016-hateful}
Zeerak Waseem and Dirk Hovy. 2016.
\newblock \href {https://doi.org/10.18653/v1/N16-2013} {Hateful symbols or hateful people? predictive features for hate speech detection on {T}witter}.
\newblock In \emph{Proceedings of the {NAACL} Student Research Workshop}, pages 88--93, San Diego, California. Association for Computational Linguistics.

\bibitem[{Wolf et~al.(2020)Wolf, Debut, Sanh, Chaumond, Delangue, Moi, Cistac, Rault, Louf, Funtowicz, Davison, Shleifer, von Platen, Ma, Jernite, Plu, Xu, Le~Scao, Gugger, Drame, Lhoest, and Rush}]{wolf-etal-2020-transformers}
Thomas Wolf, Lysandre Debut, Victor Sanh, Julien Chaumond, Clement Delangue, Anthony Moi, Pierric Cistac, Tim Rault, Remi Louf, Morgan Funtowicz, Joe Davison, Sam Shleifer, Patrick von Platen, Clara Ma, Yacine Jernite, Julien Plu, Canwen Xu, Teven Le~Scao, Sylvain Gugger, Mariama Drame, Quentin Lhoest, and Alexander Rush. 2020.
\newblock \href {https://doi.org/10.18653/v1/2020.emnlp-demos.6} {Transformers: State-of-the-art natural language processing}.
\newblock In \emph{Proceedings of the 2020 Conference on Empirical Methods in Natural Language Processing: System Demonstrations}, pages 38--45, Online. Association for Computational Linguistics.

\bibitem[{Zampieri et~al.(2019)Zampieri, Malmasi, Nakov, Rosenthal, Farra, and Kumar}]{zampieri-etal-2019-semeval}
Marcos Zampieri, Shervin Malmasi, Preslav Nakov, Sara Rosenthal, Noura Farra, and Ritesh Kumar. 2019.
\newblock \href {https://doi.org/10.18653/v1/S19-2010} {{S}em{E}val-2019 task 6: Identifying and categorizing offensive language in social media ({O}ffens{E}val)}.
\newblock In \emph{Proceedings of the 13th International Workshop on Semantic Evaluation}, pages 75--86, Minneapolis, Minnesota, USA. Association for Computational Linguistics.

\bibitem[{Zhu et~al.(2023)Zhu, Liu, Dong, Xu, Huang, Kong, Chen, and Li}]{zhu2023multilingual}
Wenhao Zhu, Hongyi Liu, Qingxiu Dong, Jingjing Xu, Shujian Huang, Lingpeng Kong, Jiajun Chen, and Lei Li. 2023.
\newblock \href {http://arxiv.org/abs/2304.04675} {Multilingual machine translation with large language models: Empirical results and analysis}.

\end{thebibliography}
\clearpage
\newpage
\appendix
\section{Appendix}

\subsection{RQ1: Extended Experiments}
\label{app:rq1}
Table \ref{tab:multibert_ms} and Table \ref{tab:multibert_ps} provide a seed-wise breakdown comparing minimum and maximum macro F1 scores when employing the multiple-checkpoints BERT \cite{sellam2022multiberts} model. In Table \ref{tab:multibert_ms}, the MLP seed ($ms$) is constant, but the pretraining seed ($ps$) varies and vice-versa in Table \ref{tab:multibert_ps}. It appears that keeping $ms$ constant leads to more variability in performance than $ps$.

\subsection{RQ2: Extended Experiments}
\label{app:rq2}
\begin{figure*}[!t]
  \includegraphics[width=0.95\textwidth]{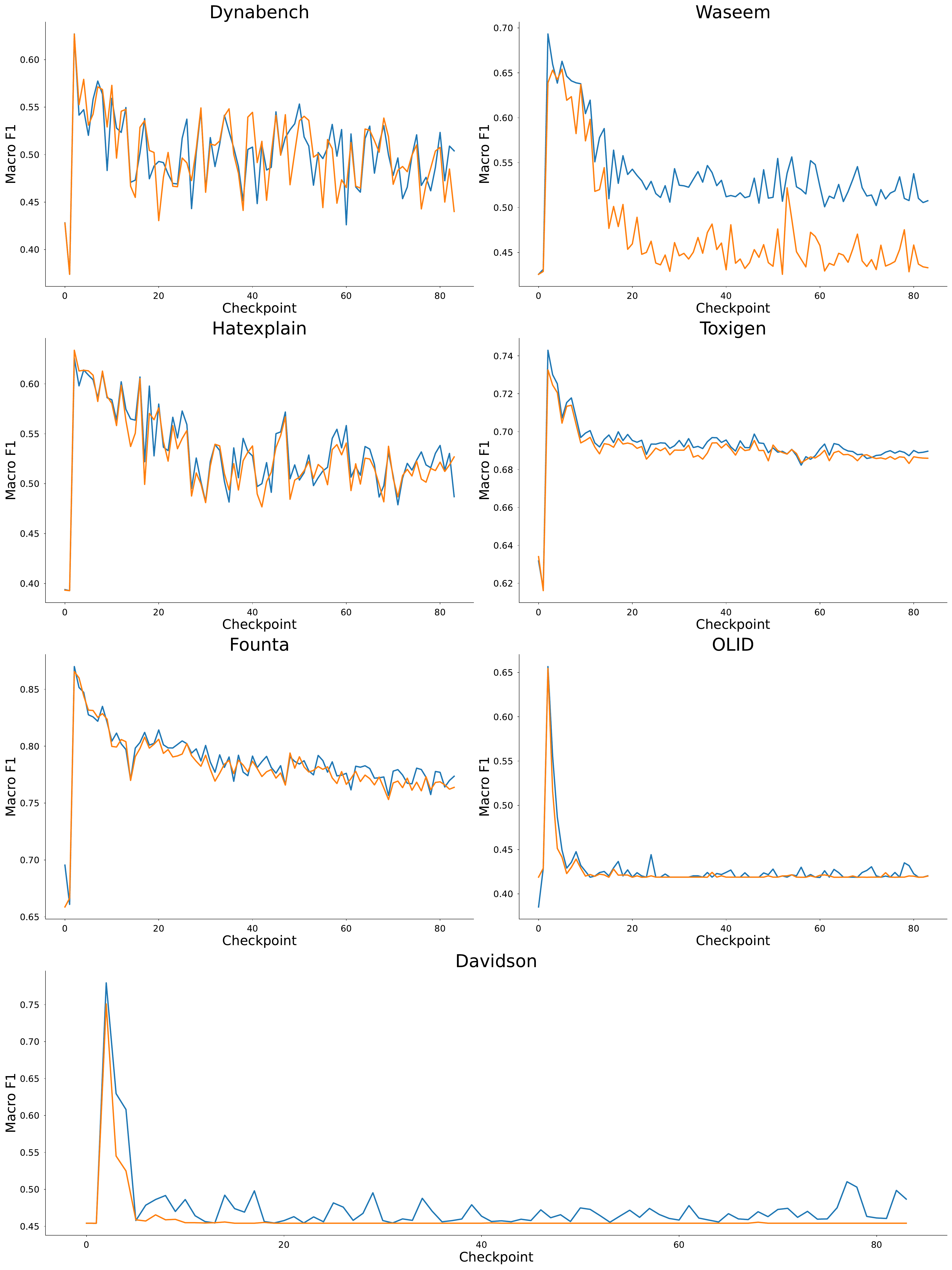}
    \caption{\textbf{RQ2:} Macro F1 (averaged over MLP seeds $ms$) attained when finetuning is done on the $n^{th} \in 1,\cdots,84$ checkpoint ($C_n$). We report the trends on all datasets for simple (yellow) and complex (blue) classification heads. Performance peaks with early checkpoints around $C_n$ are clearly visible for all configurations.}
  \label{fig:roberta_all_intermediate}
\vspace{-3mm}
\end{figure*}

In Figure \ref{fig:roberta_all_intermediate}, we showcase the trends for macro F1 on each dataset when the underlying model is picked from one of the $84$ (x-axis) intermediate checkpoints \cite{elazar2023measuring}. While simple and complex classification heads follow the same pattern overall, a significant difference in maximum macro F1 is obtained at each checkpoint (comparing simple and complex). The same is recorded in Table \ref{tab:robert_intermediate_ptest}. On the one hand, we observe that \Dvi\ and \Di\ have similar performances irrespective of the CH. On the other hand, \Di\ is a relatively new human-synthesized and much larger compared to \Dvi\ ($10k$ vs. $800$), which is obtained from Twitter. Further, we observe that for $5$ datasets, there is a significant improvement in macro F1 score when employing complex CH instead of simple. In RQ5, we also study this CH's effect on other PLM variants. 

\begin{table*}[t]
\centering
\resizebox{0.85\textwidth}{!}{
\begin{tabular}{l|lll|lll|lll}
\hline
\multirow{2}{*}{Dataset} & 12     &       &    & 127    &        &    & 451     &        & \\ \cline{2-10}
& Min F1 & Max F1 & ES & Min F1 & Max F1 & ES & Min F1 & Max F1 & ES \\ \hline
                    \Dii
                     & $S_{0}$: 0.676 & $S_{10}$: 0.731 & 0.426*
                     & $S_{5}$: 0.709 & $S_{15}$: 0.726 & 0.131
                     & $S_{0}$: 0.675 & $S_{10}$: 0.723 & 0.390* \\
                     \Dvii
                     & $S_{20}$: 0.759 & $S_{15}$: 0.791 & 0.441**
                     & $S_{10}$: 0.755 & $S_{20}$: 0.776 & 0.273*
                     & $S_{0}$: 0.745 & $S_{15}$: 0.786 & 0.491** \\
                    \Div
                     & $S_{5}$: 0.872 & $S_{10}$: 0.886 & 0.402*
                     & $S_{5}$: 0.876 & $S_{20}$: 0.888 & 0.356*
                     & $S_{0}$: 0.874 & $S_{0}$: 0.885 & 0.360* \\
                     \Dvi
                     & $S_{20}$: 0.672 & $S_{10}$: 0.718 & 0.207
                     & $S_{0}$: 0.675 & $S_{15}$: 0.725 & 0.169 
                     & $S_{0}$: 0.647 & $S_{10}$: 0.731 & 0.287* \\
                    \Diii 
                     & $S_{20}$: 0.634 & $S_{15}$: 0.679 & 0.687**
                     & $S_{5}$: 0.630 & $S_{20}$: 0.674 & 0.637**
                     & $S_{5}$: 0.636 & $S_{10}$: 0.680 & 0.588** \\
                     \Di 
                     & $S_{5}$: 0.653 & $S_{20}$: 0.660 & 0.153
                     & $S_{5}$: 0.637 & $S_{15}$: 0.659 & 0.468**
                     & $S_{15}$: 0.623 & $S_{20}$: 0.654 :& 0.600** \\
                    \Dv
                     & $S_{20}$: 0.767 & $S_{10}$: 0.771 & 0.180 
                     & $S_{5}$: 0.767 & $S_{10}$: 0.771 & 0.218
                     & $S_{5}$: 0.767 & $S_{10}$: 0.771 & 0.228 \\
                     
                     \hline    
\end{tabular}}
\caption{\textbf{RQ1:} Comparison of minimum and maximum macro F1 obtained when the MLP seed ($ms$) is constant but the pretraining seed varies ($ps$). ES stands for effect size. ** and * indicates whether the difference in minimum and maximum macro F1 is significant by $\le0.05$ and $\le0.001$ p-value, respectively.}
\label{tab:multibert_ms}
\end{table*}

\begin{table*}[!h]
\resizebox{\textwidth}{!}{
\begin{tabular}{l|lll|lll|lll|lll|lll}
\hline
\multirow{2}{*}{Dataset} & 0      &       &    & 5      &        &    & 10     &        &    & 15  &     & &20 & & \\ \cline{2-16}
         & Min F1 & Max F1 & ES & Min F1 & Max F1 & ES & Min F1 & Max F1 & ES & Min F1 & Max F1 & ES & Min F1 & Max F1 & ES   \\ \hline
        \Dii
         & $S_{451}$: 0.675 & $S_{127}$: 0.709 &  0.261
         & $S_{12}$: 0.691 & $S_{127}$: 0.709 &  0.126
         & $S_{127}$: 0.714 & $S_{12}$: 0.731 &  0.142
         & $S_{12}$: 0.711 & $S_{127}$: 0.726 &  0.123
         & $S_{12}$: 0.686 & $S_{127}$: 0.714 &  0.217 \\
        \Dvii
         & $S_{451}$: 0.745 & $S_{127}$: 0766. &  0.232
         & $S_{127}$: 0.757 & $S_{12}$: 0.763 &  0.090
         & $S_{127}$: 0.755 & $S_{12}$: 0.772 &  0.221
         & $S_{127}$: 0.757 & $S_{12}$: 0.791 &  0.435*
         & $S_{451}$: 0.755 & $S_{127}$: 0.776 &  0.291* \\
         \Div
         & $S_{12}$: 0.879 & $S_{451}$: 0.885 & 0.204
         & $S_{12}$: 0.872 & $S_{127}$: 0.876 & 0.123
         & $S_{451}$: 0.884 & $S_{127}$: 0.887 & 0.093
         & $S_{12}$: 0.885 & $S_{127}$: 0.887 & 0.087
         & $S_{12}$: 0.884 & $S_{127}$: 0.888 & 0.121 \\
         \Dvi
         & $S_{451}$: 0.647 & $S_{127}$: 0.675 & 0.089
         & $S_{451}$: 0.661 & $S_{12}$: 0.689 & 0.106
         & $S_{12}$: 0.718 & $S_{451}$: 0.731 & 0.056
         & $S_{451}$: 0.692 & $S_{127}$: 0.725 & 0.141
         & $S_{12}$: 0.672 & $S_{451}$: 0.703 & 0.113 \\
         \Diii
         & $S_{127}$: 0.658 & $S_{12}$: 0.674 &  0.215
         & $S_{127}$: 0.630 & $S_{12}$: 0.6664 &  0.483**
         & $S_{127}$: 0.640 & $S_{451}$: 0.680 & 0.504**
         & $S_{127}$: 0.660 & $S_{12}$: 0.679 & 0.300*
         & $S_{12}$: 0.634 & $S_{127}$: 0.674 & 0.591** \\
         
         \Di & $S_{451}$: 0.648 & $S_{127}$: 0.656       & 0.181    
         & $S_{127}$: 0.637 & $S_{12}$: 0.653 &  0.347*
         & $S_{451}$: 0.654 & $S_{127}$: 0.657 & 0.06
         & $S_{451}$: 0.625 & $S_{127}$: 0.659 & 0.701**
         & $S_{127}$: 0.634& $S_{12}$: 0.660 & 0.142 \\
        \Dv
         & $S_{12}$: 0.769 & $S_{127}$: 0.769 & 0.034
         & $S_{451}$: 0.767 & $S_{12}$: 0.768 & 0.075
         & $S_{12}$: 0.771 & $S_{127}$: 0.771 & 0.050
         & $S_{127}$: 0.770 & $S_{12}$: 0.770 & 0.032
         & $S_{12}$: 0.767 & $S_{127}$: 0.768 & 0.059 \\
 \hline    
\end{tabular}}
\caption{\textbf{RQ1: }Comparison of minimum and maximum macro F1 obtained when the pretraining seed (ps) is constant but the MLP seed (ms) varies. ES stands for effect size. ** and * indicate whether the difference in minimum and maximum macro F1 is significant by $\le0.05$ and $\le0.001$ p-value, respectively.}
\label{tab:multibert_ps}
\end{table*}

\begin{table*}[!h]
\resizebox{\textwidth}{!}{
\begin{tabular}{l|lll|lll|lll}
\hline
\multirow{2}{*}{Dataset} & \multicolumn{3}{l}{12}     & \multicolumn{3}{l}{127}   & \multicolumn{3}{l}{451}    \\ \cline{2-10}
                         & Sim. F1 & Com. F1 & ES & Sim. Max F1 & Com. F1 & ES & Sim. Max F1 & Com. F1 & ES \\ \hline
\Dii & $C_3$: 0.660  & $C_2$: 0.734  & 0.581**  & $C_3$:0.668  & $C_2$:0.738  & 0.547**  & $C_2$: 0.691  & $C_2$:0.775  &0.580**\\
\Dvii & $C_2$: 0.739  & $C_2$: 0.824  & 0.953**  & $C_2$:0.740  & $C_3$:0.810  & 0.852**  & $C_2$: 0.775  & $C_2$:0.764  &0.113 \\
\Div & $C_3$: 0.871  & $C_2$: 0.879  & 0.278*  & $C_2$:0.861 & $C_2$:0.880  & 0.613**  & $C_3$: 0.869  & $C_2$:0.878  &0.269 \\
\Dvi & $C_2$: 0.661  & $C_2$: 0.667  & 0.110  & $C_2$:0.649 & $C_2$:0.694  & 0.242  & $C_2$: 0.654  & $C_2$:0.672  &0.164 \\
\Diii & $C_2$: 0.640  & $C_2$: 0.687  & 0.599**  & $C_2$:0.659 & $C_2$:0.665  & 0.088  & $C_4$: 0.640  & $C_2$:0.694  &0.751** \\
\Di & $C_2$: 0.626  & $C_2$: 0.628  & 0.010  & $C_2$:0.629 & $C_2$:0.623  & 0.123  & $C_2$: 0.625  & $C_2$:0.631  &0.118 \\
\Dv & $C_2$: 0.733  & $C_2$: 0.764 & 1.810**  & $C_2$:0.732 & $C_2$:0.763  & 1.772**  & $C_2$: 0.733 & $C_2$:0.764  & 1.835** \\ \hline
 
\end{tabular}
}
\caption{\textbf{RQ2:} Comparison of maximum macro F1 obtained under varying MLP seed ($ms$) for the simple (Sim.) and complex (Com.) classification heads. ES stands for effect size. ** and * indicates whether the difference in maximum macro F1 is significant by $\le0.05$ and $\le0.001$ p-value, respectively.}
\label{tab:robert_intermediate_ptest}
\end{table*}

\begin{table*}[!t]
\resizebox{0.95\textwidth}{!}{
\begin{tabular}{l|l|lll|lll|lll|lll}
\hline
\multirow{2}{*}{Dataset} & \multirow{2}{*}{Seed} & \multicolumn{3}{|l|}{BERT} & \multicolumn{3}{|l|}{BERTweet} & \multicolumn{3}{|l|}{HateBERT} & \multicolumn{3}{|l}{mBERT} \\ \cline{3-14}
                         &                       & Min F1   & Max F1  & ES  & Min F1   & Max F1   & ES   & Min F1   & Max F1   & ES   & Min F1   & Max F1   & ES   \\ \hline

waseem & 12 & $L_{6}$: 0.758 & $L_{11}$: 0.806  & 0.484** & $L_{7}$: 0.723 & $L_{10}$: 0.786  & 0.620** & $L_{0}$: 0.758 & $L_{10}$: 0.813  & 0.558** & $L_{4}$: 0.736 & $L_{11}$: 0.788  & 0.523** \\ 

& 127 & $L_{5}$: 0.760 & $L_{4}$: 0.806  & 0.463 & $L_{6}$: 0.700 & $L_{11}$: 0.810  & 0.944** & $L_{1}$: 0.778 & $L_{10}$: 0.813  & 0.392* & $L_{8}$: 0.744 & $L_{5}$: 0.793  & 0.500** \\ 

& 451 & $L_{6}$: 0.760 & $L_{4}$: 0.799  & 0.379*
& $L_{1}$: 0.727 & $L_{11}$: 0.788  & 0.528**
& $L_{1}$: 0.752 & $L_{10}$: 0.813  & 0.614**
& $L_{9}$: 0.732 & $L_{5}$: 0.790  & 0.582** \\ \hdashline

davidson & 12 & $L_{11}$: 0.887 & $L_{1}$: 0.930  & 0.837**
& $L_{6}$: 0.887 & $L_{5}$: 0.936  & 0.895**
& $L_{7}$: 0.908 & $L_{3}$:  0.932  & 0.512**
& $L_{10}$: 0.852 & $L_{2}$:  0.920  & 1.36** \\

& 127 & $L_{2}$:  0.903 & $L_{5}$: 0.928  & 0.480**
& $L_{7}$: 0.900 & $L_{3}$:  0.935  & 0.782**
& $L_{10}$: 0.904 & $L_{5}$: 0.932  & 0.561**
& $L_{8}$: 0.888 & $L_{5}$: 0.918  & 0.576** \\

& 451 & $L_{10}$: 0.889 & $L_{4}$: 0.931  & 0.788**
& $L_{7}$: 0.905 & $L_{3}$:  0.935  & 0.671**
& $L_{7}$: 0.906 & $L_{4}$: 0.930  & 0.461**
& $L_{11}$: 0.893 & $L_{4}$: 0.923  & 0.618** \\ \hline

founta & 12 & $L_{7}$: 0.916 & $L_{4}$: 0.929  & 0.488**
& $L_{8}$: 0.921 & $L_{4}$: 0.930  & 0.378*
& $L_{2}$:  0.916 & $L_{9}$: 0.928  & 0.484**
& $L_{11}$: 0.890 & $L_{4}$: 0.924  & 1.121** \\

& 127 & $L_{0}$: 0.920 & $L_{5}$: 0.929  & 0.334*
& $L_{0}$: 0.918 & $L_{11}$: 0.928  & 0.401*
& $L_{9}$: 0.923 & $L_{4}$: 0.928  & 0.232
& $L_{10}$: 0.908 & $L_{5}$: 0.922  & 0.503** \\

& 451 & $L_{3}$: 0.921 & $L_{4}$: 0.928  & 0.280*
& $L_{6}$: 0.920 & $L_{3}$: 0.930  & 0.441*
& $L_{11}$: 0.916 & $L_{2}$:  0.928  & 0.453
& $L_{2}$:  0.904 & $L_{4}$: 0.918  & 0.489** \\ \hdashline

olid & 12 & $L_{1}$: 0.742 & $L_{9}$: 0.799  & 0.359*
& $L_{0}$: 0.747 & $L_{6}$: 0.805  & 0.388*
& $L_{0}$: 0.744 & $L_{7}$: 0.797  & 0.302*
& $L_{8}$: 0.700 & $L_{3}$:  0.750  & 0.220 \\

& 127 & $L_{0}$: 0.732 & $L_{8}$: 0.793  & 0.346*
& $L_{0}$: 0.760 & $L_{9}$: 0.817  & 0.323*
& $L_{6}$: 0.750 & $L_{8}$: 0.806  & 0.287*
& $L_{10}$: 0.624 & $L_{4}$: 0.755  & 0.509** \\

& 451 & $L_{2}$:  0.748 & $L_{11}$: 0.802  & 0.321*
& $L_{1}$: 0.764 & $L_{5}$: 0.812  & 0.307*
& $L_{0}$: 0.738 & $L_{3}$: 0.804  & 0.388*
& $L_{10}$: 0.681 & $L_{4}$: 0.765  & 0.493** \\ \hdashline

hatexplain & 12 & $L_{4}$: 0.695 & $L_{10}$: 0.766  & 1.054**
& $L_{6}$: 0.586 & $L_{9}$: 0.770  & 2.616**
& $L_{7}$: 0.638 & $L_{4}$: 0.766  & 0.1671**
& $L_{10}$: 0.647 & $L_{7}$: 0.739  & 0.1.33** \\

& 127 & $L_{9}$: 0.721 & $L_{7}$: 0.763  & 0.580**
& $L_{5}$: 0.717 & $L_{9}$: 0.757  & 0.559**
& $L_{4}$: 0.658 & $L_{3}$:  0.763  & 1.470**
& $L_{7}$: 0.616 & $L_{5}$: 0.736  & 1.724** \\

& 451 & $L_{11}$: 0.639 & $L_{4}$: 0.754  & 1.524**
& $L_{2}$:  0.691 & $L_{5}$: 0.761  & 1.024**
& $L_{1}$: 0.723 & $L_{11}$: 0.765  & 0.640**
& $L_{9}$: 0.616 & $L_{7}$: 0.737  & 1.782** \\ \hdashline

 dynabench & 12  & $L_{0}$: 0.697  & $L_{9}$: 0.746  & 1.108** 
 & $L_{0}$: 0.705  & $L_{9}$: 0.781  & 1.859**
& $L_{1}$: 0.706  & $L_{9}$: 0.765  & 1.414**
& $L_{0}$: 0.635  & $L_{4}$: 0.717  & 1.764** \\ 

 & 127 & $L_{6}$: 0.665  & $L_{10}$: 0.754  & 2.006**
& $L_{0}$: .710  & $L_{11}$: 0.783  & 1.614**
& $L_{0}$: 0.706 & $L_{10}$: 0.764  & 1.394**
& $L_{7}$: 0.661 & $L_{4}$: 0.719  & 1.316** \\

& 451 & $L_{2}$:  0.699 & $L_{9}$: 0.756  & 1.335**
& $L_{0}$: 0.711 & $L_{9}$: 0.782  & 1.716**
& $L_{0}$: 0.717 & $L_{11}$: 0.770  & 1.257**
& $L_{0}$: 0.691 & $L_{4}$: 0.720  & 0.633** \\ \hdashline

toxigen & 12 & $L_{0}$: 0.767 & $L_{11}$: 0.806  & 2.216**
& $L_{1}$: 0.780 & $L_{11}$: 0.812  & 2.026**
& $L_{0}$: 0.780 & $L_{11}$: 0.812  & 2.026**
& $L_{0}$: 0.754 & $L_{4}$: 0.777  & 1.34** \\

    & 127 & $L_{0}$: 0.769 & $L_{11}$: 0.803  & 2.044**
& $L_{1}$: 0.788 & $L_{11}$: 0.826  & 2.313**
& $L_{0}$: 0.775 & $L_{11}$: 0.816  & 2.396**
& $L_{0}$: 0.746 & $L_{5}$: 0.774  & 1.619** 
\\

    & 451 & $L_{0}$: 0.768 & $L_{11}$: 0.804  & 2.263**
& $L_{1}$: 0.787 & $L_{11}$: 0.826  & 2.551**
& $L_{0}$: 0.778 & $L_{11}$: 0.813  & 2.343**
& $L_{0}$: 0.746 & $L_{7}$: 0.775  & 1.619** 
\\ \hline
\end{tabular}}
\caption{\textbf{RQ4:} Comparison of minimum and maximum macro F1 obtained per MLP seed ($ms$) per BERT-variant. ES stands for effect size. ** and * indicates whether the difference in minimum and maximum macro F1 is significant by $\le0.05$ and $\le0.001$ p-value, respectively.}
\label{tab:layer_wise_per_seed}
\end{table*}

\subsection{RQ3: Extended Experiments}
\label{app:rq3}
The Online Language Modelling \footnote{\url{https://huggingface.co/olm}} initiative by Hugging Face is a repository of updated PLM models and tokenizers that are pretrained on regular and latest Internet snapshots obtained via Common Crawl and Wikipedia. The initiative aims to induce explicit knowledge of newer concepts and updated factual information in the PLMs. At the time of compiling this research, the OLM project had $6$ models and $19$ datasets snapshots contributed to the repository. Out of these, the two RoBERTa models released in October 2022 and December 2022 are employed in our research.

\subsection{RQ4: Extended Experiments}
\label{app:rq4}
\begin{figure*}[!t]
\centering
    \begin{subfigure}[b]{0.45\textwidth}
        \includegraphics[width=\linewidth]{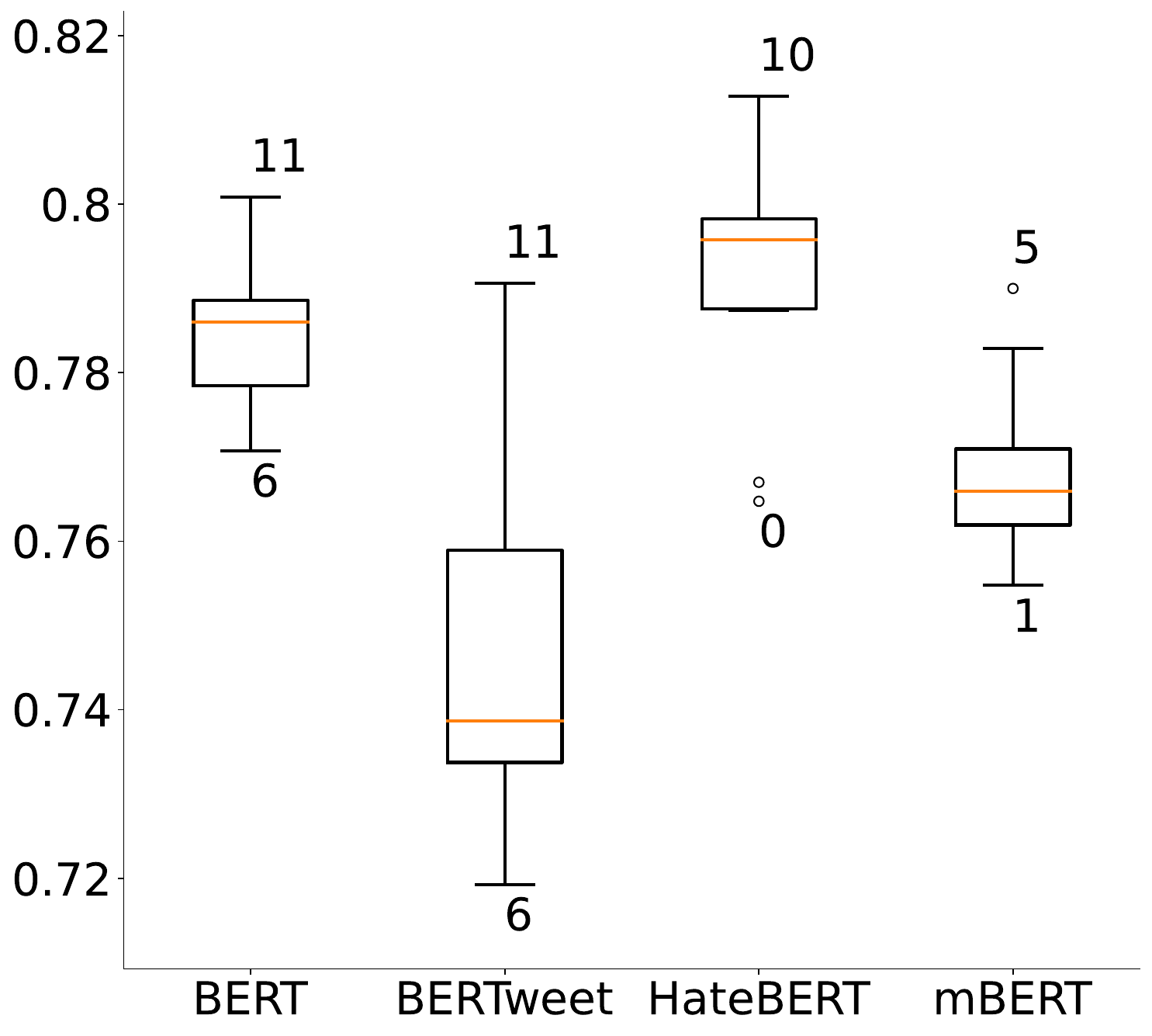}
        \caption{\Dii}
    \end{subfigure}
    \hfill
    \begin{subfigure}[b]{0.45\textwidth}
        \includegraphics[width=\linewidth]{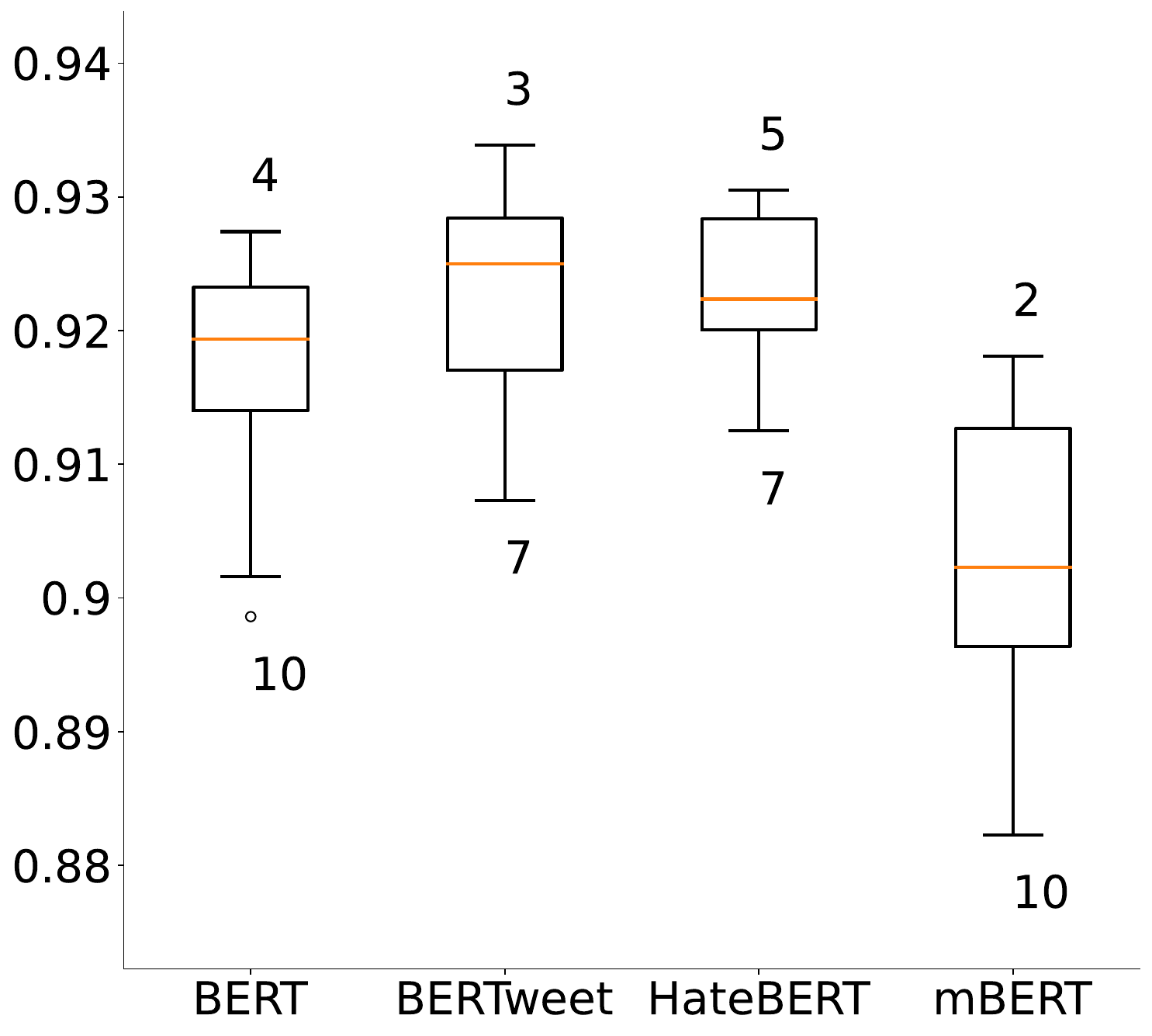}
        \caption{\Dvii}
    \end{subfigure}
    \hfill
    \begin{subfigure}[b]{0.45\textwidth}
        \includegraphics[width=\linewidth]{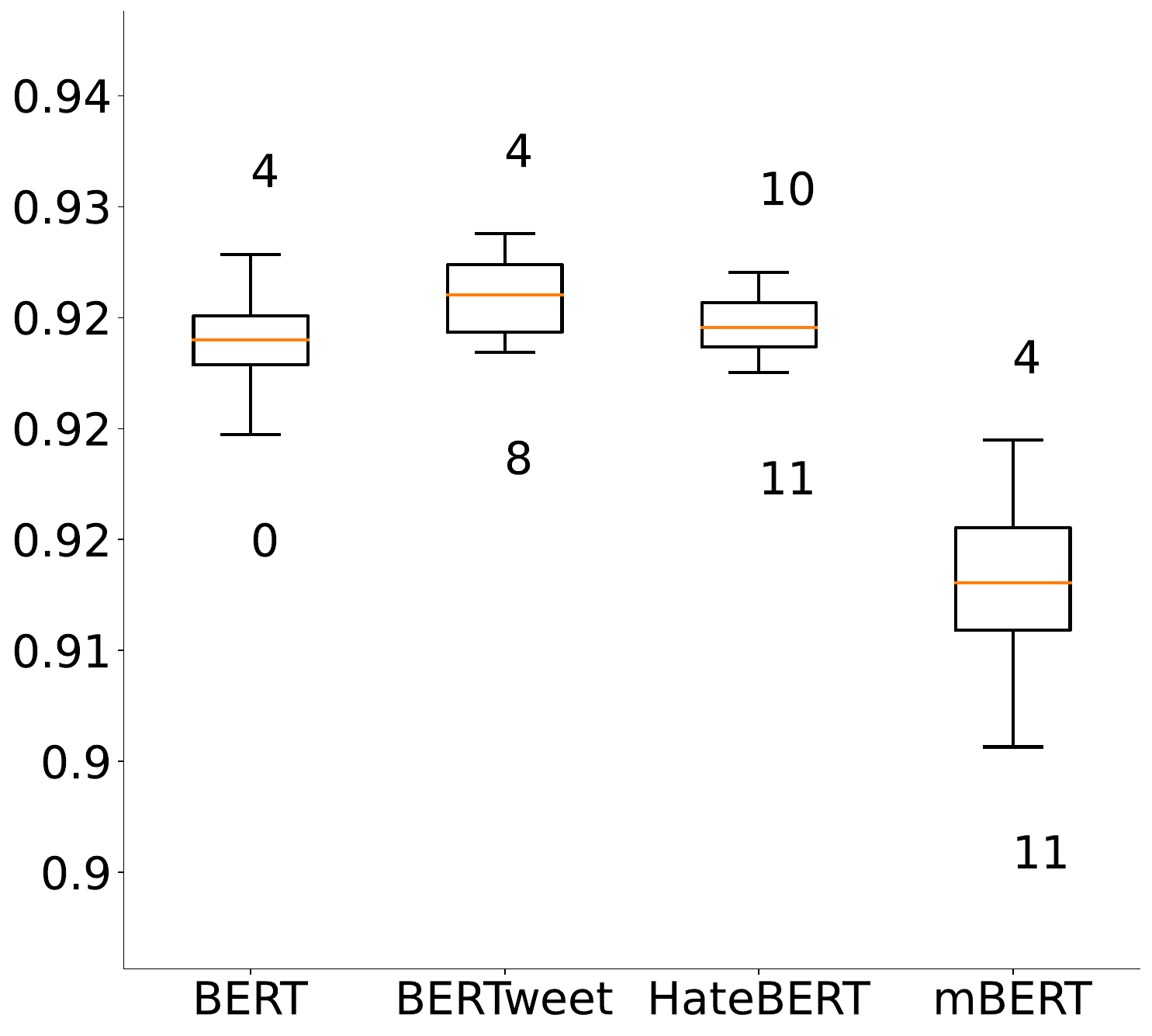}
        \caption{\Div}
    \end{subfigure}
    \hfill
    \begin{subfigure}[b]{0.45\textwidth}
        \includegraphics[width=\linewidth]{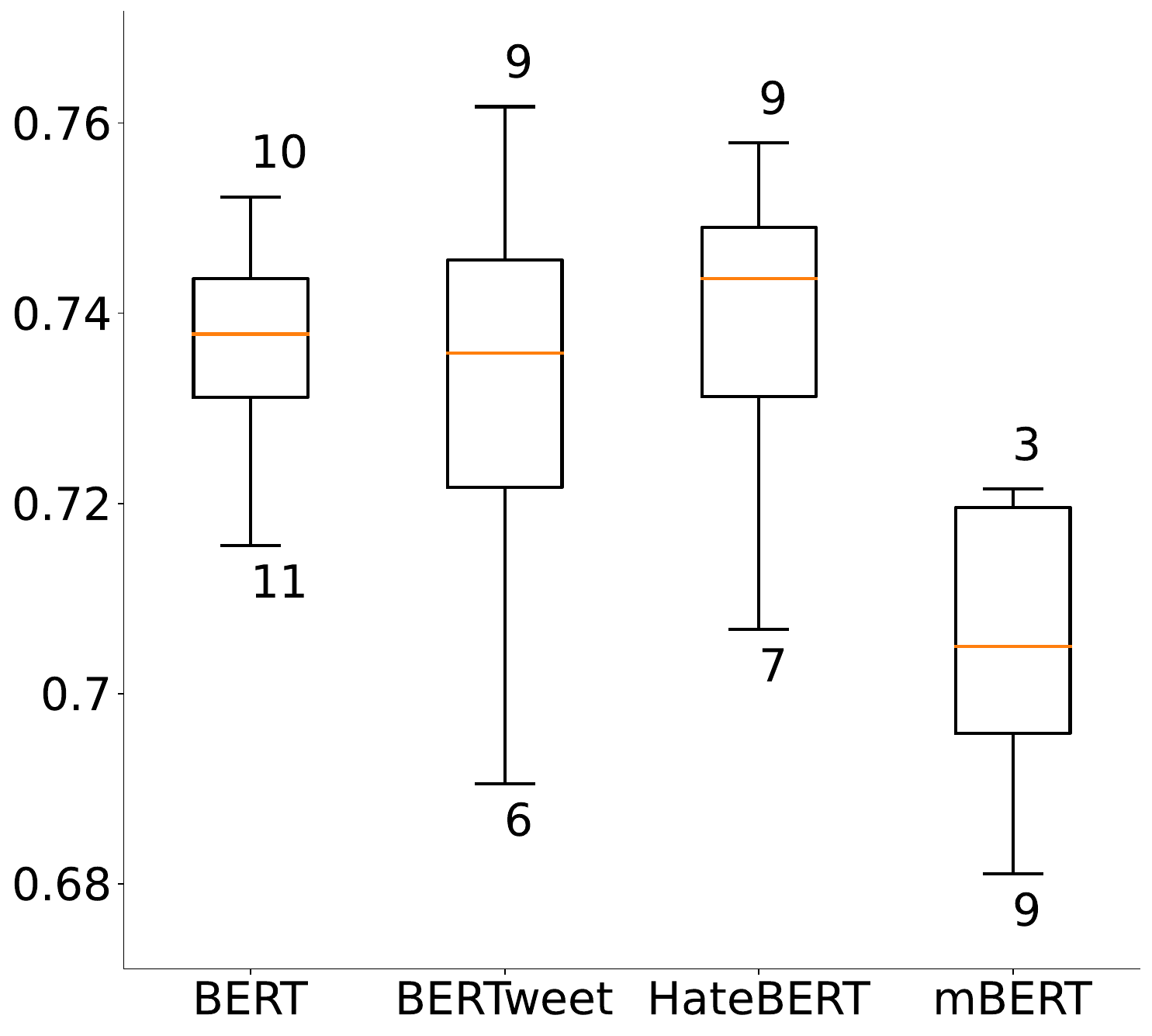}
        \caption{\Diii}
    \end{subfigure}
    \hfill
    \begin{subfigure}[b]{0.45\textwidth}
        \includegraphics[width=\linewidth]{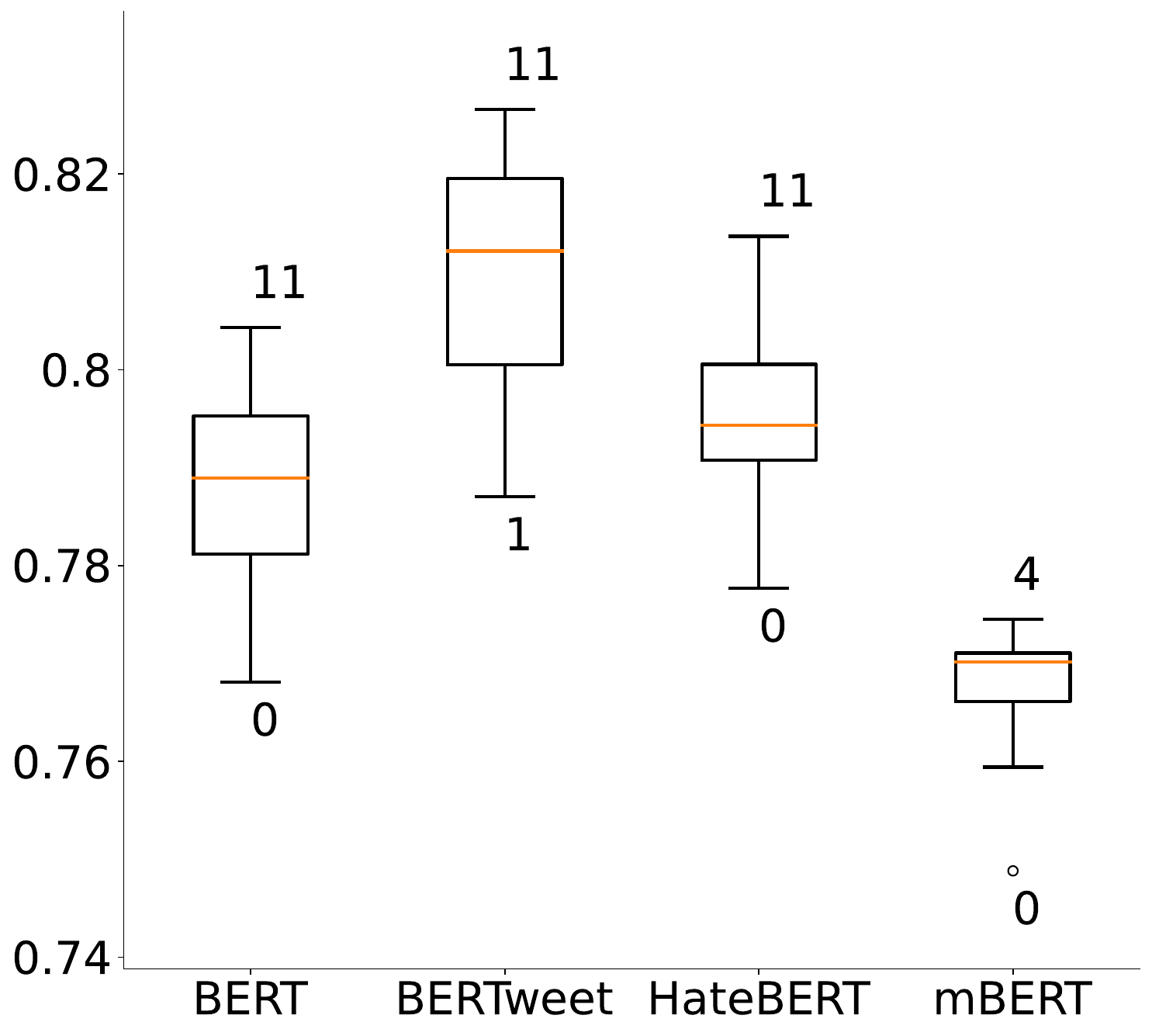}
        \caption{\Dv}
    \end{subfigure}    
    \caption{{\bf RQ4:} Extending from Figure \ref{fig:one_layer_at_a_time}(a,b) to rest of $5$ datasets -- Descriptive statistics of macro F1 when finetuning on top of individual layers of the BERT-variant highlighting the layer ($L_i$) that on average over MLP seeds ($ms$) leads to minimum and maximum macro F1. Here the $L_i$ is trainable while other layers are frozen.
    }
\label{fig:app_rest_one_layer_at_a_time}
\vspace{-3mm}
\end{figure*}

Figure \ref{fig:app_rest_one_layer_at_a_time} (a-e) provides an overview of the individual layer's contribution to performance when only the layer under consideration is trainable. Additionally, Table \ref{tab:layer_wise_per_seed} enlist the per-seed comparison of performance, respectively. We observe that there is no lottery ticket to the best/most critical layer when examined from the point of view of MLP seeds, BERT-variants, and datasets.

\begin{figure*}[!t]
\centering
    \begin{subfigure}[b]{0.45\textwidth}
        \includegraphics[width=\linewidth]{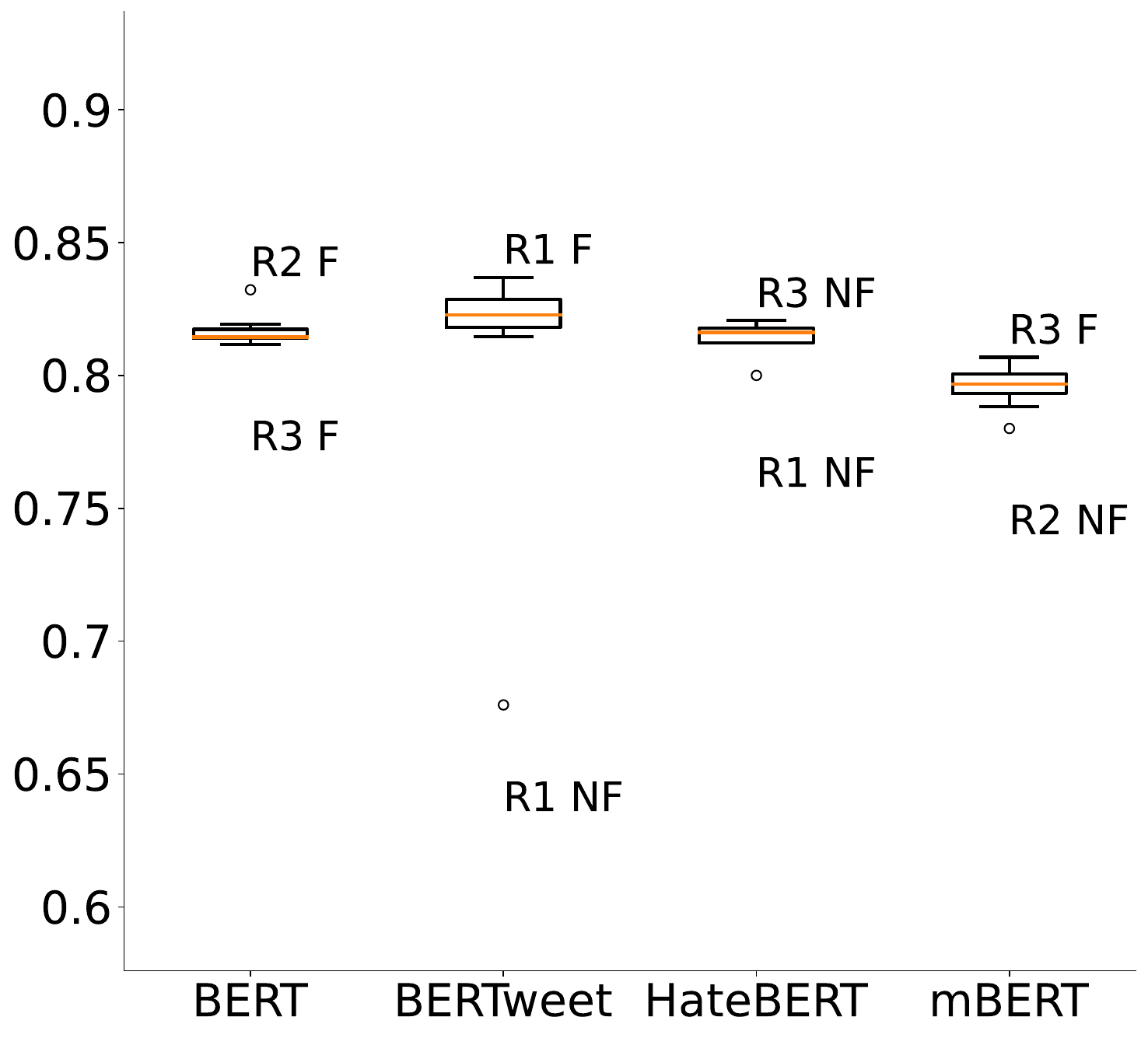}
        \caption{\Dii}
    \end{subfigure}
    \hfill
    \begin{subfigure}[b]{0.45\textwidth}
        \includegraphics[width=\linewidth]{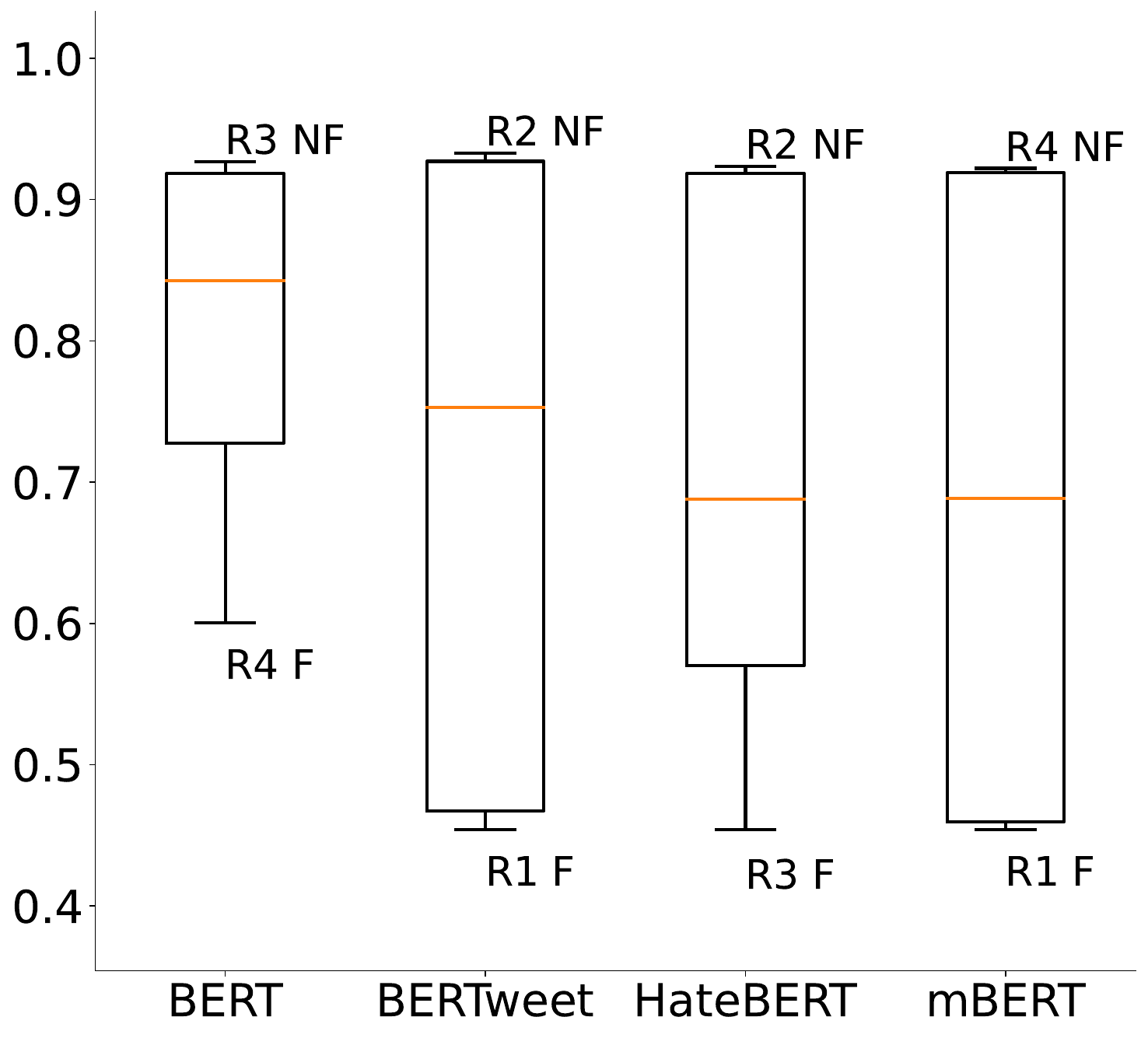}
        \caption{\Dvii}
    \end{subfigure}
    \hfill
    \begin{subfigure}[b]{0.45\textwidth}
        \includegraphics[width=\linewidth]{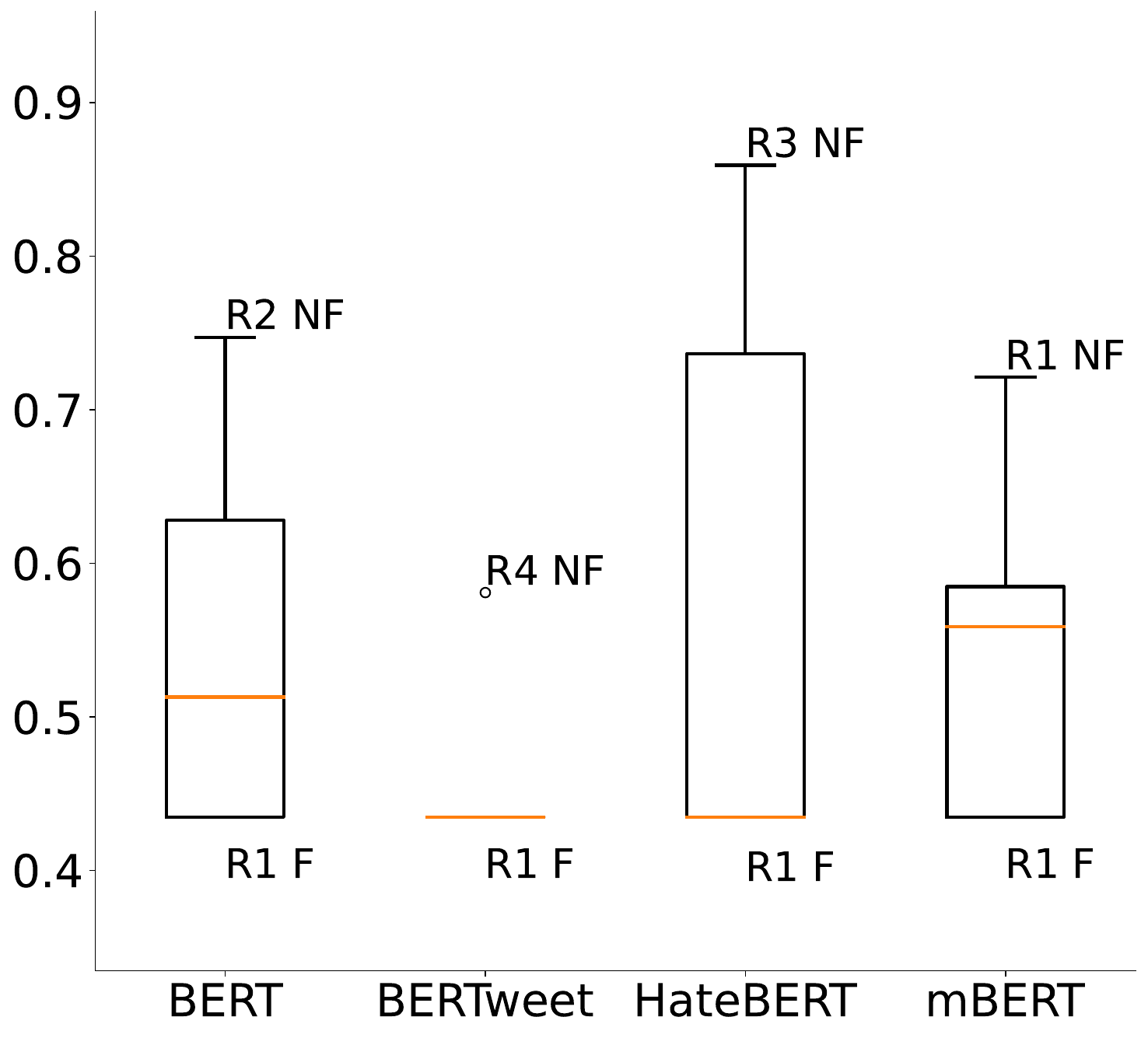}
        \caption{\Div}
    \end{subfigure}
    \hfill
    \begin{subfigure}[b]{0.45\textwidth}
        \includegraphics[width=\linewidth]{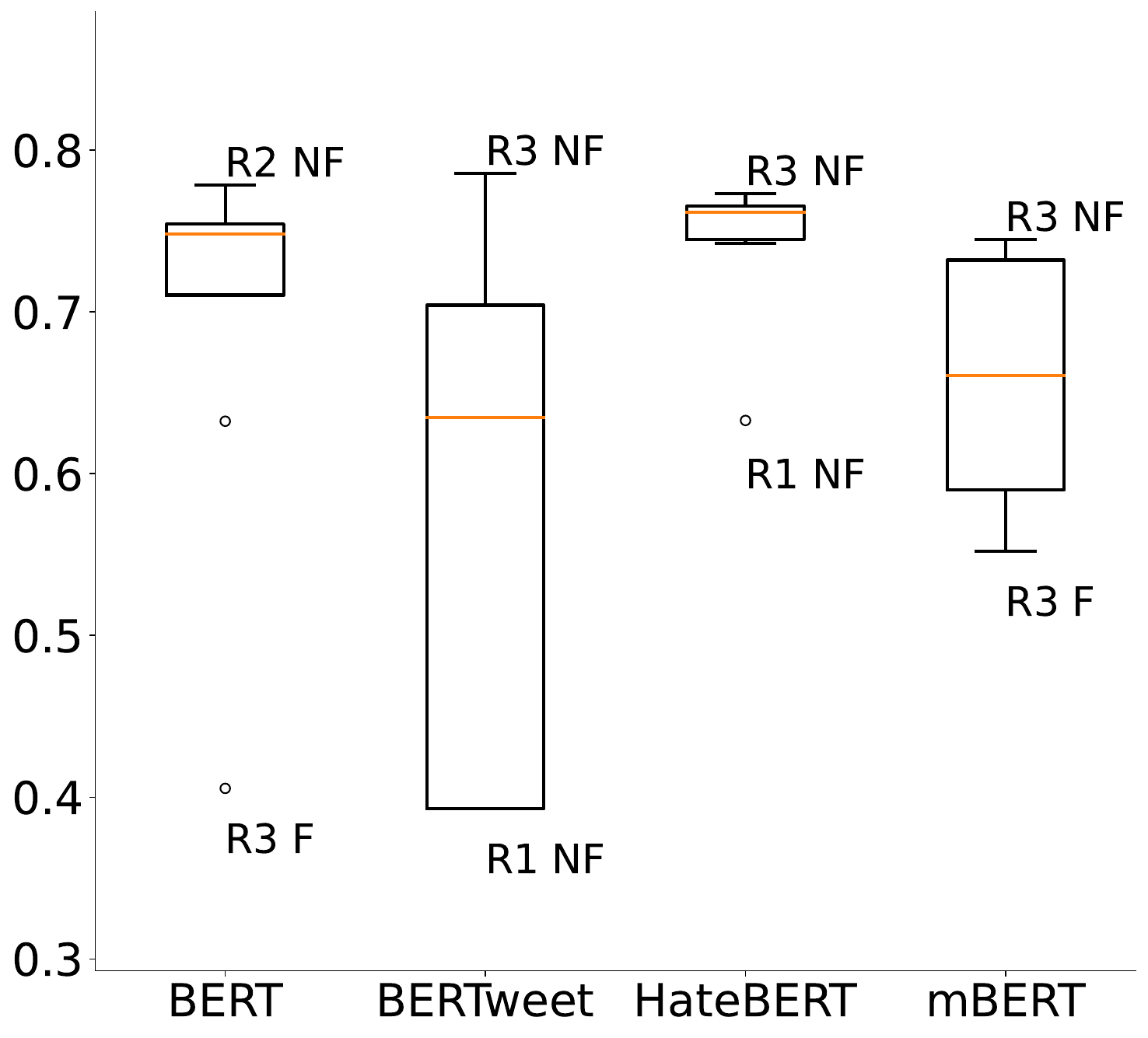}
        \caption{\Diii}
    \end{subfigure}
    \hfill
    \begin{subfigure}[b]{0.45\textwidth}
        \includegraphics[width=\linewidth]{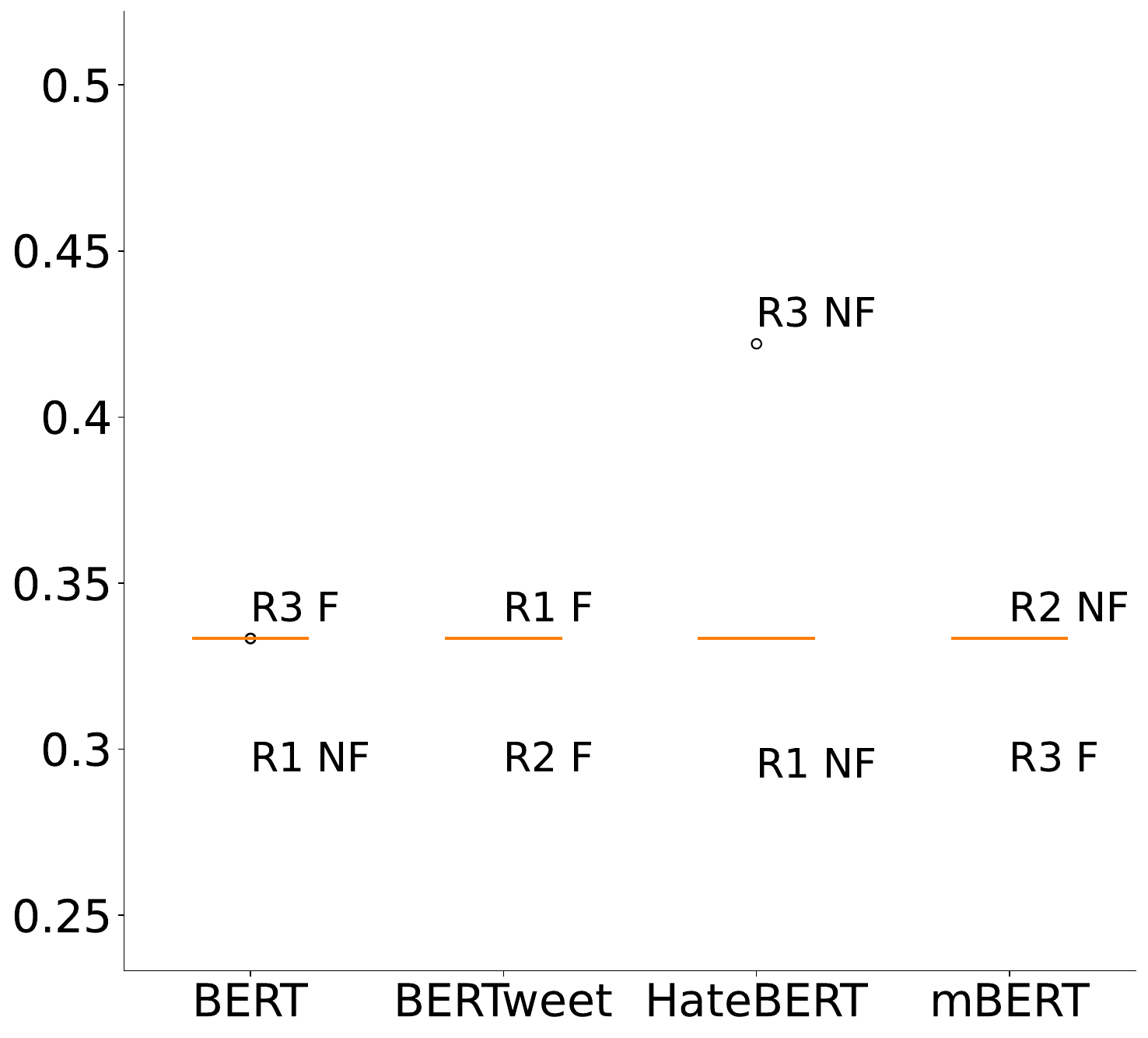}
        \caption{\Dv}
    \end{subfigure}    
    \caption{{\bf RQ4:} Extending from Figure \ref{fig:one_layer_at_a_time}(c,d) to rest of $5$ datasets -- -Descriptive statistics of macro F1 when finetuning while constraining a region of layers to be frozen (Suffix F) or non-frozen while all others are frozen (Suffix NF) for different BERT-variant highlighting the region ($R_i$) that on average over MLP seeds ($ms$) leads to minimum and maximum macro F1.
    }
\label{fig:app_rest_region}
\vspace{-3mm}
\end{figure*}

While in the layer-wise analysis so far, we looked at trainable layers one at a time, we also looked at regions of results in a (un)frozen manner in Figure \ref{fig:app_rest_region} (a-e) and Table \ref{tab:region_wise_ptest}.

\subsection{RQ 5: Extended Experiments}
\label{app:rq5}
\begin{figure*}
\centering
    \begin{subfigure}[b]{0.475\textwidth}
        \includegraphics[width=\linewidth]{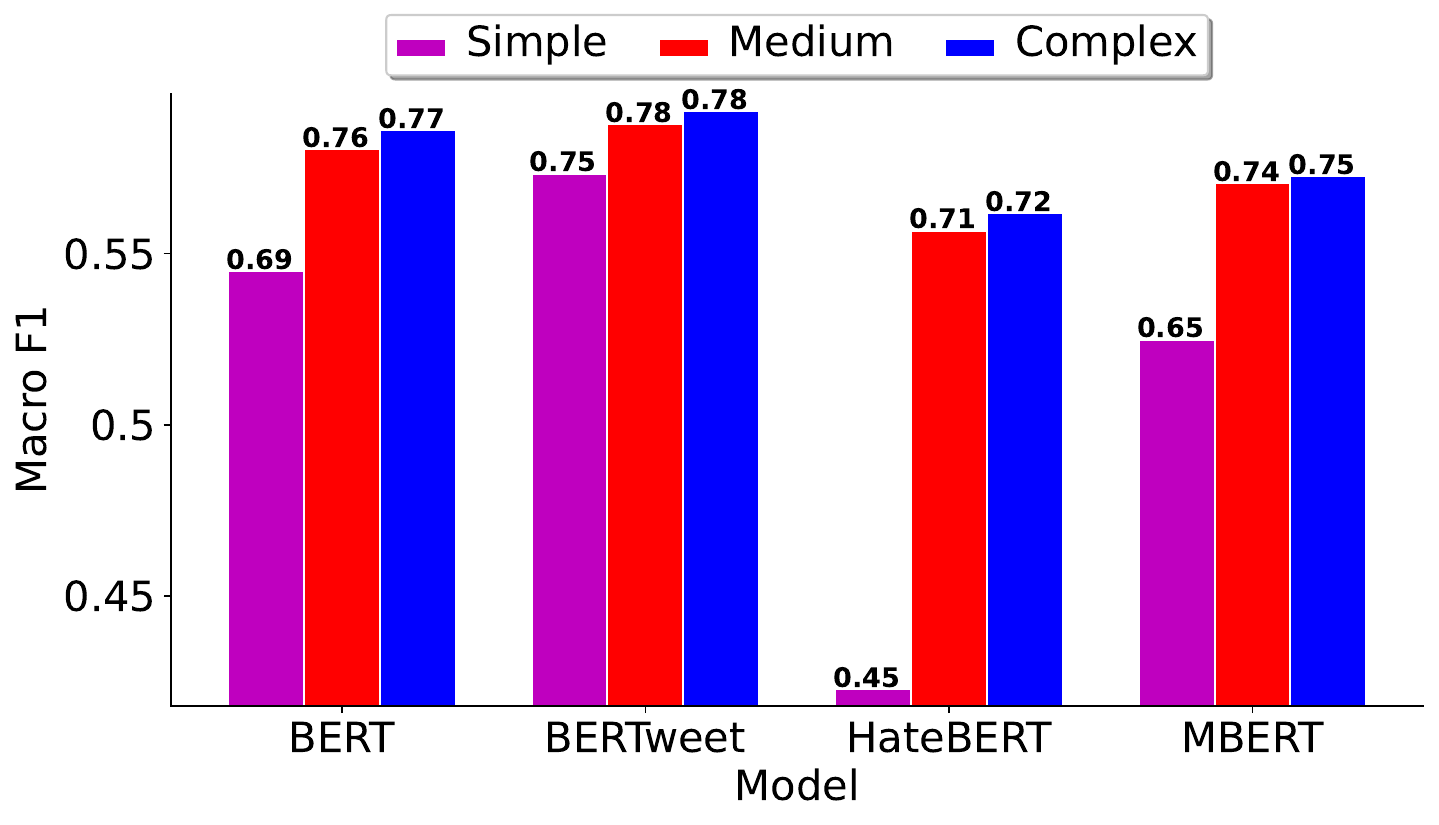}
        \caption{\Dii}
    \end{subfigure}
    \hfill
    \begin{subfigure}[b]{0.475\textwidth}
        \includegraphics[width=\linewidth]{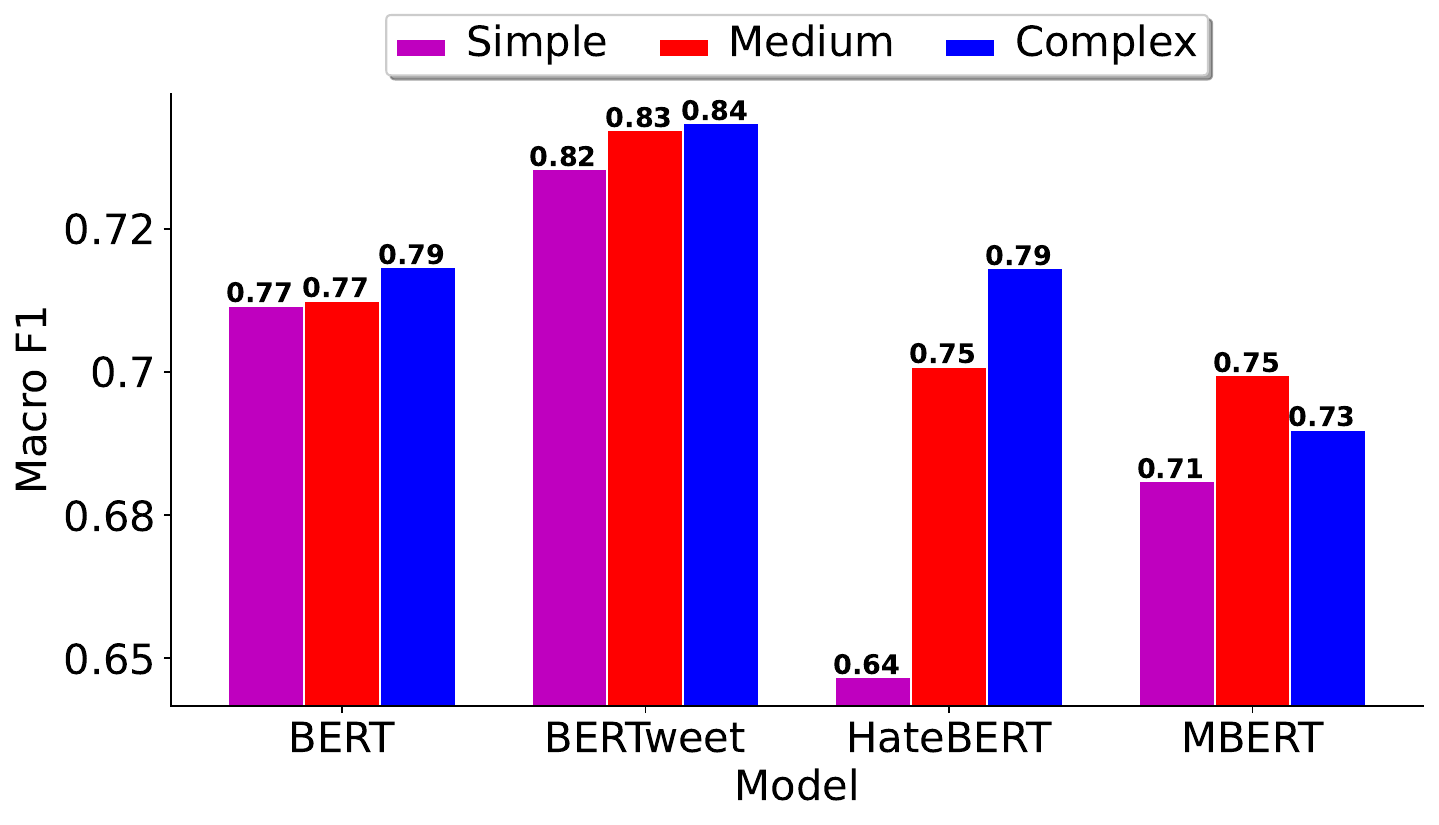}
        \caption{\Dvii}
    \end{subfigure}
    \hfill
    \begin{subfigure}[b]{0.475\textwidth}
        \includegraphics[width=\linewidth]{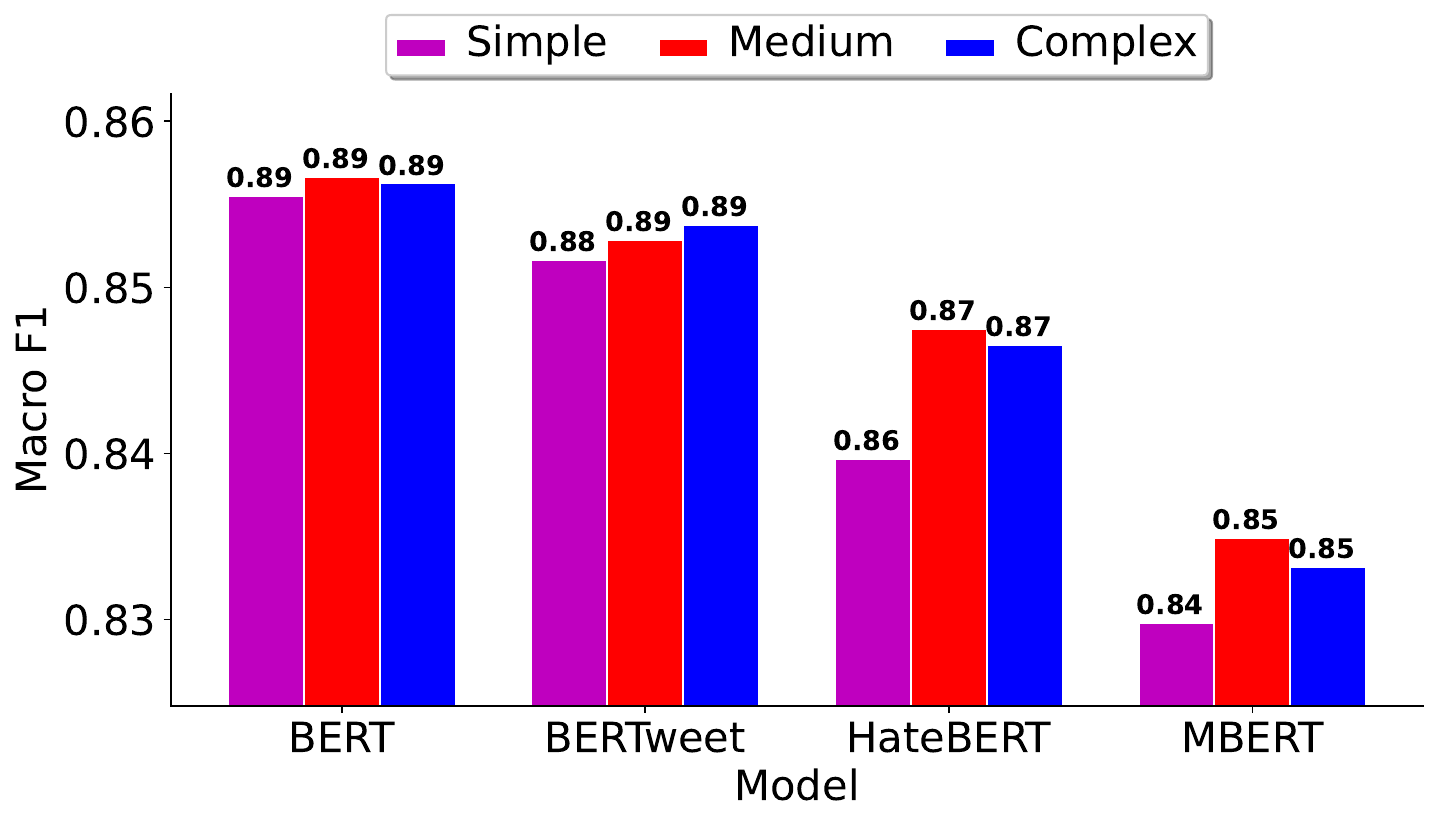}
        \caption{\Div}
    \end{subfigure}
    \hfill
    \begin{subfigure}[b]{0.475\textwidth}
        \includegraphics[width=\linewidth]{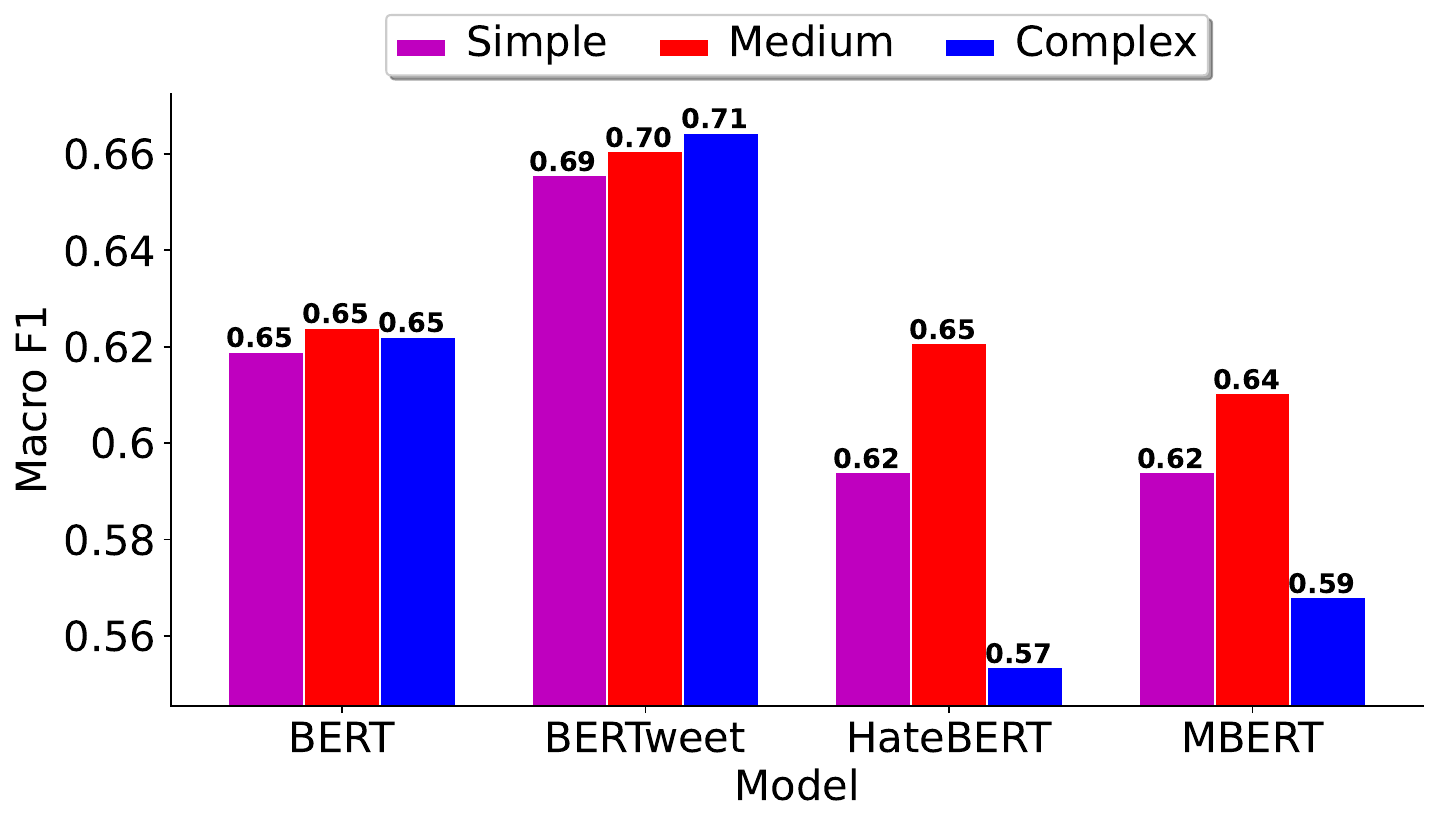}
        \caption{\Diii}
    \end{subfigure}
    \hfill
    \begin{subfigure}[b]{0.475\textwidth}
        \includegraphics[width=\linewidth]{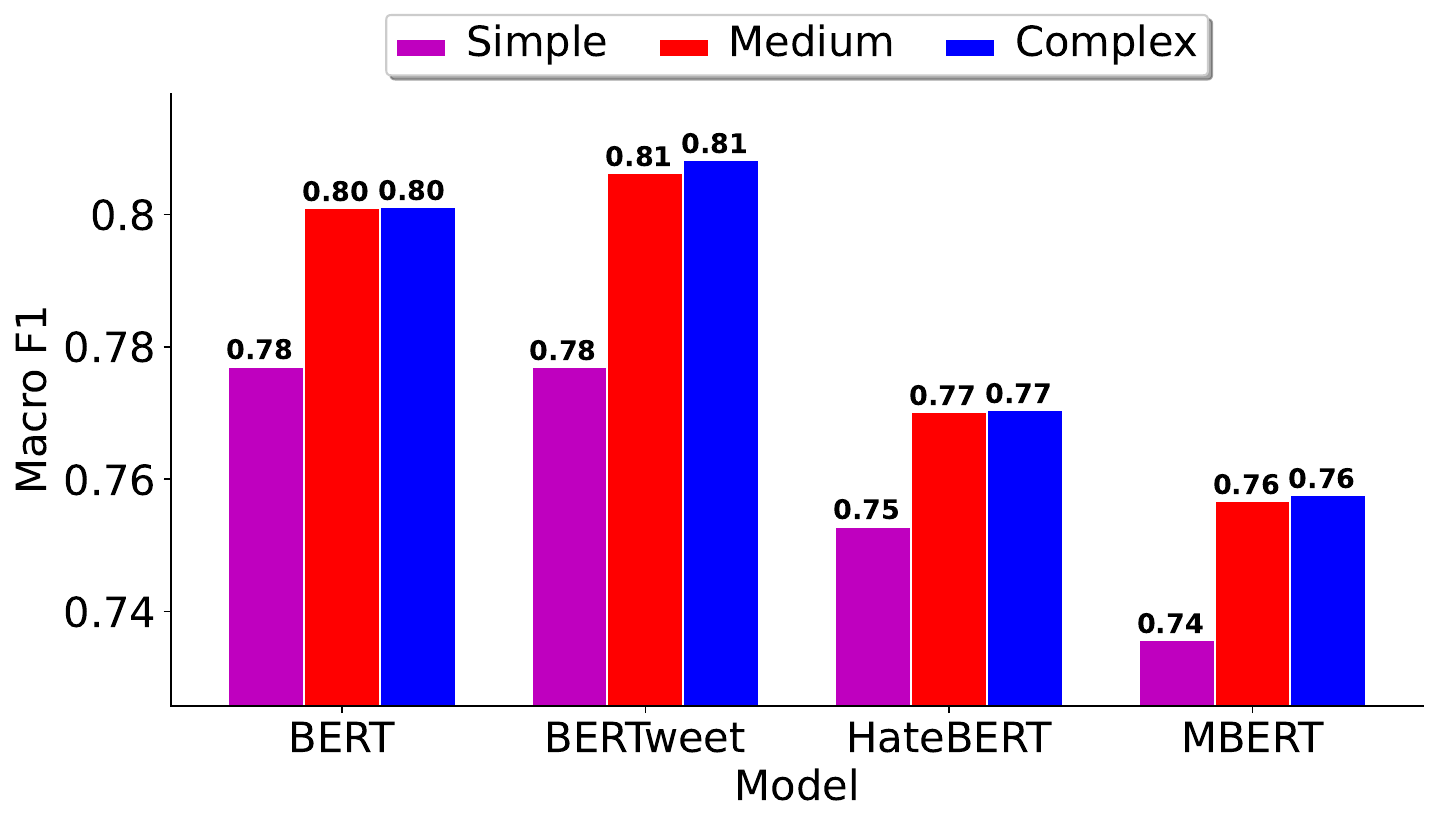}
        \caption{\Dv}
    \end{subfigure}    
    \caption{{\bf RQ5:} Extending from Figure \ref{fig:cc_analysis} to rest of $5$ datasets -- Macro F1 scores (averaged over MLP seeds $ms$) employing BERT-variants (BERT, BERTweet, HateBERT, and mBERT). Classification heads of varying complexity (simple, medium, and complex) are utilized to capture their effect on BERT-variants employed for hate detection.
    }
\label{fig:cc_analysis_all}
\vspace{-3mm}
\end{figure*}

Figure \ref{fig:cc_analysis_all} (a-e) provides an overview of the impact of classification head architecture on the finetuning performance. Granular results controlling for MLP seeds ($ms$) are enlisted in Table \ref{tab:cc_wise_ptest}.

\label{app:robert_intern}

\begin{table*}[!h]
\centering
\resizebox{0.85\textwidth}{!}{
\begin{tabular}{l|l|l|lll|lll|lll|lll}
\hline
              Dataset & BERT & SEED     &  $R_1$T &  $R_1$F & $R_1$T/F:ES &  $R_2$T &  $R_2$F & $R_2$T/F:ES &  $R_3$T &  $R_3$F & $R_3$T/F:ES &  $R_4$T &  $R_4$F & $R_4$T/F:ES \\ \hline
\Dii & BERT & 12 &                0.815 &                 0.816 &                                       0.007 &                0.840 &                 0.820 &                                       0.232 &                0.821 &                 0.822 &                                       0.009 &                0.816 &                 0.814 &                                       0.028 \\
              &                              & 127 &                0.795 &                 0.822 &                                      0.298* &                0.833 &                 0.801 &                                      0.307* &                0.786 &                 0.811 &                                       0.245 &                0.803 &                 0.831 &                                      0.297* \\
              &                              & 451 &                0.831 &                 0.812 &                                       0.189 &                0.824 &                 0.822 &                                       0.015 &                0.828 &                 0.811 &                                       0.186 &                0.824 &                 0.813 &                                       0.078 \\
              & BERTweet & 12 &                0.836 &                 0.799 &                                      0.392* &                0.814 &                 0.827 &                                       0.130 &                0.831 &                 0.820 &                                       0.086 &                0.812 &                 0.823 &                                       0.085 \\
              &                              & 127 &                0.842 &                 0.803 &                                      0.387* &                0.831 &                 0.812 &                                      0.279* &                0.820 &                 0.821 &                                       0.066 &                0.811 &                 0.842 &                                      0.352* \\
              &                              & 451 &                0.832 &                 0.426 &                                     4.936** &                0.844 &                 0.819 &                                      0.283* &                0.818 &                 0.827 &                                       0.064 &                0.821 &                 0.820 &                                       0.001 \\
              & HateBERT & 12 &                0.799 &                 0.812 &                                       0.083 &                0.831 &                 0.823 &                                       0.107 &                0.817 &                 0.812 &                                       0.086 &                0.818 &                 0.799 &                                       0.207 \\
              &                              & 127 &                0.814 &                 0.767 &                                      0.432* &                0.809 &                 0.820 &                                       0.114 &                0.815 &                 0.829 &                                       0.129 &                0.818 &                 0.828 &                                       0.146 \\
              &                              & 451 &                0.824 &                 0.820 &                                       0.034 &                0.821 &                 0.805 &                                       0.152 &                0.819 &                 0.821 &                                       0.029 &                0.800 &                 0.822 &                                       0.224 \\
              & mBERT & 12 &                0.802 &                 0.798 &                                       0.074 &                0.790 &                 0.801 &                                       0.095 &                0.806 &                 0.793 &                                       0.069 &                0.826 &                 0.806 &                                       0.183 \\
              &                              & 127 &                0.799 &                 0.805 &                                       0.037 &                0.763 &                 0.802 &                                      0.370* &                0.813 &                 0.794 &                                       0.161 &                0.788 &                 0.802 &                                       0.166 \\
              &                              & 451 &                0.791 &                 0.786 &                                       0.022 &                0.812 &                 0.738 &                                     0.733** &                0.802 &                 0.798 &                                       0.033 &                0.786 &                 0.797 &                                       0.119 \\ \hdashline
\Dvii & BERT & 12 &                0.926 &                 0.921 &                                       0.116 &                0.919 &                 0.924 &                                       0.096 &                0.454 &                 0.930 &                                    13.303** &                0.893 &                 0.922 &                                     0.551** \\
              &                              & 127 &                0.454 &                 0.905 &                                    13.147** &                0.454 &                 0.921 &                                    13.839** &                0.927 &                 0.919 &                                       0.159 &                0.454 &                 0.915 &                                    11.960** \\
              &                              & 451 &                0.918 &                 0.925 &                                       0.114 &                0.932 &                 0.910 &                                      0.454* &                0.454 &                 0.932 &                                    14.392** &                0.454 &                 0.923 &                                    12.794** \\
              & BERTweet & 12 &                0.454 &                 0.926 &                                    12.596** &                0.454 &                 0.935 &                                    13.664** &                0.862 &                 0.929 &                                     1.251** &                0.454 &                 0.931 &                                    14.453** \\
              &                              & 127 &                0.454 &                 0.924 &                                    12.368** &                0.454 &                 0.930 &                                    15.046** &                0.454 &                 0.933 &                                    15.144** &                0.506 &                 0.933 &                                     7.991** \\
              &                              & 451 &                0.454 &                 0.929 &                                    14.575** &                0.454 &                 0.934 &                                    15.645** &                0.454 &                 0.926 &                                    13.377** &                0.454 &                 0.882 &                                     9.952** \\
              & HateBERT & 12 &                0.454 &                 0.919 &                                    12.211** &                0.454 &                 0.919 &                                    12.672** &                0.454 &                 0.920 &                                    13.229** &                0.454 &                 0.928 &                                    13.370** \\
              &                              & 127 &                0.924 &                 0.924 &                                       0.037 &                0.454 &                 0.934 &                                    13.568** &                0.454 &                 0.911 &                                    12.876** &                0.454 &                 0.922 &                                    12.962** \\
              &                              & 451 &                0.454 &                 0.454 &                                       0.000 &                0.917 &                 0.917 &                                       0.026 &                0.454 &                 0.920 &                                    12.774** &                0.454 &                 0.919 &                                    13.289** \\
              & mBERT & 12 &                0.454 &                 0.913 &                                    12.393** &                0.454 &                 0.925 &                                    12.538** &                0.483 &                 0.923 &                                     9.358** &                0.454 &                 0.923 &                                    13.992** \\
              &                              & 127 &                0.454 &                 0.902 &                                    12.214** &                0.454 &                 0.916 &                                    13.964** &                0.454 &                 0.913 &                                    10.779** &                0.454 &                 0.923 &                                    13.322** \\
              &                              & 451 &                0.454 &                 0.921 &                                    12.280** &                0.476 &                 0.916 &                                     9.423** &                0.457 &                 0.924 &                                    11.758** &                0.454 &                 0.920 &                                    13.139** \\ \hdashline
\Div & BERT & 12 &                0.435 &                 0.875 &                                    16.947** &                0.435 &                 0.903 &                                    22.165** &                0.435 &                 0.435 &                                       0.000 &                0.435 &                 0.906 &                                    20.983** \\
              &                              & 127 &                0.435 &                 0.435 &                                       0.000 &                0.435 &                 0.435 &                                       0.000 &                0.435 &                 0.435 &                                       0.000 &                0.904 &                 0.435 &                                    21.930** \\
              &                              & 451 &                0.435 &                 0.901 &                                    20.681** &                0.435 &                 0.904 &                                    21.262** &                0.435 &                 0.435 &                                       0.000 &                0.435 &                 0.435 &                                       0.000 \\
              & BERTweet & 12 &                0.435 &                 0.435 &                                       0.000 &                0.435 &                 0.435 &                                       0.000 &                0.435 &                 0.435 &                                       0.000 &                0.435 &                 0.755 &                                     9.979** \\
              &                              & 127 &                0.435 &                 0.435 &                                       0.000 &                0.435 &                 0.435 &                                       0.000 &                0.435 &                 0.435 &                                       0.000 &                0.435 &                 0.435 &                                       0.000 \\
              &                              & 451 &                0.435 &                 0.435 &                                       0.000 &                0.435 &                 0.435 &                                       0.000 &                0.435 &                 0.435 &                                       0.000 &                0.435 &                 0.553 &                                     3.100** \\
              & HateBERT & 12 &                0.435 &                 0.435 &                                       0.000 &                0.435 &                 0.435 &                                       0.000 &                0.435 &                 0.910 &                                    19.936** &                0.435 &                 0.435 &                                       0.000 \\
              &                              & 127 &                0.435 &                 0.435 &                                       0.000 &                0.435 &                 0.915 &                                    24.121** &                0.435 &                 0.873 &                                    16.404** &                0.435 &                 0.871 &                                    15.600** \\
              &                              & 451 &                0.435 &                 0.435 &                                       0.000 &                0.435 &                 0.905 &                                    22.709** &                0.435 &                 0.794 &                                    11.383** &                0.435 &                 0.889 &                                    19.753** \\
              & mBERT & 12 &                0.435 &                 0.834 &                                    13.390** &                0.758 &                 0.435 &                                     9.584** &                0.435 &                 0.877 &                                    16.618** &                0.435 &                 0.435 &                                       0.000 \\
              &                              & 127 &                0.435 &                 0.435 &                                       0.000 &                0.435 &                 0.854 &                                    14.137** &                0.435 &                 0.435 &                                       0.000 &                0.435 &                 0.435 &                                       0.000 \\
              &                              & 451 &                0.435 &                 0.895 &                                    19.070** &                0.435 &                 0.435 &                                       0.000 &                0.435 &                 0.435 &                                       0.000 &                0.435 &                 0.909 &                                    20.701** \\ \hdashline
\Dvi & BERT & 12 &                0.737 &                 0.740 &                                       0.008 &                0.773 &                 0.795 &                                       0.075 &                0.777 &                 0.767 &                                       0.008 &                0.778 &                 0.790 &                                       0.081 \\
              &                              & 127 &                0.755 &                 0.762 &                                       0.052 &                0.786 &                 0.765 &                                       0.103 &                0.783 &                 0.775 &                                       0.093 &                0.767 &                 0.785 &                                       0.085 \\
              &                              & 451 &                0.771 &                 0.777 &                                       0.019 &                0.768 &                 0.800 &                                       0.186 &                0.771 &                 0.798 &                                       0.061 &                0.775 &                 0.794 &                                       0.134 \\
              & BERTweet & 12 &                0.774 &                 0.419 &                                     1.535** &                0.808 &                 0.803 &                                       0.020 &                0.419 &                 0.825 &                                     1.950** &                0.773 &                 0.815 &                                      0.307* \\
              &                              & 127 &                0.792 &                 0.419 &                                     1.644** &                0.419 &                 0.814 &                                     1.776** &                0.419 &                 0.815 &                                     1.846** &                0.419 &                 0.812 &                                     1.797** \\
              &                              & 451 &                0.804 &                 0.419 &                                     1.704** &                0.810 &                 0.811 &                                       0.048 &                0.790 &                 0.806 &                                       0.155 &                0.419 &                 0.804 &                                     1.749** \\
              & HateBERT & 12 &                0.787 &                 0.479 &                                     1.254** &                0.419 &                 0.770 &                                     1.409** &                0.764 &                 0.765 &                                       0.015 &                0.770 &                 0.795 &                                       0.187 \\
              &                              & 127 &                0.749 &                 0.749 &                                       0.042 &                0.776 &                 0.788 &                                       0.047 &                0.756 &                 0.762 &                                       0.050 &                0.751 &                 0.789 &                                       0.239 \\
              &                              & 451 &                0.769 &                 0.766 &                                       0.023 &                0.795 &                 0.793 &                                       0.024 &                0.783 &                 0.787 &                                       0.062 &                0.419 &                 0.765 &                                     1.435** \\
              & mBERT & 12 &                0.715 &                 0.735 &                                       0.094 &                0.681 &                 0.678 &                                       0.057 &                0.704 &                 0.775 &                                       0.244 &                0.740 &                 0.769 &                                       0.163 \\
              &                              & 127 &                0.780 &                 0.727 &                                       0.230 &                0.707 &                 0.763 &                                       0.266 &                0.419 &                 0.756 &                                     1.276** &                0.758 &                 0.761 &                                       0.015 \\
              &                              & 451 &                0.419 &                 0.419 &                                       0.000 &                0.764 &                 0.771 &                                       0.035 &                0.432 &                 0.772 &                                     1.343** &                0.730 &                 0.736 &                                       0.069 \\ \hdashline
\Diii & BERT & 12 &                0.747 &                 0.746 &                                       0.004 &                0.769 &                 0.776 &                                       0.133 &                0.431 &                 0.753 &                                     4.846** &                0.762 &                 0.758 &                                       0.058 \\
              &                              & 127 &                0.770 &                 0.393 &                                     7.340** &                0.718 &                 0.783 &                                     0.945** &                0.393 &                 0.721 &                                     5.729** &                0.733 &                 0.750 &                                       0.240 \\
              &                              & 451 &                0.769 &                 0.758 &                                       0.180 &                0.747 &                 0.776 &                                      0.400* &                0.393 &                 0.779 &                                     8.101** &                0.759 &                 0.702 &                                     0.817** \\
              & BERTweet & 12 &                0.775 &                 0.393 &                                     7.975** &                0.767 &                 0.775 &                                       0.137 &                0.393 &                 0.787 &                                     8.366** &                0.393 &                 0.769 &                                     6.704** \\
              &                              & 127 &                0.739 &                 0.393 &                                     6.839** &                0.393 &                 0.501 &                                     2.289** &                0.393 &                 0.779 &                                     8.771** &                0.393 &                 0.794 &                                     8.514** \\
              &                              & 451 &                0.394 &                 0.393 &                                       0.052 &                0.739 &                 0.778 &                                     0.549** &                0.393 &                 0.791 &                                     8.459** &                0.393 &                 0.722 &                                     5.741** \\
              & HateBERT & 12 &                0.758 &                 0.752 &                                       0.099 &                0.755 &                 0.780 &                                      0.406* &                0.753 &                 0.771 &                                       0.271 &                0.739 &                 0.751 &                                       0.159 \\
              &                              & 127 &                0.768 &                 0.754 &                                       0.144 &                0.725 &                 0.757 &                                      0.411* &                0.762 &                 0.770 &                                       0.150 &                0.761 &                 0.760 &                                       0.012 \\
              &                              & 451 &                0.760 &                 0.393 &                                     6.310** &                0.747 &                 0.776 &                                      0.407* &                0.768 &                 0.777 &                                       0.129 &                0.737 &                 0.780 &                                     0.695** \\
              & mBERT & 12 &                0.739 &                 0.719 &                                      0.285* &                0.393 &                 0.732 &                                     7.035** &                0.582 &                 0.736 &                                     2.129** &                0.676 &                 0.721 &                                     0.676** \\
              &                              & 127 &                0.740 &                 0.393 &                                     6.895** &                0.593 &                 0.639 &                                     0.598** &                0.682 &                 0.752 &                                     0.934** &                0.393 &                 0.738 &                                     6.722** \\
              &                              & 451 &                0.734 &                 0.745 &                                       0.179 &                0.732 &                 0.737 &                                       0.090 &                0.393 &                 0.746 &                                     7.218** &                0.719 &                 0.731 &                                       0.198 \\ \hdashline
\Di & BERT & 12 &                0.317 &                 0.349 &                                     1.573** &                0.349 &                 0.318 &                                     1.506** &                0.349 &                 0.768 &                                    12.256** &                0.349 &                 0.760 &                                    12.166** \\
              &                              & 127 &                0.349 &                 0.349 &                                       0.000 &                0.349 &                 0.732 &                                    11.640** &                0.349 &                 0.713 &                                    12.153** &                0.317 &                 0.771 &                                    13.692** \\
              &                              & 451 &                0.349 &                 0.349 &                                       0.000 &                0.349 &                 0.688 &                                    10.104** &                0.349 &                 0.349 &                                       0.000 &                0.349 &                 0.771 &                                    13.173** \\
              & BERTweet & 12 &                0.498 &                 0.349 &                                     3.944** &                0.349 &                 0.349 &                                       0.000 &                0.349 &                 0.765 &                                    14.378** &                0.349 &                 0.795 &                                    15.670** \\
              &                              & 127 &                0.317 &                 0.317 &                                       0.000 &                0.349 &                 0.730 &                                    10.885** &                0.349 &                 0.349 &                                       0.000 &                0.349 &                 0.813 &                                    15.126** \\
              &                              & 451 &                0.349 &                 0.349 &                                       0.000 &                0.317 &                 0.349 &                                     1.571** &                0.349 &                 0.777 &                                    14.698** &                0.349 &                 0.392 &                                     1.469** \\
              & HateBERT & 12 &                0.349 &                 0.691 &                                    10.451** &                0.349 &                 0.349 &                                       0.000 &                0.349 &                 0.775 &                                    13.318** &                0.349 &                 0.781 &                                    14.576** \\
              &                              & 127 &                0.349 &                 0.349 &                                       0.000 &                0.349 &                 0.727 &                                    11.989** &                0.349 &                 0.752 &                                    11.896** &                0.349 &                 0.785 &                                    13.631** \\
              &                              & 451 &                0.317 &                 0.349 &                                     1.571** &                0.349 &                 0.748 &                                    12.493** &                0.349 &                 0.742 &                                    10.092** &                0.349 &                 0.787 &                                    13.536** \\
              & mBERT & 12 &                0.349 &                 0.367 &                                     0.673** &                0.349 &                 0.349 &                                       0.009 &                0.349 &                 0.666 &                                     9.274** &                0.349 &                 0.716 &                                    10.999** \\
              &                              & 127 &                0.349 &                 0.619 &                                     7.138** &                0.349 &                 0.349 &                                       0.000 &                0.349 &                 0.675 &                                     9.671** &                0.349 &                 0.723 &                                    12.271** \\
              &                              & 451 &                0.317 &                 0.349 &                                     1.571** &                0.349 &                 0.380 &                                     1.141** &                0.349 &                 0.709 &                                     9.804** &                0.349 &                 0.724 &                                    11.119** \\ \hdashline
\Dv & BERT & 12 &                0.333 &                 0.333 &                                       0.045 &                0.333 &                 0.333 &                                       0.000 &                0.333 &                 0.333 &                                       0.000 &                0.333 &                 0.333 &                                       0.045 \\
              &                              & 127 &                0.333 &                 0.333 &                                       0.000 &                0.333 &                 0.333 &                                       0.045 &                0.333 &                 0.333 &                                       0.000 &                0.333 &                 0.333 &                                       0.000 \\
              &                              & 451 &                0.333 &                 0.333 &                                       0.000 &                0.333 &                 0.333 &                                       0.000 &                0.333 &                 0.333 &                                       0.045 &                0.333 &                 0.333 &                                       0.045 \\
              & BERTweet & 12 &                0.333 &                 0.333 &                                       0.045 &                0.333 &                 0.333 &                                       0.000 &                0.333 &                 0.333 &                                       0.045 &                0.333 &                 0.333 &                                       0.000 \\
              &                              & 127 &                0.333 &                 0.333 &                                       0.000 &                0.333 &                 0.333 &                                       0.045 &                0.333 &                 0.333 &                                       0.000 &                0.333 &                 0.333 &                                       0.045 \\
              &                              & 451 &                0.333 &                 0.333 &                                       0.045 &                0.333 &                 0.333 &                                       0.045 &                0.333 &                 0.333 &                                       0.000 &                0.333 &                 0.333 &                                       0.045 \\
              & HateBERT & 12 &                0.333 &                 0.333 &                                       0.000 &                0.333 &                 0.333 &                                       0.045 &                0.333 &                 0.599 &                                    16.715** &                0.333 &                 0.333 &                                       0.000 \\
              &                              & 127 &                0.333 &                 0.333 &                                       0.045 &                0.333 &                 0.333 &                                       0.000 &                0.333 &                 0.333 &                                       0.045 &                0.333 &                 0.333 &                                       0.045 \\
              &                              & 451 &                0.333 &                 0.333 &                                       0.000 &                0.333 &                 0.333 &                                       0.000 &                0.333 &                 0.333 &                                       0.045 &                0.333 &                 0.333 &                                       0.045 \\
              & mBERT & 12 &                0.333 &                 0.333 &                                       0.000 &                0.333 &                 0.333 &                                       0.045 &                0.333 &                 0.333 &                                       0.000 &                0.333 &                 0.333 &                                       0.000 \\
              &                              & 127 &                0.333 &                 0.333 &                                       0.000 &                0.333 &                 0.333 &                                       0.000 &                0.333 &                 0.333 &                                       0.000 &                0.333 &                 0.333 &                                       0.045 \\
              &                              & 451 &                0.333 &                 0.333 &                                       0.000 &                0.333 &                 0.333 &                                       0.000 &                0.333 &                 0.333 &                                       0.045 &                0.333 &                 0.333 &                                       0.000 \\ \hline
\end{tabular}
}
\caption{\textbf{RQ4:} Comparison of regional-wise macro F1 obtained under varying MLP seed ($ms$) for the BERT-variants. We measure the impact on performance when a region $R$ is set to trainable or unfrozen ($T$) vs. when it is non-trainable or frozen. ES stands for effect size. Further ** and * indicates whether the difference in macro F1 is significant by $\le0.05$ and $\le0.001$ p-value, respectively.}
\label{tab:region_wise_ptest}
\vspace{-3mm}
\end{table*}

\begin{table*}[!h]
\centering
\resizebox{0.625\textwidth}{!}{
\begin{tabular}{l|l|l|lll|lll}
\hline
              Dataset &  BERT-variant    & Seed     &  CH\textsubscript{S}: F1 &  CH\textsubscript{M}: F1 &  CH\textsubscript{C}: F1 &  CC\textsubscript{S,M}: ES &  CC\textsubscript{M,C}: ES &  CC\textsubscript{C,S}: ES \\ \hline
\Dii & BERT & 12 &   0.703 &   0.752 &    0.773 &             0.481** &                0.201 &              0.667** \\
              &                              & 127 &   0.668 &   0.766 &    0.776 &             0.627** &                0.066 &              0.704** \\
              &                              & 451 &   0.697 &   0.765 &    0.767 &             0.533** &                0.030 &              0.552** \\
              & BERTweet & 12 &   0.455 &   0.718 &    0.715 &             2.514** &                0.016 &              2.463** \\
              &                              & 127 &   0.454 &   0.734 &    0.731 &             2.939** &                0.070 &              2.609** \\
              &                              & 451 &   0.429 &   0.689 &    0.725 &             2.516** &               0.343* &              3.200** \\
              & HateBERT & 12 &   0.737 &   0.771 &    0.783 &              0.313* &                0.119 &               0.433* \\
              &                              & 127 &   0.751 &   0.781 &    0.787 &               0.236 &                0.073 &               0.319* \\
              &                              & 451 &   0.752 &   0.775 &    0.779 &               0.254 &                0.019 &               0.280* \\
              & mBERT & 12 &   0.666 &   0.738 &    0.742 &             0.621** &                0.014 &              0.622** \\
              &                              & 127 &   0.639 &   0.742 &    0.750 &             0.896** &                0.066 &              0.972** \\
              &                              & 451 &   0.644 &   0.742 &    0.744 &             0.832** &                0.026 &              0.858** \\ \hdashline
\Dvii & BERT & 12 &   0.781 &   0.722 &    0.811 &             0.800** &              1.229** &               0.453* \\
              &                              & 127 &   0.768 &   0.789 &    0.811 &               0.272 &               0.290* &              0.558** \\
              &                              & 451 &   0.771 &   0.813 &    0.738 &             0.551** &              0.905** &               0.355* \\
              & BERTweet & 12 &   0.604 &   0.693 &    0.741 &             0.968** &              0.480** &              1.472** \\
              &                              & 127 &   0.701 &   0.777 &    0.821 &             0.937** &              0.602** &              1.593** \\
              &                              & 451 &   0.626 &   0.786 &    0.797 &             1.802** &                0.165 &              1.979** \\
              & HateBERT & 12 &   0.824 &   0.842 &    0.850 &               0.275 &                0.148 &               0.423* \\
              &                              & 127 &   0.825 &   0.832 &    0.818 &               0.111 &                0.186 &                0.070 \\
              &                              & 451 &   0.813 &   0.829 &    0.843 &               0.195 &                0.200 &               0.397* \\
              & mBERT & 12 &   0.724 &   0.759 &    0.723 &              0.428* &               0.443* &                0.018 \\
              &                              & 127 &   0.698 &   0.764 &    0.713 &             0.850** &              0.670** &                0.127 \\
              &                              & 451 &   0.713 &   0.723 &    0.754 &               0.135 &               0.389* &              0.522** \\ \hdashline
\Div & BERT & 12 &   0.891 &   0.892 &    0.892 &               0.030 &                0.010 &                0.040 \\
              &                              & 127 &   0.890 &   0.894 &    0.891 &               0.168 &                0.128 &                0.046 \\
              &                              & 451 &   0.892 &   0.893 &    0.894 &               0.028 &                0.042 &                0.069 \\
              & BERTweet & 12 &   0.861 &   0.876 &    0.873 &              0.383* &                0.080 &               0.301* \\
              &                              & 127 &   0.855 &   0.879 &    0.873 &             0.693** &                0.157 &              0.523** \\
              &                              & 451 &   0.863 &   0.870 &    0.873 &               0.174 &                0.078 &                0.261 \\
              & HateBERT & 12 &   0.886 &   0.888 &    0.890 &               0.047 &                0.074 &                0.126 \\
              &                              & 127 &   0.883 &   0.886 &    0.888 &               0.086 &                0.053 &                0.134 \\
              &                              & 451 &   0.881 &   0.884 &    0.885 &               0.074 &                0.040 &                0.118 \\
              & mBERT & 12 &   0.840 &   0.849 &    0.846 &               0.224 &                0.058 &                0.162 \\
              &                              & 127 &   0.839 &   0.849 &    0.845 &               0.267 &                0.108 &                0.168 \\
              &                              & 451 &   0.840 &   0.852 &    0.848 &              0.327* &                0.108 &                0.209 \\ \hdashline
\Dvi & BERT & 12 &   0.672 &   0.685 &    0.720 &               0.028 &                0.154 &                0.185 \\
              &                              & 127 &   0.675 &   0.708 &    0.672 &               0.165 &                0.185 &                0.023 \\
              &                              & 451 &   0.640 &   0.733 &    0.677 &              0.311* &                0.149 &                0.145 \\
              & BERTweet & 12 &   0.419 &   0.674 &    0.630 &             1.051** &                0.160 &              0.817** \\
              &                              & 127 &   0.506 &   0.722 &    0.608 &             1.015** &              0.530** &               0.412* \\
              &                              & 451 &   0.453 &   0.707 &    0.582 &             0.966** &              0.483** &              0.455** \\
              & HateBERT & 12 &   0.659 &   0.742 &    0.730 &              0.421* &                0.074 &               0.341* \\
              &                              & 127 &   0.623 &   0.712 &    0.726 &              0.388* &                0.097 &              0.503** \\
              &                              & 451 &   0.674 &   0.699 &    0.726 &               0.147 &                0.113 &                0.260 \\
              & mBERT & 12 &   0.507 &   0.555 &    0.591 &               0.172 &                0.162 &               0.328* \\
              &                              & 127 &   0.538 &   0.617 &    0.647 &               0.239 &                0.117 &               0.348* \\
              &                              & 451 &   0.574 &   0.614 &    0.504 &               0.125 &               0.353* &                0.226 \\ \hdashline

hatexplain label & BERT & 12 &   0.661 &   0.661 &    0.685 &               0.010 &               0.358* &               0.363* \\
                 &                              & 127 &   0.677 &   0.679 &    0.676 &               0.045 &                0.037 &                0.009 \\
                 &                              & 451 &   0.674 &   0.688 &    0.692 &               0.230 &                0.035 &                0.274 \\
                 & BERTweet & 12 &   0.621 &   0.663 &    0.655 &             0.551** &                0.112 &               0.437* \\
                 &                              & 127 &   0.616 &   0.651 &    0.619 &             0.478** &               0.430* &                0.036 \\
                 &                              & 451 &   0.626 &   0.680 &    0.683 &             0.764** &                0.031 &              0.763** \\
                 & HateBERT & 12 &   0.691 &   0.697 &    0.714 &               0.076 &                0.228 &               0.309* \\
                 &                              & 127 &   0.677 &   0.705 &    0.709 &              0.391* &                0.067 &               0.450* \\
                 &                              & 451 &   0.708 &   0.715 &    0.724 &               0.097 &                0.150 &                0.238 \\
                 & mBERT & 12 &   0.655 &   0.660 &    0.663 &               0.052 &                0.047 &                0.101 \\
                 &                              & 127 &   0.658 &   0.670 &    0.658 &               0.163 &                0.163 &                0.002 \\
                 &                              & 451 &   0.647 &   0.654 &    0.637 &               0.086 &                0.240 &                0.155 \\ \hdashline
\Di & BERT & 12 &   0.658 &   0.673 &    0.663 &              0.316* &                0.219 &                0.086 \\
              &                              & 127 &   0.648 &   0.637 &    0.681 &               0.226 &              0.851** &              0.640** \\
              &                              & 451 &   0.663 &   0.663 &    0.674 &               0.020 &                0.201 &                0.231 \\
              & BERTweet & 12 &   0.622 &   0.628 &    0.564 &               0.128 &              1.271** &              1.105** \\
              &                              & 127 &   0.590 &   0.607 &    0.496 &              0.381* &              2.464** &              2.076** \\
              &                              & 451 &   0.571 &   0.611 &    0.608 &             0.825** &                0.065 &              0.771** \\
              & HateBERT & 12 &   0.686 &   0.707 &    0.703 &             0.493** &                0.095 &               0.367* \\
              &                              & 127 &   0.681 &   0.657 &    0.702 &             0.512** &              0.969** &               0.461* \\
              &                              & 451 &   0.685 &   0.709 &    0.696 &             0.532** &               0.282* &                0.232 \\
              & mBERT & 12 &   0.641 &   0.644 &    0.547 &               0.052 &              1.894** &              1.908** \\
              &                              & 127 &   0.577 &   0.648 &    0.649 &             1.621** &                0.018 &              1.514** \\
              &                              & 451 &   0.626 &   0.650 &    0.648 &             0.490** &                0.036 &               0.459* \\ \hdashline
\Dv & BERT & 12 &   0.777 &   0.800 &    0.801 &             1.407** &                0.052 &              1.468** \\
              &                              & 127 &   0.776 &   0.802 &    0.802 &             1.450** &                0.003 &              1.509** \\
              &                              & 451 &   0.778 &   0.801 &    0.801 &             1.368** &                0.000 &              1.407** \\
              & BERTweet & 12 &   0.753 &   0.770 &    0.770 &             0.898** &                0.062 &              0.916** \\
              &                              & 127 &   0.753 &   0.770 &    0.769 &             0.723** &                0.027 &              0.670** \\
              &                              & 451 &   0.753 &   0.771 &    0.772 &             1.033** &                0.045 &              1.111** \\
              & HateBERT & 12 &   0.776 &   0.806 &    0.809 &             1.882** &                0.182 &              1.986** \\
              &                              & 127 &   0.777 &   0.807 &    0.808 &             1.557** &                0.116 &              1.989** \\
              &                              & 451 &   0.777 &   0.806 &    0.807 &             1.534** &                0.070 &              1.539** \\
              & mBERT & 12 &   0.735 &   0.757 &    0.758 &             1.182** &                0.061 &              1.233** \\
              &                              & 127 &   0.736 &   0.757 &    0.758 &             1.228** &                0.017 &              1.250** \\
              &                              & 451 &   0.736 &   0.756 &    0.758 &             1.134** &                0.140 &              1.329** \\ \hline
\end{tabular}
}
\caption{\textbf{RQ5:} Comparison of maximum macro F1 obtained under varying MLP seed ($ms$) for the simple ($S$), medium ($M$) and complex ($C$) classification heads ($CH$). $CH_{x,y}$ captures the difference in performance when comparing the given configuration under heads $x$ and $y$. ES stands for effect size. ** and * indicates whether the difference in maximum macro F1 is significant by $\le0.05$ and $\le0.001$ p-value, respectively.}
\label{tab:cc_wise_ptest}
\vspace{-3mm}
\end{table*}

\end{document}